\documentclass[10pt,final,journal]{IEEEtran}

\usepackage{algorithm,algorithmicx,algpseudocode} 
\usepackage{amsmath,amsfonts,amssymb,amsthm,mathtools} 
\usepackage{pifont,stmaryrd} 
\usepackage{graphicx,epstopdf,caption,subcaption} 
\usepackage{threeparttable,multirow} 
\usepackage[colorlinks]{hyperref} 
\usepackage[normalem]{ulem} 
\usepackage[usenames]{xcolor} 
\usepackage{cite} 
\usepackage{blindtext} 
\usepackage[utf8]{inputenc}
\usepackage[T1]{fontenc}
\usepackage{listings}
\usepackage{xparse}
\usepackage{balance}
\usepackage[export]{adjustbox}
\IEEEoverridecommandlockouts

\newcommand{\bA}        {\mathbf{A}}

\newcommand{\bB}        {\mathbf{B}}

\newcommand{\E}         {\mathbb{E}}

\newcommand{\be}        {\mathbf{e}}

\newcommand{\cE}        {\mathcal{E}}
\newcommand{\bcE}       {\boldsymbol{\cE}}

\newcommand{\bI}        {\mathbf{I}}

\newcommand{\bK}        {\mathbf{K}}

\newcommand{\bM}        {\mathbf{M}}

\newcommand{\N}         {\mathbb{N}}

\newcommand{\cN}        {\mathcal{N}}

\newcommand{\cO}        {\mathcal{O}}

\newcommand{\bq}        {\mathbf{q}}
\newcommand{\bQ}        {\mathbf{Q}}

\newcommand{\R}         {\mathbb{R}}

\newcommand{\bR}        {\mathbf{R}}

\newcommand{\bu}        {\mathbf{u}}

\newcommand{\bv}        {\mathbf{v}}
\newcommand{\bV}        {\mathbf{V}}

\newcommand{\bW}        {\mathbf{W}}

\newcommand{\bx}        {\mathbf{x}}
\newcommand{\bX}        {\mathbf{X}}

\newcommand{\bZ}        {\mathbf{Z}}

\newcommand{\bSigma}    {\boldsymbol{\Sigma}}

\newcommand{\bone}      {\mathbf{1}}

\DeclareMathOperator*{\argmin}  {arg\,min}                              

\newcommand{\tT}        {\mathrm{T}}                                    

\newcounter{counter}

\newcommand{\lowerromannumeral}[1]{\setcounter{counter}{#1}\roman{counter}}
\theoremstyle{plain}
\newtheorem{lemma}{Lemma}
\newtheorem{theorem}{Theorem}

\newtheorem{proposition}{Proposition}

\theoremstyle{definition}

\theoremstyle{remark}
\newtheorem{remark}{Remark}

\graphicspath{{images/}} 

\begin{document}
\title{Distributed Principal Subspace Analysis for Partitioned Big Data: Algorithms, Analysis, and Implementation}

\author{Arpita Gang, Bingqing Xiang, and Waheed U.\ Bajwa
%
\thanks{A.\ Gang and B.\ Xiang contributed equally to the paper. The results reported in this paper first appeared in the MS thesis of B.\ Xiang~\cite{bqthesis.2020}, which was completed within the Department of Electrical and Computer Engineering, Rutgers University--New Brunswick, NJ 08854 in 2020. A.\ Gang and W.\ U.\ Bajwa are with the Department of Electrical and Computer Engineering, Rutgers University--New Brunswick, NJ 08854 (Emails: {\tt \{arpita.gang,~waheed.bajwa\}@rutgers.edu}). B.\ Xiang is now with ByteDance Ltd.\ (Email: {\tt xiangbqxyy@gmail.com}). }%
%
\thanks{This work was supported in part by the National Science Foundation under Awards CCF-1453073, CCF-1907658, and OAC-1940074, and by the Army Research Office under Awards W911NF-17-1-0546 and W911NF-21-1-0301.}}


\maketitle

\makeatletter
\setlength{\@fptop}{0pt}
\DeclarePairedDelimiter\abs{\lvert}{\rvert}%
\DeclarePairedDelimiter\norm{\lVert}{\rVert}%

\let\oldabs\abs
\def\abs{\@ifstar{\oldabs}{\oldabs*}}

\let\oldnorm\norm
\def\norm{\@ifstar{\oldnorm}{\oldnorm*}}
\makeatother
\begin{abstract}
Principal Subspace Analysis (PSA)---and its sibling, Principal Component Analysis (PCA)---is one of the most popular approaches for dimensionality reduction in signal processing and machine learning. But centralized PSA/PCA solutions are fast becoming irrelevant in the modern era of big data, in which the number of samples and/or the dimensionality of samples often exceed the storage and/or computational capabilities of individual machines. This has led to the study of distributed PSA/PCA solutions, in which the data are partitioned across multiple machines and an estimate of the principal subspace is obtained through collaboration among the machines. It is in this vein that this paper revisits the problem of distributed PSA/PCA under the general framework of an arbitrarily connected network of machines that lacks a central server. The main contributions of the paper in this regard are threefold. First, two algorithms are proposed in the paper that can be used for distributed PSA/PCA, with one in the case of data partitioned across samples and the other in the case of data partitioned across (raw) features. Second, in the case of sample-wise partitioned data, the proposed algorithm and a variant of it are analyzed, and their convergence to the true subspace at linear rates is established. Third, extensive experiments on both synthetic and real-world data are carried out to validate the usefulness of the proposed algorithms. In particular, in the case of sample-wise partitioned data, an MPI-based distributed implementation is carried out to study the interplay between network topology and communications cost as well as to study the effects of straggler machines on the proposed algorithms.
%
\end{abstract}


\section{Introduction}\label{sec:intro}
In the current world of machine learning, data tends to be huge in both dimension and size, i.e., the number of samples. To tackle the massiveness of dimension, measures have to be taken to reduce the data dimensionality, which aids in storage and subsequent processing of the data. Also, the massiveness of size of the data makes it difficult to store and process the data at a single location/machine and hence use of multiple units has become inevitable. This motivates the need to explore distributed dimensionality reduction solutions, wherein one can keep data distributed across machines and still process them together. The most fundamental tool for dimension reduction is Principal Component Analysis (PCA)~\cite{hotelling1933analysis}, which extracts a smaller set of uncorrelated features from the data that carry maximum information. Quite often though, one only needs a smaller set of features that approximate the data well enough and uncorrelatedness is not a necessary condition. This technique is more appropriately called Principal Subspace Analysis (PSA), which falls under the larger umbrella of low-rank approximation techniques~\cite{lowrank2017literature}. PSA~\cite{Pearson.1901} is an unsupervised learning technique that is used for dimension reduction of data, before utilizing it for further applications like classification, regression, etc., to help with faster processing and computations. These aforementioned reasons are the motivations for this paper in which we explore PSA/PCA in a distributed environment so as to derive a smaller set of important data features efficiently when data is distributed across machines.

Mathematically speaking, for a data point $\bx \in \R^{d}$, PSA aims to represent it by a smaller $r$-dimensional vector $\tilde{\bx} \in \R^r$ $(r \ll d)$ such that it is an `efficient' representation of $\bx$. This is accomplished by finding an $r$-dimensional subspace, represented by its orthonormal basis $\bQ \in \R^{d\times r}$, such that $\tilde{\bx}= \bQ^\tT\bx$ has features that retain maximum information contained in original data point $\bx \in \R^d$. In other words, when $\bx$ is reconstructed from $\tilde{\bx}$ as $\bQ\tilde{\bx} = \bQ\bQ^\tT\bx$ (subject to $\bQ^\tT\bQ = \bI$), it has the minimum approximation error in Frobenius norm. For data samples drawn from any distribution, the directions that contain maximum information (energy) are given by the leading eigenvectors of the covariance matrix of that distribution~\cite{baldi.hornik.1989}. This implies the subspace that would retain the most amount of information is the one spanned by those eigenvectors, i.e., the principal eigenspace. Thus, dimension reduction that would result in a smaller set of features can be achieved only when the said matrix $\bQ$ is the basis of the principal eigenspace of the data covariance matrix $\bSigma = \E\begin{bmatrix}\bx\bx^\tT\end{bmatrix}$. The PCA problem, in addition, requires $\bQ$ to be precisely the eigenvectors of $\bSigma$, as opposed to any orthonormal basis of the principal eigenspace.


Even though principal eigenspace estimation is a well studied problem when data is available at a single location, the enormity of the amount of available data or inherent distributed nature of the data generation like in Internet-of-Things, from an array of sensors, etc., makes it absolutely necessary to look for solutions for the case when data is distributed across locations. Collating such data at one location can be prohibitive due to storage and computation constraints and/or to maintain the privacy of data. It is in this regard that we first and foremost aim to find solutions for PSA in a distributed setup. Interestingly, however, our algebraic approach to the PSA problem ends up being applicable to distributed PCA also in the case of the covariance matrix having distinct eigenvalues. Nonetheless, to keep the exposition simple, we mainly limit ourselves to usage of the term ``distributed PSA'' in much of the remainder of this section. 

Note that distributed setups can be broadly of two types: \lowerromannumeral{1}) when all the entities (data centers, sensors, etc.) are connected to a central server, and \lowerromannumeral{2}) when the entities are connected as an arbitrary network without any central server. The terms distributed and decentralized are interchangeably used for both these setups in the literature and are explained in more detail in~\cite{Yang.Gang.Bajwa.2020}. In this paper, we focus on the latter kind of setting with no central server because of its more general architecture; here onwards we use the term \textit{distributed} for the setup and the term \textit{nodes} for the entities forming the distributed network.

Within any distributed setting, splitting of the data among the nodes can happen in two ways: \lowerromannumeral{1}) by samples, and \lowerromannumeral{2}) by raw features. Sample-wise splitting means each node has access to some but not all samples of the data, but each sample has its full set of raw features. This kind of data partitioning naturally occurs in cases like Internet-of-Things, where devices are scattered geographically, each device (node) carries a subset of the entire information (samples) spread across the network and the data cannot be brought together for reasons like privacy or communication bandwidth constraints. The feature-wise splitting means each node has few features for all samples of data. A natural example of this type of data partitioning occurs in sensor array applications, where different sensors capture different parts of the same signal. In this paper, we consider both kinds of data partitioning and propose distributed PSA algorithms for each of them. The end goal in each case is to find the principal eigenspace of the covariance matrix when data is distributed across a network.

\subsection{Relationship to Prior Work}\label{ssec:prior_work}
PCA and PSA are age-old tools for dimensionality reduction with seminal work appearing as early as 1901 in~\cite{Pearson.1901}. In~\cite{hotelling1933analysis}, Hotelling proposed a solution for estimating the eigenvectors of data covariance matrix to compress a set of data points. Since then many more solutions for dimensionality reduction have been proposed, which include iterative methods like power method, orthogonal iteration~\cite{van1983matrix}, and Lanczos algorithm~\cite{Lanczos.1950}. These methods are shown to have convergence guarantees for subspace estimation in case of symmetric matrices, a category covariance matrices fall under. Data compression has also been a topic of interest in the neural network community, with autoencoders being an important tool for data compression. The work in~\cite{baldi.hornik.1989} showed that a single-layer fully connected autoencoder that has linear activation and squared error cost function will have weights given by the space spanned by the eigenvectors, i.e., the eigenspace of the input covariance matrix. Thus autoencoders are efficient tools for principal eigenspace estimation. 

In contrast to the centralized setting, solutions for PSA in the distributed setup are very recent and few. As noted earlier, the partitioning of data is possible in two ways: by features and by samples. In the case when the partitioning is by features of the data, each node estimates one or a subset of the features of the eigenspace. For this particular kind of partitioning, the work in~\cite{kempe2008decentralized} estimates top-$r$ eigenvectors of the graph adjacency matrix of a network, while another significant work in~\cite{scaglione2008decentralized} proposed an algorithm for estimation of top-$r$ eigenvectors of the covariance matrix sequentially, starting from the eigenvector corresponding to the largest eigenvalue. This sequential approach slows down the convergence of the algorithm when a higher-dimensional eigenspace needs to be estimated. To speed up the subspace estimation process, an ideal situation here would be to estimate all the basis vectors simultaneously rather than one-by-one sequentially. Furthermore, the detailed analysis of the subspace estimation algorithm in~\cite{scaglione2008decentralized} given in~\cite{esprit} shows that this sequential approach requires the $(r+1)$ largest eigenvalues of the covariance matrix to be distinct, which is a strong condition. To address these issues for the case of feature-wise partitioned data, we propose an algorithm based on orthogonal iterations (OI) to find the principal eigenspace of the covariance matrix simultaneously by using a distributed QR factorization algorithm~\cite{strakova2011distributed}.

When data is partitioned by samples, even though each node has access to few samples, the goal is that every node estimates complete eigenspace of the covariance matrix of the entire data. In addition, all nodes need to agree with each other, i.e., a consensus in the network is an important requirement for distributed solutions in this case. The works in~\cite{raja2015cloud, cksvd.allerton.2013, depm} give solutions for this particular kind of distributed setup, proposing a variant of the power method. These methods focus on extracting only the top eigenvector and have been shown to converge at a linear rate by using explicit consensus iterations~\cite{xiao2004fast} after each iteration of the power method to ensure the nodes in the network agree with each other. Although estimation of the next dominant eigenvectors can be done sequentially using the distributed power method, the convergence analysis provided in these papers are only for the dominant eigenvector. Additionally, similar to feature-wise partitioned case, using distributed power method for sequentially estimating the subspace basis vectors would require distinct eigenvalues since that is a basic requirement of power method for convergence. Another method for the estimation of top eigenvector in distributed but streaming data case was proposed in~\cite{raja.bajwa.2020}. A recently proposed method in~\cite{gang.raja.bajwa.2019} uses a Hebbian update rule in the distributed setting to find top-$r$ eigenvectors and is proved to converge linearly to a neighbourhood of the true solution~\cite{gang.bajwa.2021}. The review paper~\cite{wu2018review} provides a detailed coverage of distributed PCA/PSA solutions for both types of data partitioning, namely, by features and by samples (referred to as DRO and DCO, respectively, therein). 

Note that PSA is a nonconvex problem due to its nonconvex constraint that the solution must lie on the Stiefel manifold. Recently, some work has also been done for solving general nonconvex problems in the distributed setting that can be related to sample-wise distributed PSA problem in some sense. The work in~\cite{next} does convex approximations of a nonconvex objective function but assumes that the constraint set is convex, while~\cite{proxpda} shows convergence to a stationary point of unconstrained nonconvex problems. The method in~\cite{wai.scaglione.lafond.2016} also requires the constraint set to be convex in case of nonconvex objective functions. A recent work in~\cite{mingyi.2021.decentralized} proposes a Riemannian gradient descent method for optimization of nonconvex problems over a Stiefel manifold in a distributed network. It is shown to converge only to a stationary point of the nonconvex function. Thus, none of these methods are directly applicable to the PSA problem in the distributed setup. In this paper, we propose an orthogonal iterations-based approach that uses consensus averaging as a solution to the sample-wise distributed PSA problem. This is an extension of the distributed power method algorithm proposed as a subroutine in~\cite{raja2015cloud} to the case of $r>1$ and is shown to converge to the eigenspace of the covariance matrix at linear rate without the strong assumption of distinct top-$(r+1)$ eigenvalues of the covariance matrix.
\subsection{Our Contributions}
The main contributions of this paper are \lowerromannumeral{1}) a novel algorithm for feature-wise distributed PSA called F-DOT, \lowerromannumeral{2}) a novel algorithm for sample-wise distributed PSA called S-DOT along with a variant SA-DOT that adaptively changes the number of consensus iterations for each orthogonal iteration, \lowerromannumeral{3}) theoretical convergence guarantees for S-DOT and SA-DOT, \lowerromannumeral{4}) experiments that use Message Passing Interface (MPI)~\cite{mpi.1993} to understand communication cost in real-world settings, and \lowerromannumeral{5}) extensive numerical experiments to demonstrate the efficiency of all the proposed algorithms as compared to existing distributed and baseline methods. 

The main goal of this paper is to find solutions for PSA when data is partitioned either by features or by samples over an arbitrary network of interconnected nodes. To fulfill the purpose of dimension reduction in the distributed setting for the two types of mentioned data splits, we propose algorithms that would find the principal eigenspace of the data covariance matrix even in the absence of a central entity that can collate the data or co-ordinate among the nodes. Orthogonal iteration (OI) is a very useful algorithm for eigenspace estimation in centralized settings~\cite{van1983matrix} and it also forms the fundamental building block of all our proposed solutions. Maintaining orthonormality in case of F-DOT and network consensus in case of S-DOT and SA-DOT requires careful considerations while adapting OI to the distributed setup. The theoretical guarantees of the S-DOT and SA-DOT algorithms show that our proposed solution has linear convergence rates for the case of a subspace with $r>1$, unlike the existing theoretical results in the literature that only provide guarantees for the case of $r=1$. Extensive experimental results are presented that further support our claims. Even though we do not provide any theoretical guarantees for F-DOT algorithm, experimental simulations demonstrate its efficiency. For extensive experimental study, we have also simulated real-world distributed networks using the MPI protocol as well as studied the effects of various parameters associated with the algorithms like network connectivity, data dimension, etc. Finally, as noted earlier, since our distributed PSA developments are based on OI, they generalize to the distributed PCA problem in the case of distinct top-$(r+1)$ eigenvalues of the covariance matrix~\cite{qr1982}. Going forward, however, we do not insist on distinct eigenvalues and, as such, limit ourselves to the distributed PSA problem.

\begin{remark}\label{remark:remark1}
During the revision of this paper, whose results first appeared in~\cite{bqthesis.2020}, a related work~\cite{ye2021deepca} for distributed PSA of sample-wise partitioned data appeared as a preprint. Both~\cite{ye2021deepca} and our work are extensions of the ideas in our prior work~\cite{raja2015cloud}. The authors in~\cite{ye2021deepca} have made use of the idea of ``gradient tracking'' from distributed optimization literature~\cite{next,extra} to improve on the communications cost of distributed PSA. When compared to this work, our method has the same algorithmic complexity but the communications complexity has an additional log factor. Nonetheless, the work in this paper predates~\cite{ye2021deepca}; in addition, we also discuss feature-wise partitioned data and carry out an extensive MPI-based implementation that helps study the impacts of different real-world design choices and constraints on distributed PSA solutions.
\end{remark}

\subsection{Notation and Organization}
The following notational convention is used throughout the rest of this paper. We use the standard notation $:=$ to denote definitions of terms. The notation $|\cdot|$ is used for both the cardinality of a set and the absolute value of a real number. Similarly, $\|\cdot\|_2$ is used for both the $\ell_2$-norm of a vector and the operator 2-norm of a matrix. The notation $\setminus$ denotes the set difference operation. Finally, we make use of the following ``\emph{Big--O}'' notation for scaling relations: $f(n) = \cO(g(n))$ if $\exists c_o > 0, n_o : \forall n \geq n_o, f(n) \leq c_o g(n)$, and $f(n) = \Omega(g(n))$ if $g(n) = \cO(f(n))$.

The rest of this paper is organized as follows: In Section~\ref{sec:formulation}, we describe and mathematically formulate the distributed PSA problem for both kinds of data partitioning. Section~\ref{sec:algorithm} describes the three proposed algorithms, while Section~\ref{sec:analysis} provides convergence analysis of the S-DOT and SA-DOT algorithms, and discusses the computational complexity and communication cost of the three algorithms. We provide numerical results in Section~\ref{sec:results} to show efficacy of the proposed methods and conclude in Section~\ref{sec:conc}. The detailed proofs of our main mathematical results are in Appendix~\ref{app:lemma1n2} and Appendix~\ref{app:thm1n2}.

\section{Problem Formulation}\label{sec:formulation}
The goal of principal subspace analysis (PSA) is to compress data without losing much information. Specifically, to compress a data point $\bx \in \R^d$ such that it has only $r$ $(r \ll d)$ features, PSA finds the $r$-dimensional eigenspace spanned by the eigenvectors corresponding to the $r$ largest eigenvalues of the population covariance matrix $\bSigma = \E\begin{bmatrix}(\bx-\E\begin{bmatrix}\bx\end{bmatrix})(\bx-\E\begin{bmatrix}\bx\end{bmatrix})^\tT\end{bmatrix}$. If the resulting eigenspace is given as $\bQ = \begin{bmatrix}\bq_1, \ldots, \bq_r\end{bmatrix} \in \R^{d \times r}$, then the reduced set of features will be given by $\bQ^\tT\bx$. In practice the actual distribution and hence $\bSigma$ is unknown, and therefore a sample covariance matrix is used instead. For the data matrix $\bX = \begin{bmatrix}\bx_1,\ldots, \bx_n\end{bmatrix} \in \R^{d \times n}$ with sample mean $\bar{\bx} = \frac{1}{n}\sum_{t=1}^{n}\bx_t$, the sample covariance matrix is $\bM = \frac{1}{n-1}\sum_{t=1}^{n}(\bx_t - \bar{\bx})(\bx_t - \bar{\bx})^\tT$. Without loss of generality, we will assume $\bar{\bx} = 0$, since even otherwise the sample mean can be easily computed and subtracted from the samples, thus making the sample covariance matrix $\bM = \frac{1}{n}\sum_{t=1}^{n}\bx_t\bx_t^\tT = \frac{1}{n}\bX\bX^\tT$. With the goal of finding the subspace that can be used to reconstruct data points with minimum error, PSA is formulated in the centralized case as:
\begin{align}\nonumber
    \bQ_c &= \underset{\bQ_c \in \R^{d\times r}}  \argmin \quad f(\bQ_c) = \underset{\bQ_c \in \R^{d\times r}} \argmin \quad \|(\bI - \bQ_c\bQ_c^\tT)\bX\|_F^2\\ \label{eq:psa}
    &\qquad \text{such that} \quad  \bQ_c^\tT\bQ_c = \bI.
\end{align}
The constraint $\bQ_c^\tT\bQ_c = \bI$ implies that the solution should lie on the Stiefel manifold. This formulation returns an orthogonal basis of the $r$-dimensional eigenspace of $\bM$. Not only do we want a solution to the PSA problem~\eqref{eq:psa} in this paper, we are also looking at an added challenge of non-availability of data at a single location, thus requiring to solve PSA in a distributed manner. We consider the following distributed setup for this problem: a network that is defined by an undirected graph given as $\mathcal{G}=(\mathcal{N}, \mathcal{E})$ where $\mathcal{N} = \{1,2,\dots,N\}$ is the set of nodes in the network and $\mathcal{E}$ is the set of edges $(i,j)$. For each node $i$, we record its neighbors (including itself) in the set $\mathcal{N}_i=\{j| (i,j) \in \mathcal{E}\} \cup i$. 

\subsection{The Types of Data Partitions}
As mentioned earlier, data partitioning is most commonly done in two major ways: by samples and by features. In case of sample-wise distribution, mathematically, each node $i$ consists of a set of samples denoted by $\bX_i \in \R^{d \times n_i}$ such that $\sum_{i=1}^{N}n_i = n$. The local covariance matrix at node $i$ is thus $\bM_i = \frac{1}{n_i}\bX_i\bX_i^\tT$ and it is straightforward to see that $n\bM = \sum_{i=1}^{N}n_i\bM_i$. Also, every node $i$ maintains its own copy $\bQ_{s,i}$ of the true estimate $\bQ_s$ in the absence of any central server. Thus for node $i$, if we were to focus on local PSA only then~\eqref{eq:psa} can be re-written as follows:
\begin{align}\nonumber
     \bQ_{s,i} &= \underset{\bQ_{s,i} \in \R^{d \times r}}\argmin  \left [f_i(\bQ_{s,i})\coloneqq \norm{(\bI-\bQ_{s,i}\bQ_{s,i}^\tT)\bX_i}^2_F \right] \\
     &\qquad\text{such that} \quad \bQ_{s,i}^\tT\bQ_{s,i} = \bI.
\end{align}
Through collaboration, however, the ultimate goal is that all nodes reach the same estimate of the space spanned by the eigenvectors of the global covariance matrix $\bM$, i.e., $\bQ_{s,1}=\bQ_{s,2}=\ldots=\bQ_{s,N}=\bQ_s$. Thus, the overall optimization problem to be solved in the network is:
\begin{align}\nonumber
     &\underset{\{\bQ_{s,i}^\tT\bQ_{s,i} = \bI\}_{i=1}^{N}}\argmin  \sum_{i=1}^{N}\left [f_i(\bQ_{s,i})\coloneqq \norm{(\bI-\bQ_{s,i}\bQ_{s,i}^\tT)\bX_i}^2_F \right] \\ \label{eq:dpsa_sample}
     &\text{such that} \quad \bQ_{s,1}=\bQ_{s,2}=\ldots=\bQ_{s,N}=\bQ_s .
\end{align}
Note that if $\bQ_{s,1}=\bQ_{s,2}=\ldots=\bQ_{s,N}=\bQ_s$, $\sum_{i=1}^{N}f_i(\bQ_{s,i}) = f(\bQ_{s})$, which is consistent with the formulation~\eqref{eq:psa} of centralized PSA. 

In the case of feature-wise partitioning, the view of distributed PSA is significantly different from the sample-wise case. Here, for a data sample $\bx_t \in \R^d$, each node $i$ has access to some of the $d$ features of the sample, i.e., node $i$ has access to a part $\bX_i \in \R^{d_i \times n}$ of the complete data such that $\sum_{i=1}^{N}d_i = d$. The goal of distributed PSA in this case is that each node $i$ learns a part $\bQ_{f,i}\in \R^{d_i \times r}$ of the estimate of eigenspace of $\bM$ by using its local data $\bX_i$ and collaborating with other nodes in such a way that $\bQ_f = \begin{bmatrix}\bQ_{f,1}^\tT,\ldots, \bQ_{f,N}^\tT\end{bmatrix}^\tT$ represents the estimate of $\bQ$, the whole $r$-dimensional eigenspace. Unlike the sample-wise partitioned case, the centralized PSA formulation~\eqref{eq:psa} is inseparable in the feature-wise partitioned case.

It is well known that orthogonal iteration (OI)~\cite{van1983matrix} is an iterative method that finds the dominant $r$-dimensional eigenspace of a symmetric matrix $\bM$ at a linear rate under the assumption that if $\lambda_1, \ldots, \lambda_d$ are its eigenvalues then the condition $\lambda_1 \geq \lambda_2 \ldots \geq \lambda_r > \lambda_{r+1}\geq \ldots \geq \lambda_d$ holds true. In both cases of partitions described here, the unavailability of $\bX$ and hence $\bM$ at a single location makes the centralized OI solution unusable, unless the data is collected at a single location. Since this is often impossible as discussed before, we aim to modify OI such that it can be used in distributed networks for both feature-wise and sample-wise data partitions.


\section{Proposed Algorithms}\label{sec:algorithm}


Even though orthogonal iteration (OI) is a simple and effective solution when the matrix whose eigenspace is to be computed is available at a single location, using it in either sample-wise or feature-wise data partitioned case has its challenges. The sample-wise distributed case requires all nodes in an arbitrarily connected network to reach a common solution given by the eigenspace of $\bM$ without having access to entire matrix at any of the nodes. The nodes are only allowed to collaborate with their immediate neighbors and not exchange any raw data. In feature-wise case, consensus is not a requirement but each node is required to compute a part of the eigenvectors of $\bM$ while it is not available in entirety at any one node. Even though there is no common solution that the nodes have to reach, collaboration is still a vital part here to maintain the orthogonality of the estimated solution. We propose algorithms to deal with these challenges and use OI effectively in both kinds of data partitioning settings.

\subsection{PSA for Sample-wise Partitioned Data}
We begin with the setup where data is partitioned by samples, i.e., each node has access to a few samples stored in $\bX_i$, resulting in a local covariance matrix $\bM_i$. Ignoring the scaling factors as those do not affect the eigenspace, one can write $\bM = \sum_{i=1}^{N}\bM_i$. Under the eigengap assumption required for OI, we first propose an algorithm \textit{Sample-wise Distributed Orthogonal iTeration (S-DOT)} that estimates the dominant $r$-dimensional eigenspace of $\bM$ at each node $i$ while using only its local data $\bM_i$ and a subroutine called consensus averaging~\cite{xiao2004fast}. The complete algorithm is given in Algorithm~\ref{alg:cdot}.

S-DOT is a two-scale iterative method, where for each iteration of OI (outer loop) performed locally at each node, there is an inner loop of $T_c$ consensus iterations. We define $\bQ_{s,i}^{(t)}$ as the estimate of $\bQ_s$ at node $i$ after $t$ iterations of the outer loop. Now during the outer loop orthogonal iteration $t$, each node locally computes the product $\bM_i\bQ_{s,i}^{(t-1)}$ as given in Step~\ref{algostep:doi} of Algorithm~\ref{alg:cdot}. Then, we apply $T_c$ iterations of consensus averaging using a doubly stochastic weight matrix $\bW$ defined based on the graph topology to approximate $\frac{1}{N}\sum _{i=1}^N \bM_i\bQ_{s,i}^{(t-1)}$. It is known that if $T_c \rightarrow \infty$, then the averaging would be exact~\cite{xiao2004fast}. Let us assume for a moment that $\bQ_{s,i}^{(t-1)} = \bQ_s^{(t-1)} \forall i$, then Step~\ref{algostep:doi} at node $i$ would be $\bZ_i^{(0)}=\bM_i\bQ_{s}^{(t-1)}$. Performing exact consensus averaging step $\bZ_i^{(t_c)}= \sum_{j \in \mathcal{N}_i} w_{i,j}\bZ_j^{(t_c-1)}$ infinitely many times on these resulting $\bZ_i^{(0)}$ will result in $\bZ_i^{(\infty)} = \frac{1}{N}\sum _{j=1}^N \bM_j\bQ_{s}^{(t-1)} = \frac{1}{N}\bM\bQ_{s}^{(t-1)}$, which is the same as an update of centralized OI at all nodes across the network. This shows that using averaging consensus can lead to the eigenspace of the global covariance matrix $\bM$ at each node $i$. However, infinite consensus iterations is not possible in the real world for any $t$ and hence after a finite number of consensus iterations $T_c$, each $\bV_{s,i}^{(t)} = \frac{\bZ_i^{(T_c)}}{[\bW^{T_c}\be_1]_i}$, where $\be_1 = \begin{bmatrix}1,0,\ldots,0\end{bmatrix}^T$, incurs some error due to imperfect averaging, i.e., $\bV_{s,i}^{(t)} =\sum_{j=1}^N\bM_j\bQ_{s,j}^{(t-1)}+\bcE_{c,i}^{(t)}$. Quantifying the error $\bcE_{c,i}^{(t)},\,\forall i,t$ is one of our main contributions in convergence analysis. In the final step of the $t^{th}$ outer loop iteration, every node locally performs a QR decomposition of $\bV_{s,i}^{(t)}$ to ensure that the estimated basis vectors are orthonormal.


\begin{algorithm}
\caption{Sample-wise Distributed Orthogonal Iteration}\label{alg:cdot}
\begin{algorithmic}[1]

\State {\textbf{Input:} $\bW$; $\bM_i, i = 1,\ldots, N$}
\State{\textbf{Initialize:} Set $t \leftarrow 0$ and $\bQ^{(t)}_{s,i} \leftarrow \bQ_{\text{init}}$ where $\bQ_{\text{init}} \in \R^{d \times r}: \bQ_{\text{init}}^\tT\bQ_{\text{init}} = \bI$ }

    \While{stopping criteria} 
        \State {$t \leftarrow t+1$}
        \State {$\bZ_i^{(0)}\leftarrow \bM_i \bQ_{s,i}^{(t-1)},i=1,2,\ldots,N$} \label{algostep:doi}
        \State {\textbf{Begin consensus loop:} $t_c \leftarrow 0$}
        \While{$t_c < T_c$}
            \State{$ t_c \leftarrow t_c+1 $}
            \State{$ \bZ_i^{(t_c)}\leftarrow \sum_{j \in \mathcal{N}_i} w_{i,j}\bZ_j^{(t_c-1)} $}
        \EndWhile
        \State{$\bV_{s,i}^{(t)} \leftarrow \frac{\bZ_i^{(t_c)}}{[\bW^{T_c}\be_1]_i}$}\label{algostep:consensus_result}
        \State{$\bQ_{s,i}^{(t)}, \bR_{s,i}^{(t)} \leftarrow  \text{QR factorization}(\bV_{s,i}^{(t)})$} \label{algostep:doi_qr}
    \EndWhile
\State {\textbf{Return:} $ \bQ_{s,i}^{(t)}$}
\end{algorithmic}
\end{algorithm}

It is well known that OI converges, i.e., the principal angle between the subspaces spanned by $\bQ$ and $\bQ_{s,i}^{(t-1)}$ is larger than that between $\bQ$ and $\bQ_{s,i}^{(t)}$, and the convergence is at a linear rate. Performing a large number of consensus iterations during the initial orthogonal iterations (outer loop) would be of not much consequence given that the quantities being averaged have inherently huge errors. This implies that communication costs between the nodes in the initial iterations of the outer loop can be reduced without major loss to the final result. This idea motivates us to consider an adaptive version of the S-DOT algorithm, wherein the number of consensus iterations per outer loop iteration increase with time. We call this variant \textit{Sample-wise Adaptive Distributed Orthogonal iTeration (SA-DOT)}. For SA-DOT, we define $\bar{T}_c= \left [T_{c,1},T_{c,2},\dots,T_{c,T_o} \right ]$, where $T_o$ is the total number of outer loop iterations and $T_{c,1} < T_{c,2} \ldots < T_{c,T_o}$. In the $t^{th}$ outer iteration of SA-DOT, we employ $T_{c,t}$ averaging consensus at each site. The algorithm flow for S-DOT and SA-DOT is otherwise congruent. We show in our analysis and experiments the utility of this adaptive method.
\subsection{PSA for Feature-wise Partitioned Data}
The other kind of data partition we consider in this paper is feature-wise. In this case, each node $i$ has access to a few features of all the samples available in a data. As described earlier, if the part of the data available at node $i$ is $\bX_i \in \R^{d_i \times n}$ then the whole data matrix is $\bX = \begin{bmatrix} \bX_1^\tT,\ldots,\bX_N^\tT\end{bmatrix}^\tT$. The goal is to find the dominant $r$-dimensional eigenspace of $\bM = \bX\bX^T$ collaboratively such that each node computes the features of the principal eigenspace corresponding to the data features it carries. In other words, a node carrying the data portion $\bX_i \in \R^{d_i \times n}$ will estimate the corresponding part of $\bQ_f = \begin{bmatrix}\bQ_{f,1}^\tT,\ldots, \bQ_{f,N}^\tT \end{bmatrix}^\tT$ such that $\bQ_{f,i} \in \R^{d_i \times k}$. Similar to the sample-wise data partitioned case, we operate under the assumption that the eigenvalues of $\bM$ follow the order $\lambda_1 \geq \ldots \lambda_r > \lambda_{r+1}\geq \ldots \lambda_d$.

In order to develop our algorithm we recall that each iteration in the centralized OI has two steps: an update step that computes $\widetilde{\bQ} = \bM\bQ$ followed by a QR orthonormalization step. Taking a closer look at the update step when data is partitioned by features, we have
\begin{align}\nonumber
    \bM\bQ &= \bX\bX^\tT\bQ = \bX\begin{bmatrix} \bX_1^\tT,\ldots,\bX_N^\tT\end{bmatrix}\begin{bmatrix}\bQ_{f,1}\\\vdots \\\bQ_{f,N} \end{bmatrix} \\\label{eq:feature_expln}
    &= \bX \Big(\sum_{i=1}^N\bX_i^\tT\bQ_{f,i}\Big) = \begin{bmatrix}\bX_1\Big(\sum_{i=1}^N\bX_i^\tT\bQ_{f,i}\Big)\\ \vdots \\ \bX_N\Big(\sum_{i=1}^N\bX_i^\tT\bQ_{f,i}\Big) \end{bmatrix}.
\end{align}
This shows that the update step computation can be easily distributed as follows: having access to $\bX_i$ and $\bQ_{f,i}$, each node $i$ computes $\bX_i^\tT\bQ_{f,i}$. This is followed by a round of consensus averaging in the network to get the (approximate) sum $\sum_{i=1}^N\bX_i^\tT\bQ_{f,i}$ at each node followed by computing $\bV_{f,i} = \bX_i\Big(\sum_{i=1}^N\bX_i^\tT\bQ_{f,i}\Big)$ at each node $i$. But the orthonormalization step is not as straightforward as in Algorithm~\ref{alg:cdot} because no node has access to full set of vectors. To tackle this, we use a distributed QR decomposition method proposed in~\cite{strakova2011distributed}. This method again uses the weight matrix $\bW$ and exchanges $\bV_{f,i}$ among the nodes to orthonormalize the eigenvectors without the need for any collation of $\bV_{f,i}$. The use of distributed QR evades the necessity of computing the eigenvectors sequentially as proposed in~\cite{scaglione2008decentralized}. Our solution, called \textit{Feature-wise Distributed Orthogonal iTeration (F-DOT)}, is given in Algorithm~\ref{alg:rdot}.

\begin{algorithm}
\caption{Feature-wise Distributed Orthogonal Iteration}\label{alg:rdot}
\begin{algorithmic}[1]

\State {\textbf{Input:} $\bW$; $\bX_i$, $i = 1,\ldots,N$}
\State{\textbf{Initialize:} Set $t \leftarrow 0$ and $\bQ^{(t)}_{f,i} \leftarrow \bQ_{\text{init}}$, where $\bQ_{\text{init}} \in \R^{d \times r}: \bQ_{\text{init}}^\tT\bQ_{\text{init}} = \bI$}

    \While{stopping rule} 

        \State {$t \leftarrow t+1$}
        \State {$\bZ_i^{(t_c)}\leftarrow \bX_i^\tT \bQ_{f,i}^{t-1},i=1,2,\ldots,N$} \label{algostep:doi_f}
        \State {\textbf{Begin consensus loop:} Set $t_c\leftarrow 0$, }
        \While{$t_c < T_c$}
        
            \State{$ t_c \leftarrow t_c+1 $}
            \State{$ \bZ_i^{(t_c)}\leftarrow \sum_{j \in \mathcal{N}_i} w_{i,j}\bZ_j^{(t_c-1)} $} \label{algostep:consensus_f}
            
        \EndWhile
        \State{$\bV_{f,i}^{(t)} \leftarrow \frac{N}{m}\bX_i\frac{\bZ_i^{(t_c)}}{[\bW^{t_c}e_1]_i}$}\label{algostep:consensus_result_f}
        \State{$\bQ_{f,i}^{(t)}, \bR_{f,i}^{(t)} \leftarrow \text{Distributed QR}(\bV_{f,i}^{(t)})$}~\cite{strakova2011distributed} \label{algostep:distQR}

    \EndWhile
\State {\textbf{Return:} $ \bQ_{f,i}^{(t)}$}
\end{algorithmic}
\end{algorithm}

\section{Convergence Analysis and Discussion}\label{sec:analysis}
In the following, we provide a detailed analysis of the convergence behavior of Sample-wise Distributed Orthogonal iTeration (S-DOT) and Sample-wise Adaptive Distributed Orthogonal iTeration (SA-DOT). 
The results need an entity called mixing time of the Markov chain associated with the doubly stochastic matrix $\bW$. It is defined as
\begin{equation}
    \tau_{\text{mix}} = \underset{i=1,\ldots,N}{\text{max}}\underset{t\in \N}{\text{inf}}\left\{t:\|\be_i^\tT\bW^t - \frac{1}{N}\bone^\tT\|_2 \leq \frac{1}{2}\right\},
\end{equation}
where $\bone$ is a vector of ones. We also require the following result from literature~\cite{kempe2008decentralized} that quantifies the convergence behaviour of matrix consensus as a function of the number of consensus iterations.
\begin{proposition}\label{theo:consensus}
\cite[Theorem~5]{kempe2008decentralized} Define $\bZ_i^{(T_c)} \in \R^{d \times r}$ as the matrix at node $i$ after $T_c$ consensus iterations for $i \in \{1, \ldots, N\}$, where the initial value at each site $i$ is $\bZ_i^{(0)}$. Let $\bZ=\sum_{i=1}^{N}\bZ_i^{(0)}$, and define $\bZ^\prime=\sum_{i=1}^{N}\abs{\bZ_i^{(0)}}$, such that the $(j,k)$ entry of $\bZ^\prime$ is the sum of absolute values of the $(j,k)^{th}$ entry of $\bZ_i^{(0)}$ at all nodes $i$. For any $\delta>0$, and $T_c=O(\tau_{\text{mix}}\log{\delta^{-1}})$, the approximation error of averaging consensus is $\norm{\frac{\bZ_i^{(T_c)}}{[\bW^{T_c}\be_1]_i}-\bZ}_F \leq \delta \norm{\bZ^\prime}_F$, $\forall i$.
\end{proposition}
The main theorem of this paper is based on an induction argument, which utilizes the following theorem.
\begin{lemma}\label{lemm:c-dot}
Let $\bM_i, i=1,\ldots, N,$ be the covariance matrix available at node $i$, and define $\bM := \sum\limits_{i=1}^N\bM_i$. Suppose we are at $(t_o+1)^{th}\leq T_o$ iteration of either S-DOT or SA-DOT, where $T_o$ is the maximum number of iterations. Next, define:
\begin{itemize}
    \item $\bQ_c$ to be the eigenspace estimate computed by centralized OI after $t_o$ iterations and $\bQ_{s,i}$ to be the estimate computed after $t_o$ iterations at node $i$ by either S-DOT or SA-DOT,
    \item $\bQ_c^\prime $ and $\bQ_{s,i}^\prime $ to be the eigenspace estimates from OI and S-DOT / SA-DOT after $(t_o+1)$ orthogonal iterations, respectively,
    \item $\bK_c^{(t_o)} \coloneqq \bV_c^{(t_o)^\tT}\bV_c^{(t_o)}=\bR_c^{(t_o)^\tT}\bR_c^{(t_o)}$, where $\bR_c^{(t_o)}$ is the Cholesky decomposition of $\bK_c^{(t_o)}$, and $\bV_c^{(t_o)}=\bM\bQ_c^{(t_o)} = \bM\bQ_c$, and
    \item the constants $\alpha \coloneqq \sum\limits_{i=1}^N \norm{\bM_i}_2$, $\gamma \coloneqq \sqrt{\sum\limits_{i=1}^N \norm{\bM_i}^2_2}$, and $\beta := \underset{t_o=1, \ldots, T_o}{\max} \norm{\bR_c^{-1^{(t_o)}}}_2$.
\end{itemize}
Then for any $\epsilon \in (0,1)$ and a fixed $\delta$, if $\forall i$, $i=1,\ldots,N$, we have 
\begin{equation}\label{eq:ineq6}
    \norm{\bQ_c-\bQ_{s,i}}_F+\frac{\delta \gamma \sqrt{N r}}{\alpha} \leq \frac{1}{2\alpha^2\beta^3\sqrt{r}(2\alpha\sqrt{r}+\delta \gamma \sqrt{N r})}
\end{equation}
and 
\begin{equation}
    T_c=\cO(\tau_{\text{mix}}\log \delta^{-1}),
\end{equation}
then the following is true:
\begin{equation}
     \small \norm{\bQ_c^\prime-\bQ_{s,i}^\prime}_F \leq (3\alpha\beta\sqrt{r})^4\left(\max_i \norm{\bQ_c-\bQ_{s,i}}_F+\frac{\delta\gamma\sqrt{N r}}{\alpha}\right),
\end{equation}
where the parameter $\delta$ is given as:
\begin{itemize}
    \item $\delta = \frac{\alpha}{\gamma\sqrt{N r}}\epsilon^{T_o}\left(\frac{1}{3\sqrt{r}\alpha\beta}\right)^{4T_o}$ for S-DOT, and
    \item $\delta = \frac{\alpha}{T_o\gamma\sqrt{N r}}\epsilon^{T_o}\left(\frac{1}{3\sqrt{r}\alpha\beta}\right)^{4t_o}$ for SA-DOT.
\end{itemize}
\end{lemma}
The proof of Lemma~\ref{lemm:c-dot} is provided in Appendix~\ref{app:lemma1n2}. This lemma states that if the difference between the estimate of the eigenspace obtained using the S-DOT / SA-DOT algorithm and that using the centralized OI is bounded at the beginning of an iteration, then it remains bounded at the end of the iteration too. Notice that the inequality~\eqref{eq:ineq6} is trivially true if the centralized OI and S-DOT / SA-DOT are initialized at the same set of basis vectors. By induction, \eqref{eq:ineq6} and hence the Lemma holds true for every subsequent iteration.

With this lemma in hand, we state our main theorem that guarantees linear convergence of the proposed S-DOT and SA-DOT algorithms.
\begin{theorem}\label{theo:c-dot}
  Let the eigenvalues of $\bM$ be $\lambda_1,\lambda_2,\ldots ,\lambda_d$ such that $\lambda_1 \geq \ldots \geq\lambda_r > \lambda_{r+1}\geq \ldots\lambda_d$ and the true $r$-dimensional principal eigenspace of $\bM$ be represented by $\bQ$. Assume OI, S-DOT and SA-DOT are all initialized to $\bQ_c^{(0)}=\bQ_{s,i}^{(0)}=\bQ_{\text{init}}$, where $\bQ_{\text{init}}$ is a random $d \times r$ matrix with orthonormal columns, and let $\bQ_{\text{init}}$ be such that it satisfies
\begin{equation}
    \abs{\cos{(\theta)}}= \min_{\bu \in \bQ,\bv \in \bQ_{\text{init}}} \frac{\abs{\bu^\tT\bv}}{\norm{\bu}_2\norm{\bv}_2}>0.
\end{equation}
If during the $t^{th}$ S-DOT / SA-DOT iteration, the respective algorithm runs:
\begin{itemize}
    \item $T_c$ consensus iterations in the case of S-DOT with $\small T_c=\Omega \left (T_o\tau_{\text{mix}} \log{(3\sqrt{r}\alpha\beta )}+T_o\tau_{\text{mix}} \log(\frac{1}{\epsilon})+\tau_{\text{mix}}\log{\left(\frac{\gamma \sqrt{N r}}{\alpha}\right)} \right )$,
    \item $T_{c,t}$ consensus iterations for SA-DOT with $\small T_{c,t}=\Omega \left(t\tau_{\text{mix}} \log{(3\sqrt{r}\alpha\beta)}+T_o\tau_{\text{mix}} \log{(\frac{1}{\epsilon})}+\tau_{\text{mix}}\log{\left(T_o\frac{\gamma \sqrt{N r}}{\alpha}\right)}\right)$,
\end{itemize}
where $\epsilon \in (0,1)$ and $\alpha, \beta, \gamma$ are as defined in Lemma~\ref{lemm:c-dot}, then the following is true $\forall i$, $i = 1, \ldots, N$:
\begin{align}\label{eq:2_5}
    \norm{\bQ\bQ^\tT-\bQ_{s,i}^{(T_o)}(\bQ_{s,i}^{(T_o)})^\tT}_2 \leq c\abs{\frac{\lambda_{r+1}}{\lambda_r}}^{T_o}+ c^\prime \epsilon^{T_o},
\end{align}
where $c$ is a positive numerical constant, while $c^\prime = 3$ for S-DOT and $c^\prime = 2$ for SA-DOT.
\end{theorem}
A detailed proof of this theorem, which establishes that $\forall i, \bQ_{s,i} \stackrel{t}{\rightarrow} \pm \bQ$ at a linear rate for both variants of our proposed algorithm, is provided in Appendix~\ref{app:thm1n2}. Note that the first term on the right-hand side of~\eqref{eq:2_5} decays geometrically as a function of the $r^{th}$ eigengap of $\bM$ in accordance with the convergence behaviour of centralized OI, while the second term is the error incurred due to inexact consensus in both S-DOT and SA-DOT. Thus, Theorem~\ref{theo:c-dot} shows that with proper initialization and an adequate fixed number of consensus steps $T_c$ per orthogonal iteration, S-DOT converges at a linear rate to the true $r$-dimensional eigenspace of the global covariance matrix $\bM$. As pointed out earlier, this incurs some unnecessary communication overhead, which may limit the convergence speed of the algorithm. The algorithm SA-DOT improves this communication cost as it adaptively increases the number of consensus iterations $T_{c,t}$ with every orthogonal iteration (notice the $t$ in the definition of $T_{c,t})$.

\subsection{Computation Complexity and Communication Cost}
We now discuss the computation complexity and communication cost of the three algorithms. In the case of sample-wise partitioned data, the local covariance matrices $\bM_i$ are computed only once before the start of the algorithm and hence its computation does not affect the overall complexity of S-DOT and SA-DOT algorithms. The two computationally dominant steps in Algorithm~\ref{alg:cdot} are Steps~\ref{algostep:doi} and~\ref{algostep:doi_qr} requiring $\cO(d^2r)$ and $\cO(r^2d)$ computations per iteration respectively, at every node $i \in \{1,\ldots, N\}$. Since $d \gg r$, Step~\ref{algostep:doi} dominates the overall computational complexity of the algorithm, which is $\cO(d^2rN)$ per iteration for all the $N$ nodes in the network. It is to be noted that Step~\ref{algostep:doi} is an unavoidable step in any OI or power-method based PSA algorithm for sample-wise partitioned data.

In the case of feature-wise partitioned data, the number of operations per iteration in Step~\ref{algostep:doi_f} and Step~\ref{algostep:consensus_result_f} of Algorithm~\ref{alg:rdot} is $O(nd_ir)$ at each node $i$, making the total computational cost of the two steps per iteration $\cO(ndr)$. Furthermore, the computational cost of Step~\ref{algostep:distQR} is $\cO(r^2\log N + \frac{r^2d}{N})$ per iteration. In the case of massive data, $n \gg d$ and hence the computation cost per iteration is dominated by $\cO(ndr)$. Therefore, F-DOT does not work well with data that has large number of samples. In the future we want to develop distributed PSA algorithms that work with big data $\bX$ that has both large $d$ and large $n$.

Now, let us assume that the cost of communicating one $\R^{d\times r}$ matrix in the network is one unit in the case of sample-wise partitioned data. It is clear from Theorem~\ref{theo:c-dot} that for S-DOT, $T_{c}$ is a sum of three terms: the first and second terms are proportional to the maximum number of S-DOT iterations $T_o$ and the third term is proportional to a constant. Also, in the case of SA-DOT it is evident from Theorem~\ref{theo:c-dot} that $T_{c,t}$ is again a sum of three terms: the first term is proportional to the current SA-DOT iteration index $t$, second term is proportional to the maximum number of SA-DOT iterations $T_o$, and the third term is proportional to $\log T_o$. Since $t \leq T_o$, the lower bound of $T_{c,t}$ can be written as $\Omega(T_o)$. It is to be noted from~\eqref{eq:2_5} that $T_o = O(\log{(\frac{1}{\eta})})$ for $O(\eta)$ error. Thus, the lower bound of both $T_c$ and $T_{c,t}$ can be written as $\Omega\left(\log{(\frac{1}{\eta})}\right)$. This implies that the communication complexity for both S-DOT and SA-DOT is $O(T_o T_{c,t}) = O(\log^2 \frac{1}{\eta})$ per node, making the total communication cost $O(N\log^2 \frac{1}{\eta})$.

In the case of feature-wise partitioned data, message exchanges occur in two steps, namely Step~\ref{algostep:consensus_f} and Step~\ref{algostep:distQR}. The size of the message sent from node $i$ in Step~\ref{algostep:consensus_f} is $\R^{n\times r}$, while it is $\R^{d_i\times r}$ in Step~\ref{algostep:distQR}. Let us assume that the cost of communicating one $r$-dimensional vector in the network is one unit. Thus, the communication cost of Step~\ref{algostep:consensus_f} per outer loop iteration is $\cO(n N T_c)$, where $T_c$ is the number of consensus iterations, and that of Step~\ref{algostep:distQR} is $\cO(dNr^2 T_{ps})$, where $T_{ps}$ is the number of push-sum iterations used in distributed QR. It is pointed out in~\cite{strakova2011distributed} that for an $O(\eta)$ error, the number of push-sum iterations in a network of $N$ nodes is $T_{ps} = \cO(\log N + \log \frac{1}{\eta})$. Assuming we use $T_c = O(\log \frac{1}{\eta})$, the total communication cost of F-DOT algorithm will be $\cO(n N \log \frac{1}{\eta} + dNr^2 \log N + dNr^2\log \frac{1}{\eta})$, which is linear in the number of samples $n$ and the total dimension $d$ of the data.


\section{Experimental Results}\label{sec:results}
In this section, we demonstrate the convergence behavior of S-DOT, SA-DOT and F-DOT algorithms through numerical experiments. We generate an undirected connected network having $N$ nodes for each experiment with three different topologies, viz., Erdős–Rényi, ring and star. If not specified, the network topology would be Erd\H{o}s-Rényi with network connectivity parameter $p$. The weight matrix $\bW$ used during the consensus iterations is designed by using the local-degree weights method described in~\cite{xiao2004fast}. The maximum number of consensus iterations is set to $50$, unless otherwise specified. We also emulate real-world distributed synchronous networks using MPI-based blocking point-to-point communications and use that to calculate the number of point-to-point (P2P) communications between different nodes of the network. Since our experiments were carried out using Python on a distributed cluster, we used the MPI for Python package~\cite{mpiforpython} as a wrapper around the Open MPI v2.1.1 implementation of the MPI standard. The Open MPI implementation~\cite{openmpi}, in the case one has both an IP network and at least one high-speed network (such as InfiniBand), automatically switches from TCP/IP to the higher-speed connection. The cluster we utilized, the Amarel cluster of Rutgers, uses the Mellanox InfiniBand fabric. The columns labeled ``P2P'' in all tables in this section stand for the average number of point-to-point communications per node for an experiment using MPI, which is calculated using~\cite{mattaway2000point}. 

The default number of iterations for S-DOT, SA-DOT and F-DOT is $200$ in these tables and $(K)$ represents $1000's$ of P2P communications. Furthermore, the P2P values for the central node and peripheral nodes are marked separately for a star network. The quantity $\Delta_r = \abs{\frac{\lambda_{r+1}}{\lambda_{r}}}$ corresponds to the $r^{th}$ eigengap of the global covariance matrix $\bM$. If $\widehat{\bQ} \in \R^{d\times r}$ is an estimate of the eigenspace and the true low-rank principal subspace is given by $\bQ$ then the error metric used is the average of square of the sine of the principal angles between $\widehat{\bQ}$ and $\bQ$, given as
\begin{equation}
    E = \frac{1}{r}\sum_{i=1}^{r}(1-\sigma_i^2(\bQ^\tT\widehat{\bQ})),
\end{equation}
where $\sigma_i(\bQ^\tT\widehat{\bQ})$ denotes the $i^{th}$ singular value of $\bQ^\tT\widehat{\bQ}$, which gives the cosine of the $i^{th}$ principal angle. The squared-sine distance is simply the chordal distance~\cite{schubert.2016}, which is equivalent to the distance between the projection matrices of $\bQ$ and $\widehat{\bQ}$ quantified in Theorem~\ref{theo:c-dot}.

\subsection{Experiments Using Synthetic Data}
In every experiment with synthetic data, samples were generated such that each site $i$ has $n_i=500$ data points in $\R^{20}$, i.e., $d=20$. Samples are randomly generated from the Gaussian distribution with different $r^{th}$ eigengaps $\Delta_r = \frac{\lambda_{r+1}}{\lambda_r}$. The number of nodes used in the generated network were $N\in \{10,20\}$ and we did $20$ Monte-Carlo trials for each experiment on synthetic data.

\begin{figure}
        \centering
        \vspace{-10pt}
        \begin{subfigure}{0.25\textwidth}
                \includegraphics[width=\linewidth]{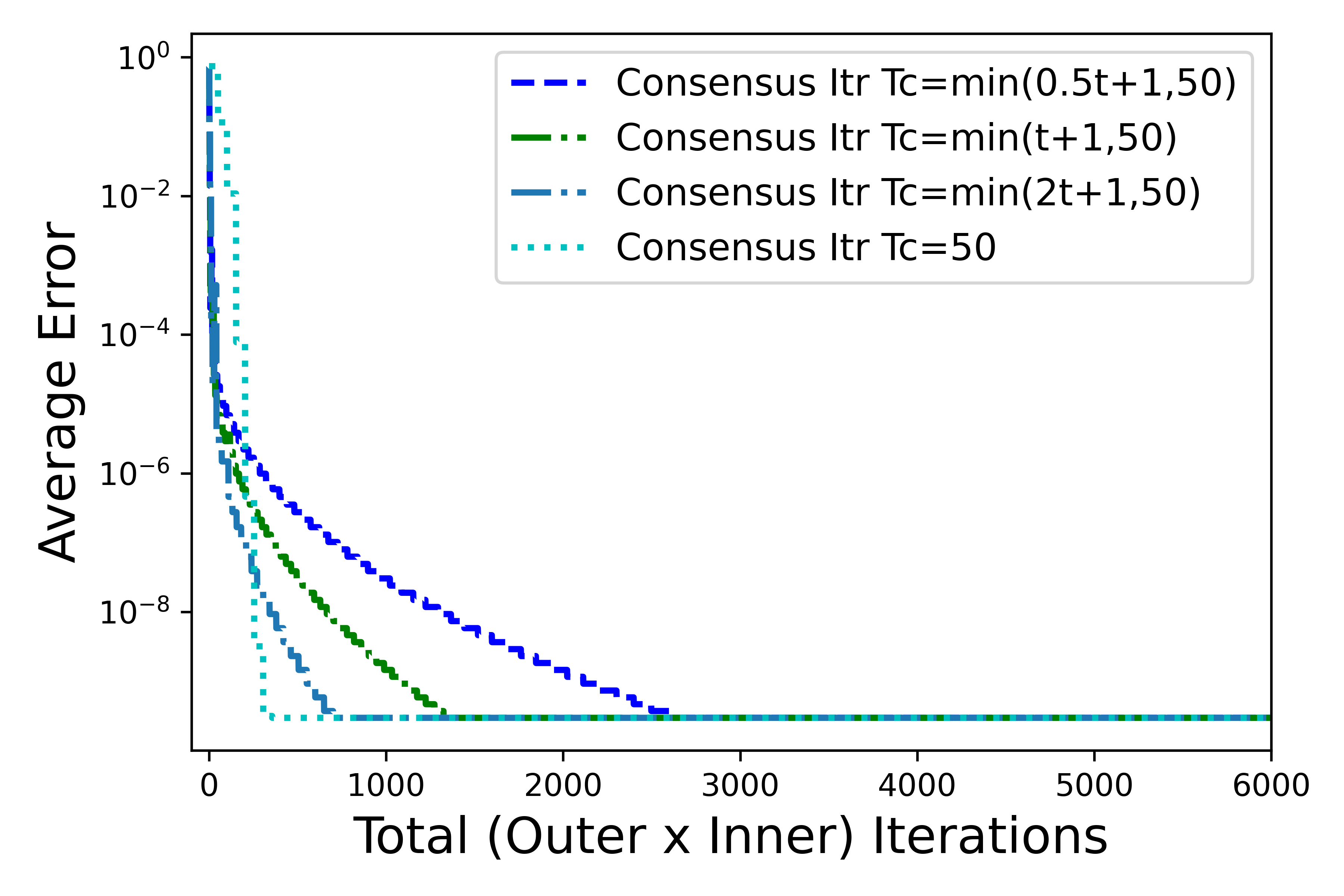}
                \caption{$\Delta_r=0.3$}
                \label{fig:cadotgapa}
        \end{subfigure}%
        \begin{subfigure}{0.25\textwidth}
                \includegraphics[width=\linewidth]{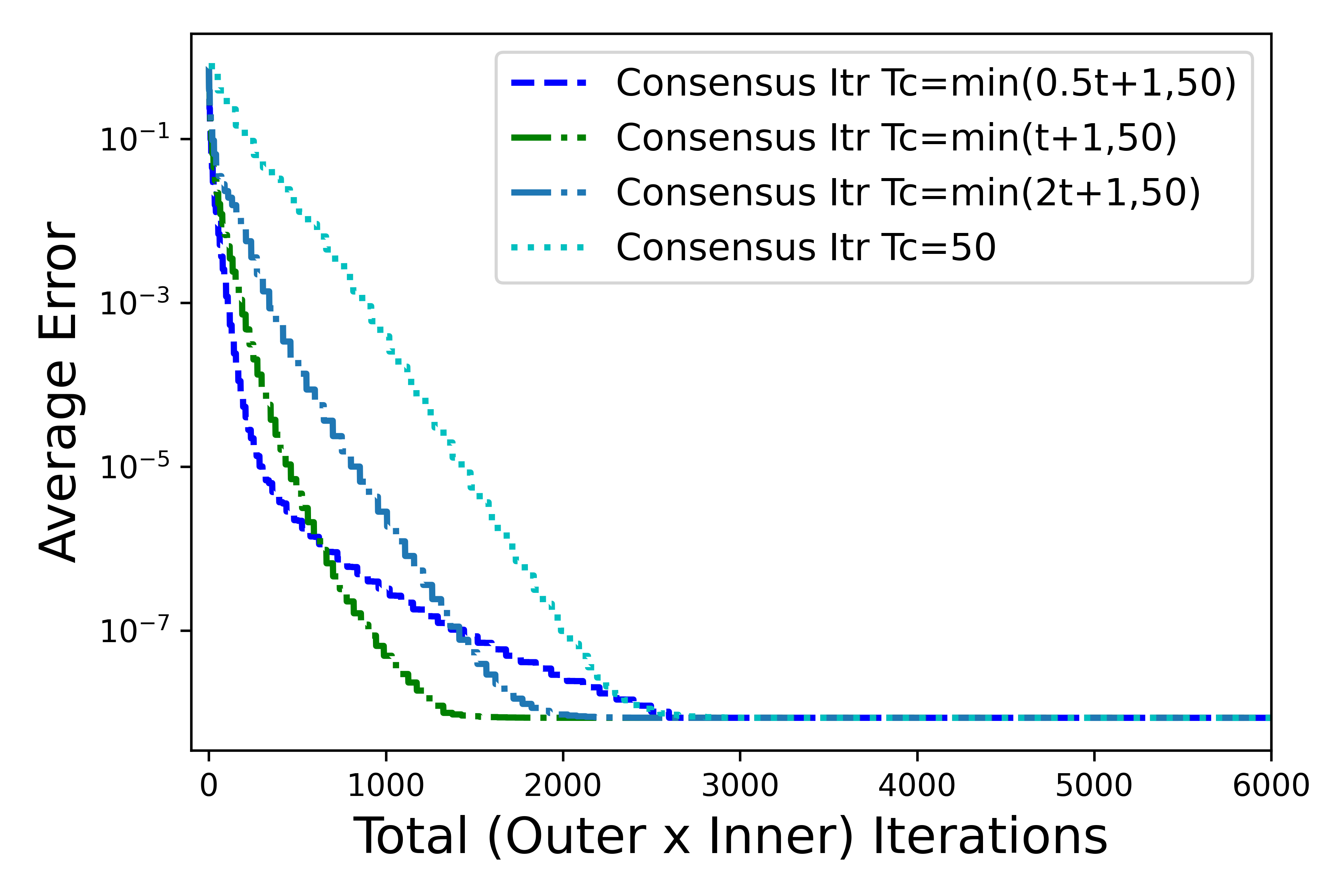}
                \caption{$\Delta_r=0.9$}
                \label{fig:cadotgapc}
        \end{subfigure}%
        \caption{Comparison of S-DOT and SA-DOT for different eigengaps in terms of average error.}\label{fig:cadot_gap}
\end{figure}

First, we show a comparison between the two variants of the proposed algorithm, S-DOT and SA-DOT for sample-wise partitioned data. Specifically we show the effects of using varying number of consensus iterations (in the case of SA-DOT) versus a fixed number of consensus iterations (in the case of S-DOT) per orthogonal iteration in terms of the average number of point-to-point communications (P2P) per node. Table~\ref{Tab:S2} lists P2P communications in the case of different $\Delta_r$ for fixed $T_c=50$ consensus iterations for S-DOT and varying iteration rules for SA-DOT. It is clear from the table that using lesser number of consensus iterations in the beginning can significantly reduce the communication cost. To further depict the effect of different consensus iteration rules on convergence results, Figure~\ref{fig:cadot_gap} provides a comparison for two different eigengaps. The plots show how average error across the nodes changes with the total number of iterations in the network. In accordance with our theoretical results, for a larger eigengap the convergence rate of orthogonal iterations is slower and hence initial iterations have larger errors, which implies having smaller number of communications initially is indeed overall cost effective.
\begin{table}
\caption{Comparison of P2P communications for S-DOT and SA-DOT for different eigengaps}
\centering
\begin{tabular}{|l|l|l|l|l|l|}
\hline
 $N$ & Erdős–Rényi: $p$ & $r$ & $\Delta_r$ & Consensus Itr  $T_c$ & P2P $(K)$\\
\hline
\hline
 20  &  0.25 & 5 & 0.3 & $\lceil 0.5t+1 \rceil$  & 34.88   \\ 
     &      &   &      & $t+1$     & 40.54   \\ 
     &      &   &      & $2t+1$    & 43.31   \\ 
     &      &   &      & 50      & 46.2    \\ 
\hline              
 20  &  0.25 & 5 & 0.7 &$\lceil 0.5t+1 \rceil$  & 37.37  \\ 
     &      &   &      & $t+1$      & 43.44  \\ 
     &      &   &      & $2t+1$     & 46.41  \\ 
     &      &   &      & 50      & 49.5   \\ 
\hline            
 20  &  0.25 & 5 & 0.9 & $\lceil 0.5t+1 \rceil$ & 36.47    \\ 
     &      &   &      & $t+1$     & 42.38    \\ 
     &      &   &      & $2t+1$    & 52.28    \\  
     &      &   &      & 50      & 48.3     \\ 
 \hline
\end{tabular}
  \label{Tab:S2}
\end{table}

\begin{figure}
        \centering
        \vspace{-10pt}
        \begin{subfigure}{0.25\textwidth}
                \includegraphics[width=\linewidth]{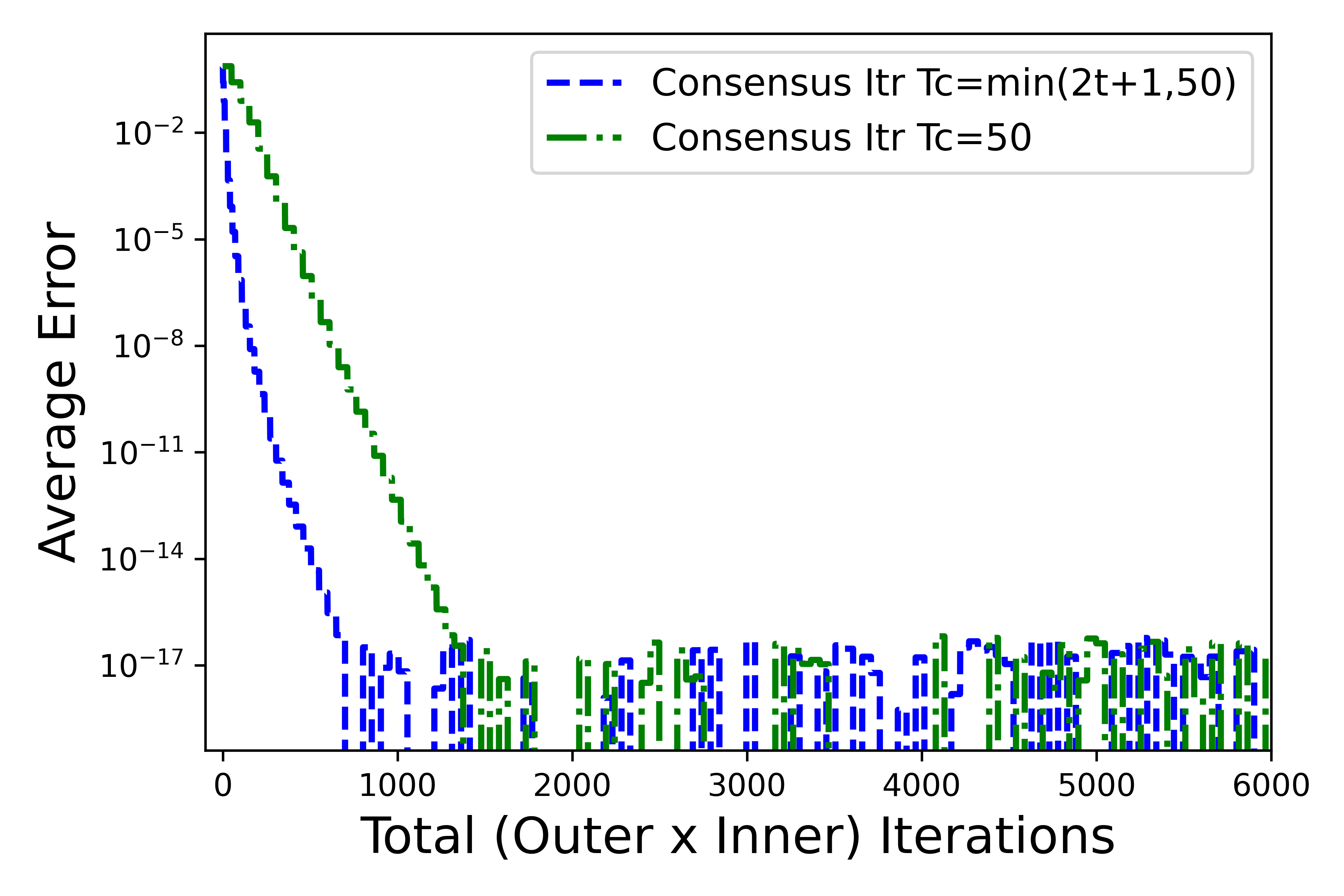}
                \caption{$p=0.5$}
                \label{fig:cadotpa}
        \end{subfigure}%
        \begin{subfigure}{0.25\textwidth}
                \includegraphics[width=\linewidth]{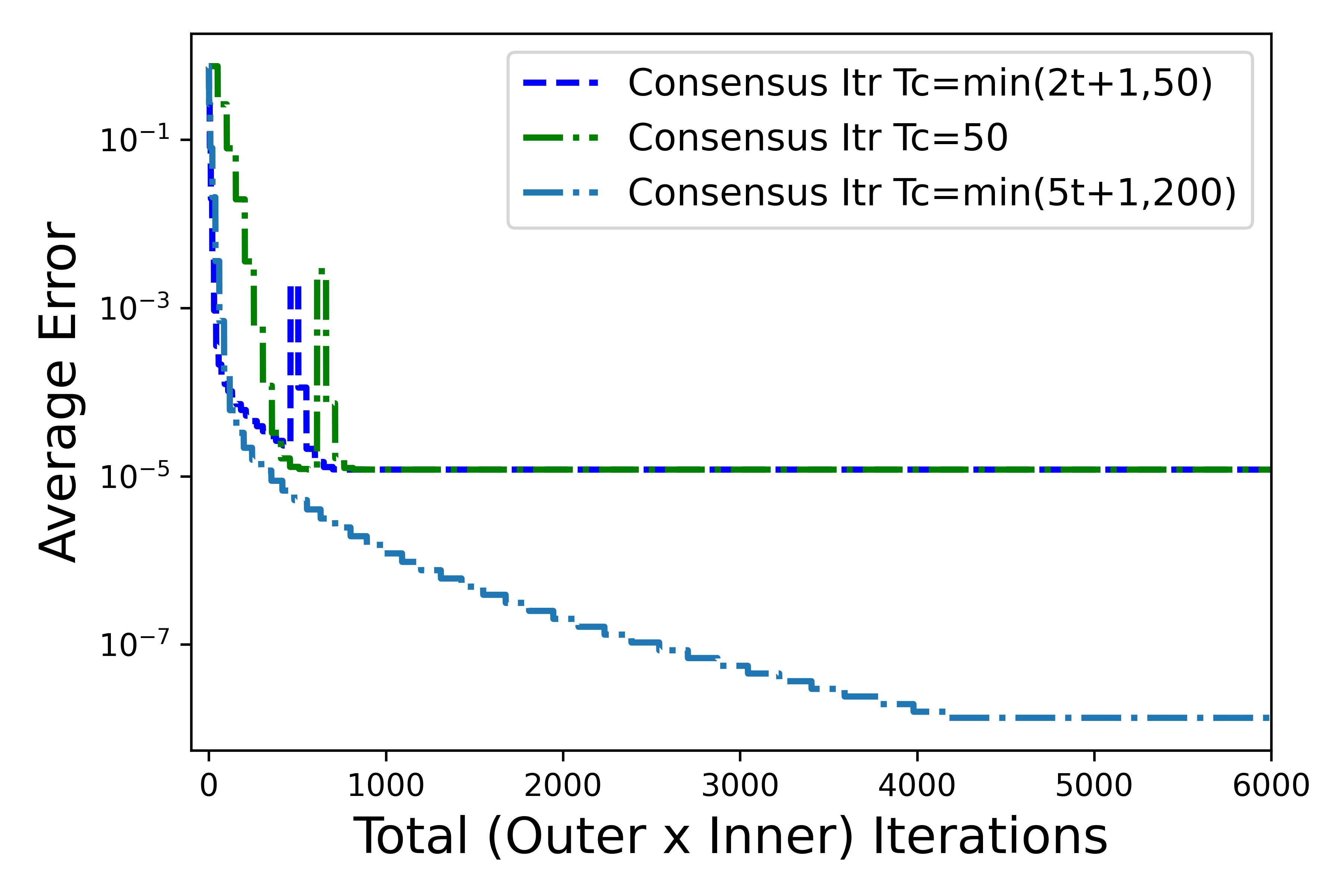}
                \caption{$p=0.1$}
                \label{fig:cadotpc}
        \end{subfigure}%
        \caption{Effect of network connectivity on algorithm performance for sample-wise partitioned data.}\label{fig:cadot_p}
\end{figure}

We also investigate the effect of network connectivity on convergence of the two variants of our proposed algorithm S-DOT and SA-DOT. For this we simulate Erd\H{o}s-Rènyi network topology with different values of connectivity parameter $p$. From the P2P column in Table \ref{Tab:S3}, we can conclude that the number of point-to-point communication increases as $p$ increases. Also, different $p$ leads to different mixing time $\tau_{mix}$ for the corresponding weight matrix $\bW$ for the underlying network, which can also affect the error floor, as indicated in Theorem \ref{theo:c-dot}. Results in Fig. \ref{fig:cadotpc} show that a sparser network can lead to slower convergence. This confirms there is a direct relation between network connectivity and performance of the algorithms. For a sparser network, even though overall communication cost will be lower, but the sparsity hampers information diffusion and hence the final performance of the algorithms.
\begin{table}
\caption{Effect of network connectivity on P2P communications for S-DOT and SA-DOT}
\centering
\begin{tabular}{|l|l|l|l|l|l|}
\hline
 $N$ & Erdős–Rényi: $p$ & $r$ & $\Delta_r$ & Consensus Itr  $T_c$ & P2P $(K)$\\
\hline
\hline
 20  &  0.5 & 5 & 0.7  & $2t+1$    & 90.66   \\ 
     &      &   &      & 50      & 96.7    \\ 
\hline  
 20  &  0.25 & 5 & 0.7 & $2t+1$    & 46.41  \\ 
     &      &   &      & 50      & 49.5   \\ 
\hline            
 20  &  0.1 & 5 & 0.7  & $2t+1$   & 22.97  \\ 
     &      &   &      & 50     & 24.5   \\ 
     &      &   &      & $\min(5t+1, 200)$       & 88.05  \\ 
 \hline
\end{tabular}
  \label{Tab:S3}
\end{table}

\begin{figure}
        \vspace{-10pt}
        \begin{subfigure}[b]{0.25\textwidth}
                \includegraphics[width=\linewidth]{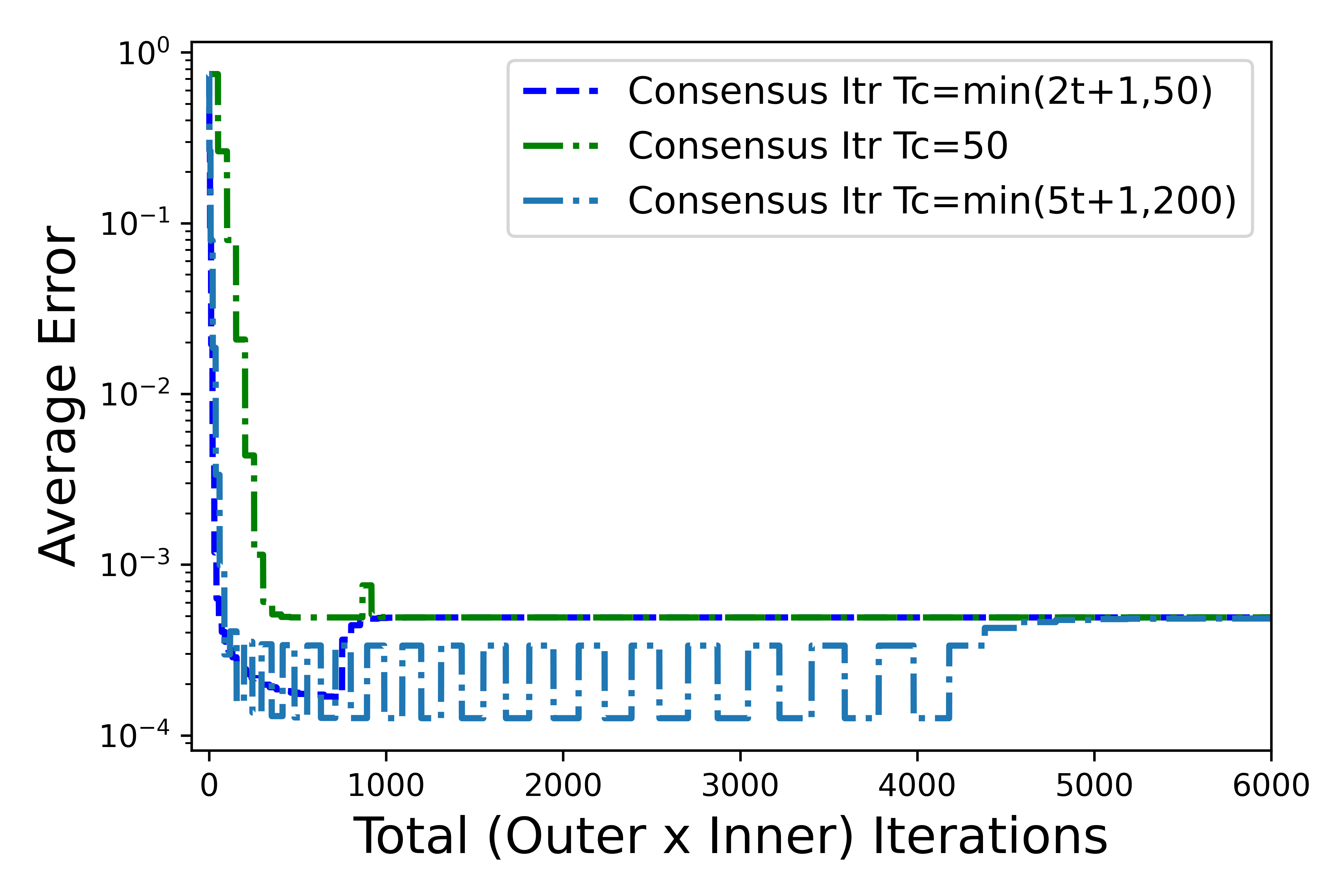}
                \caption{Ring topology}
                \label{fig:cadotringstara}
        \end{subfigure}%
        \begin{subfigure}[b]{0.25\textwidth}
                \includegraphics[width=\linewidth]{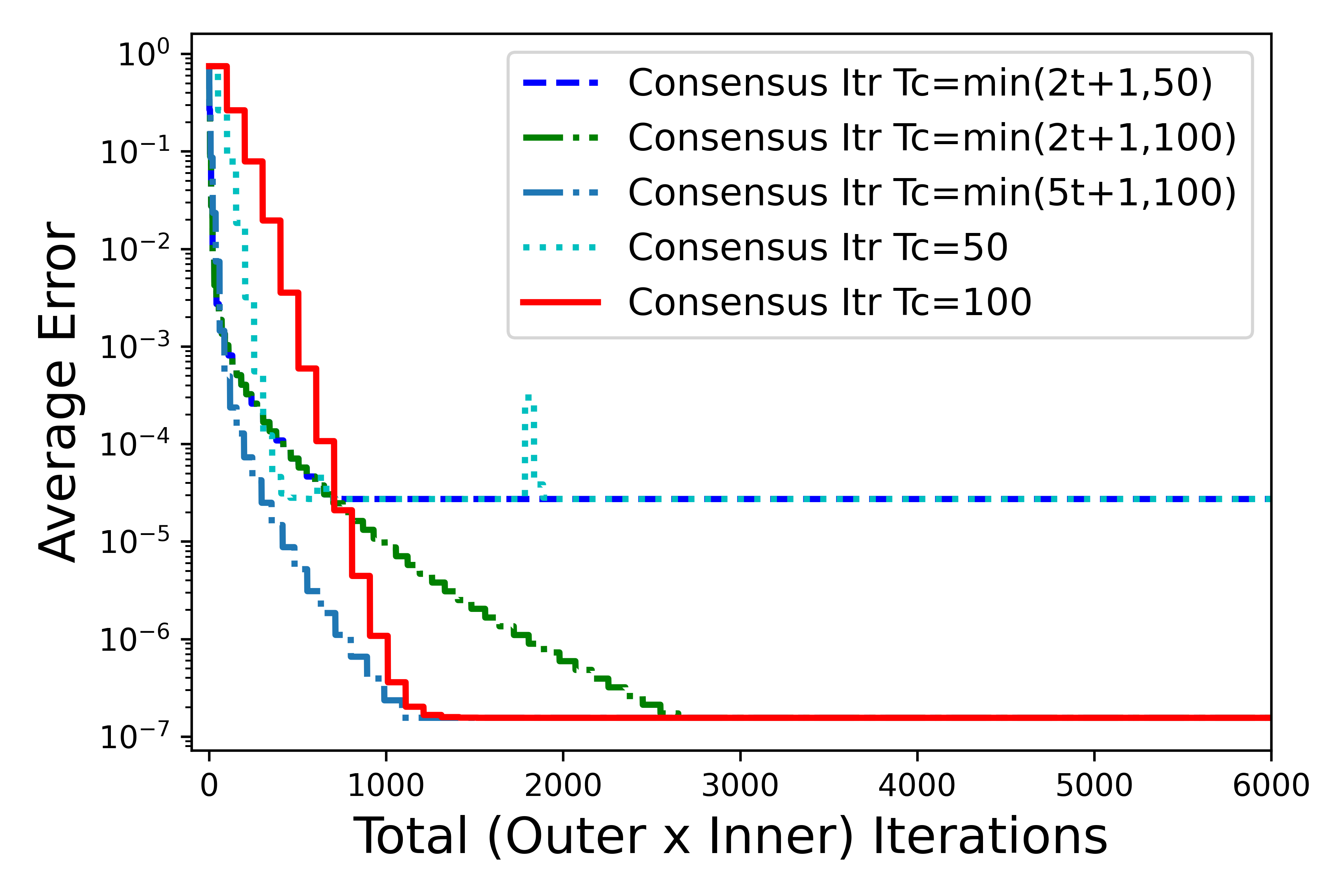}
                \caption{Star topology}
                \label{fig:cadotringstarb}
        \end{subfigure}%
        \caption{Comparison of S-DOT and SA-DOT for ring and star topologies in terms of average error.}\label{fig:cadot_ring_star}
\end{figure}

We also demonstrate the performance of our algorithms on ring and star topologies for sample-wise partitioned data. Table~\ref{Tab:s4} gives the parameter details and P2P communications for a ring network. For star topology, the number of P2P communications are different for the center node and other (edge) nodes. In Table \ref{Tab:s5}, the number of point-to-point communication at the center node is equal to the sum of all edge nodes, which creates a bottleneck effect at the central node that can lead to slow convergence rate for an algorithm. The results for ring topology in Fig. \ref{fig:cadot_ring_star} show that S-DOT and SA-DOT do not perform too well since ring topology is a periodic Markov chain \cite{gagniuc2017markov} that cannot converge to a steady-state distribution. The steady-state distribution exists if the Markov chain with a finite number of states is aperiodic and irreducible, therefore, $\tau_{mix}\rightarrow \infty$ for ring topologies. 
\begin{table}
\caption{Parameters and P2P communication for ring topology}
\centering
\begin{tabular}{|l|l|l|l|l|}
\hline
 $N$ & $r$ & $\Delta_r$ & Consensus Itr   & P2P $(K)$\\
\hline
\hline
 20   & 5 & 0.7  & $2t+1$        & 18.75    \\ 
      &   &      & 50          & 20       \\ 
      &   &      & $\min(5t+1, 200)$   & 71.88  \\  
\hline  
\end{tabular}
  \label{Tab:s4}
\end{table}
\begin{table}
\caption{Parameters and P2P communication for star topology}
\centering
\begin{tabular}{|l|l|l|l|l|l|}
\hline
 $N$ & $r$ & $\Delta_r$ & Consensus Itr   & Center P2P $(K)$ & Edge P2P $(K)$\\
\hline  
\hline
 20   & 5 & 0.7  & $2t+1$            & 178.13   & 9.38   \\  
      &   &      & 50               & 190      & 10     \\  
      &   &      & $\min(2t+1, 100)$        & 332.5   & 17.5   \\  
      &   &      & $\min(5t+1, 100)$          & 360.43   & 18.97  \\  
      &   &      & 100               & 380     & 20      \\ 
\hline 
\end{tabular}
  \label{Tab:s5}
\end{table}

Next, we investigate the effect of straggler nodes in a network on convergence speed. The straggler effect delays the job completion for distributed algorithms because of the presence of a slow node in the network~\cite{straggler.2017}. In this experiment, we emulate the straggler effect by setting a $0.01$ second delay during each iteration at a randomly selected site $i$ that changes every iteration. Since our algorithms are designed for synchronous networks, the impact of a straggler node is significant on S-DOT and SA-DOT, as shown in Table \ref{Tab:s7} for an Erd\H{o}s-Rènyi topology. The execution time of experiments shown in Table \ref{Tab:s7} indicates that a slow node can slow down the job completion for the entire network to a good extent. Speeding up the algorithms in the presence of straggler nodes requires dealing with asynchronicity in the networks and we leave that work for future.
\begin{table}
\caption{Effect of straggler nodes on execution time of S-DOT and SA-DOT}
\centering
\begin{tabular}{|l|l|l|l|l|l|l|l|}
\hline
 $N$ & $p$ & $r$ & $\Delta_r$ & Cons. Itr & Time (in s) & P2P $(K)$ & Straggler\\
\hline
\hline
 10  &  0.5 & 5 & 0.7  & $2t+1$ & 101.33   & 45    & Yes    \\
     &      &   &      & $2t+1$ & 5.18     & 45   &  No   \\
     &      &   &      & 50   & 108.56   & 48   & Yes    \\  
     &      &   &      & 50   & 19.5     & 48   &  No   \\ 
\hline              
 20  & 0.25 & 5 & 0.7  & $2t+1$ & 98.5    & 47.81      & Yes    \\
     &      &   &      & $2t+1$ & 5.08    & 47.81      & No    \\
     &      &   &      & 50   & 105.59   & 51     & Yes   \\
     &      &   &      & 50   & 5.74     & 51     &  No   \\ 
 \hline
\end{tabular}
  \label{Tab:s7}
\end{table}

Having demonstrated the dynamics of our proposed algorithms for sample-wise partitioning with respect to various factors like network connectivity, eigengap, etc., we now show the comparison of our algorithms with other existing work in both centralized and distributed domains. We compare with two centralized methods, orthogonal iteration (OI)~\cite{van1983matrix}, where the whole subspace is estimated at once, and sequential power method (SeqPM), where each basis vector of the $r$-dimensional subspace is estimated sequentially. We also provide comparisons with some distributed algorithms, namely, distributed Sanger's algorithm (DSA), which is a recently proposed Hebbian-based learning algorithm~\cite{gang.bajwa.2021}, distributed projected gradient descent (DPGD), which is a common gradient-based method to solve constrained problems, sequential distributed  power method (SeqDistPM), which is the distributed version of SeqPM, and a recently proposed gradient tracking based subspace estimation method called DeEPCA~\cite{ye2021deepca}. Note that DPGD involves two significant steps per iteration: first is a distributed gradient descent step at every node $i$ that takes the form $\sum_{j\in \cN_i}w_{ij}\bQ_j + \alpha\nabla f_i(\bQ_i)$ as in~\cite{dgd} using trace maximization of the function $f_i(\bQ_i) =  Tr(\bQ_i^\tT\bM_i\bQ_i)$ as the objective function. This is followed by a projection step at each node to ensure the orthogonality constraint $\bQ_i^{\tT}\bQ_i = \bI$, when the orthogonalization is accomplished using QR decomposition. In these set of experiments, the number of nodes in the network $N$ was set to 10, with each node having $n_i = 1000$ samples in $\R^{20}$, i.e., $d=20$. The number of consensus iterations used for S-DOT was 50 and was $\min (t+1,50)$ in the $t^{th}$ iteration of SA-DOT.

\begin{figure}
        \centering
        \vspace{-10pt}
        \begin{subfigure}{0.25\textwidth}
                \includegraphics[width=\linewidth]{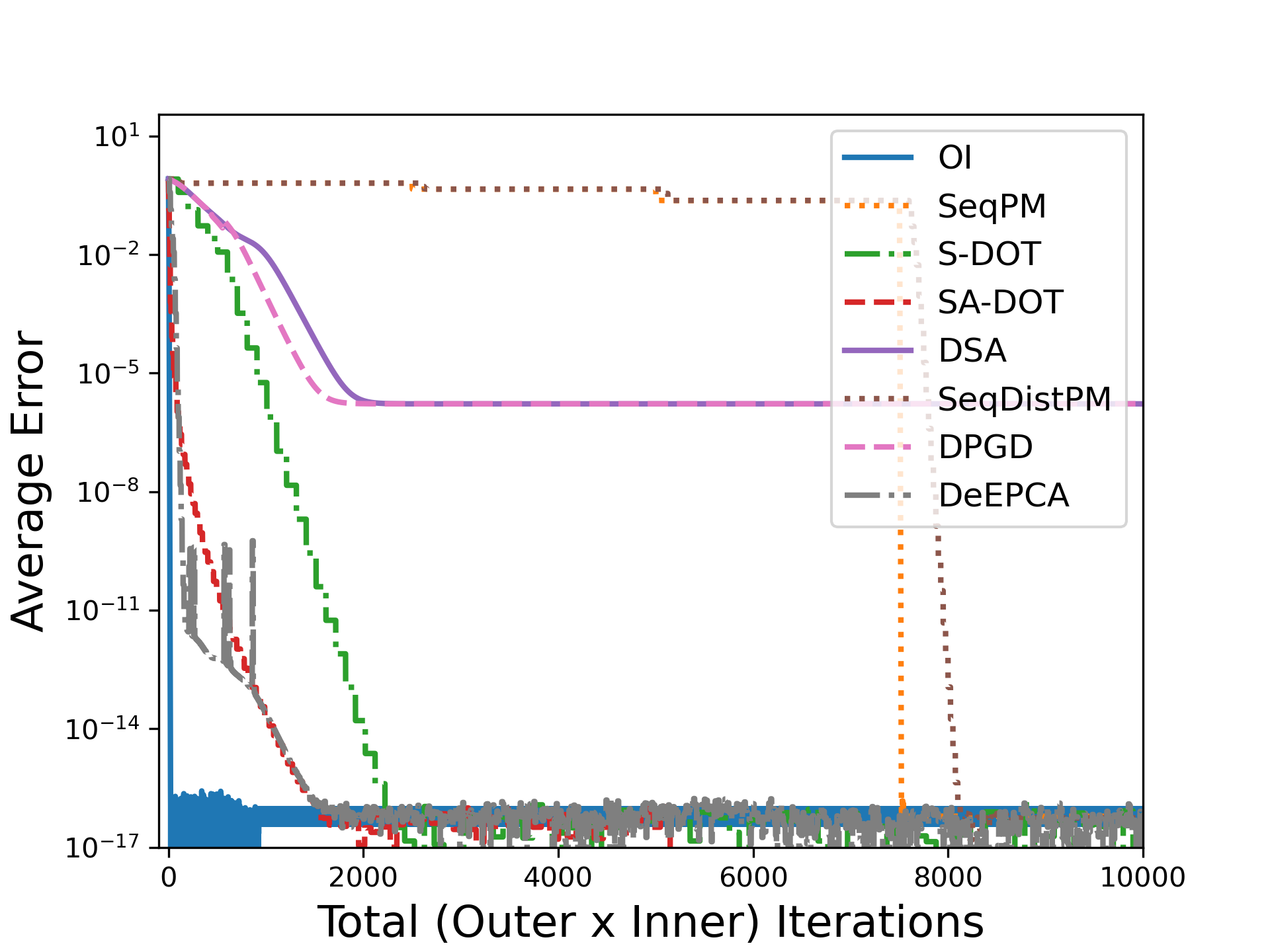}
                \caption{$r=4, \Delta_r = 0.4$}
                \label{fig:c1a}
        \end{subfigure}%
        \begin{subfigure}{0.25\textwidth}
                \includegraphics[width=\linewidth]{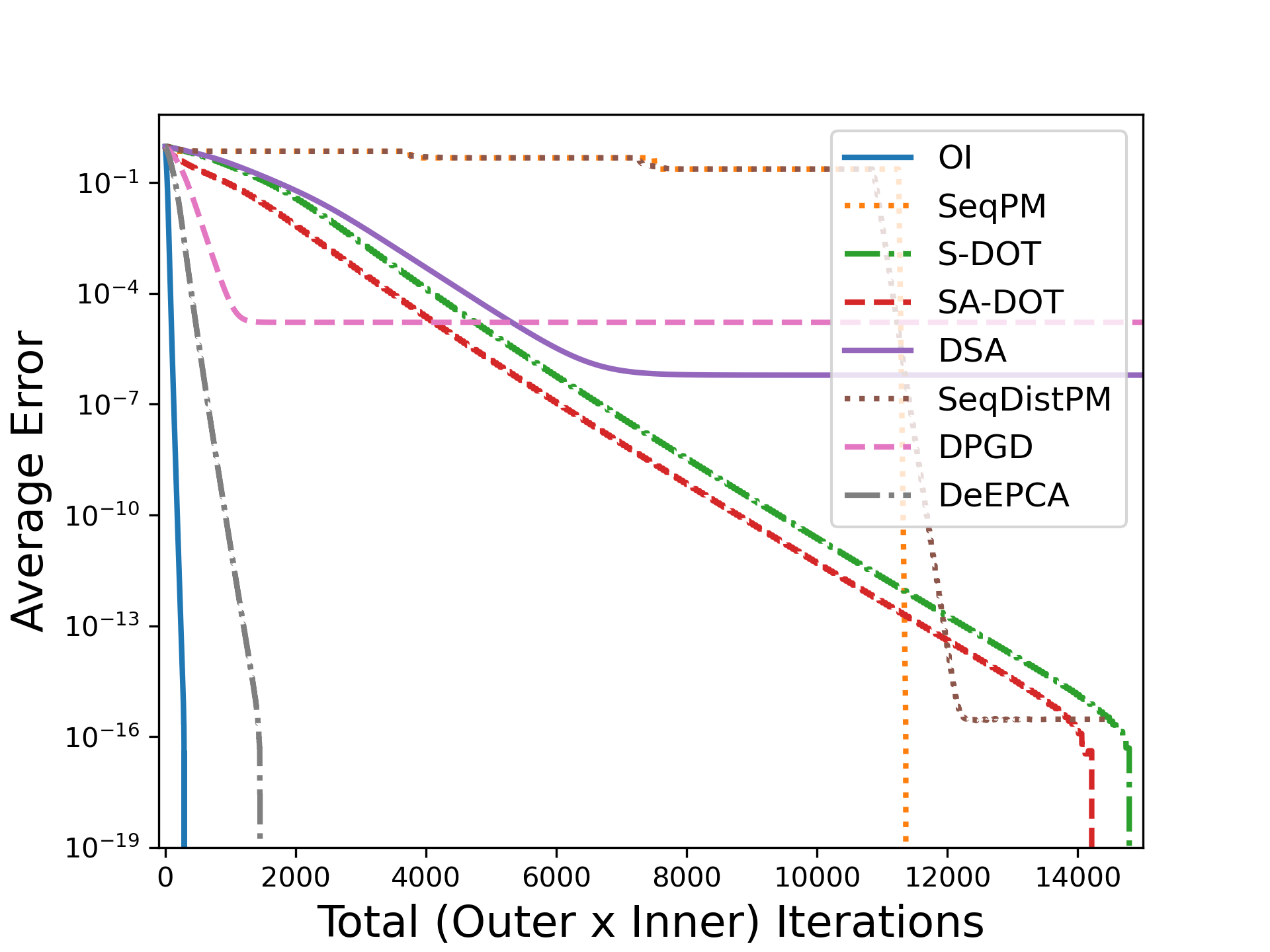}
                \caption{$r=4, \Delta_r = 0.85$}
                \label{fig:c1b}
        \end{subfigure}%
        
        \begin{subfigure}{0.25\textwidth}
                \includegraphics[width=\linewidth]{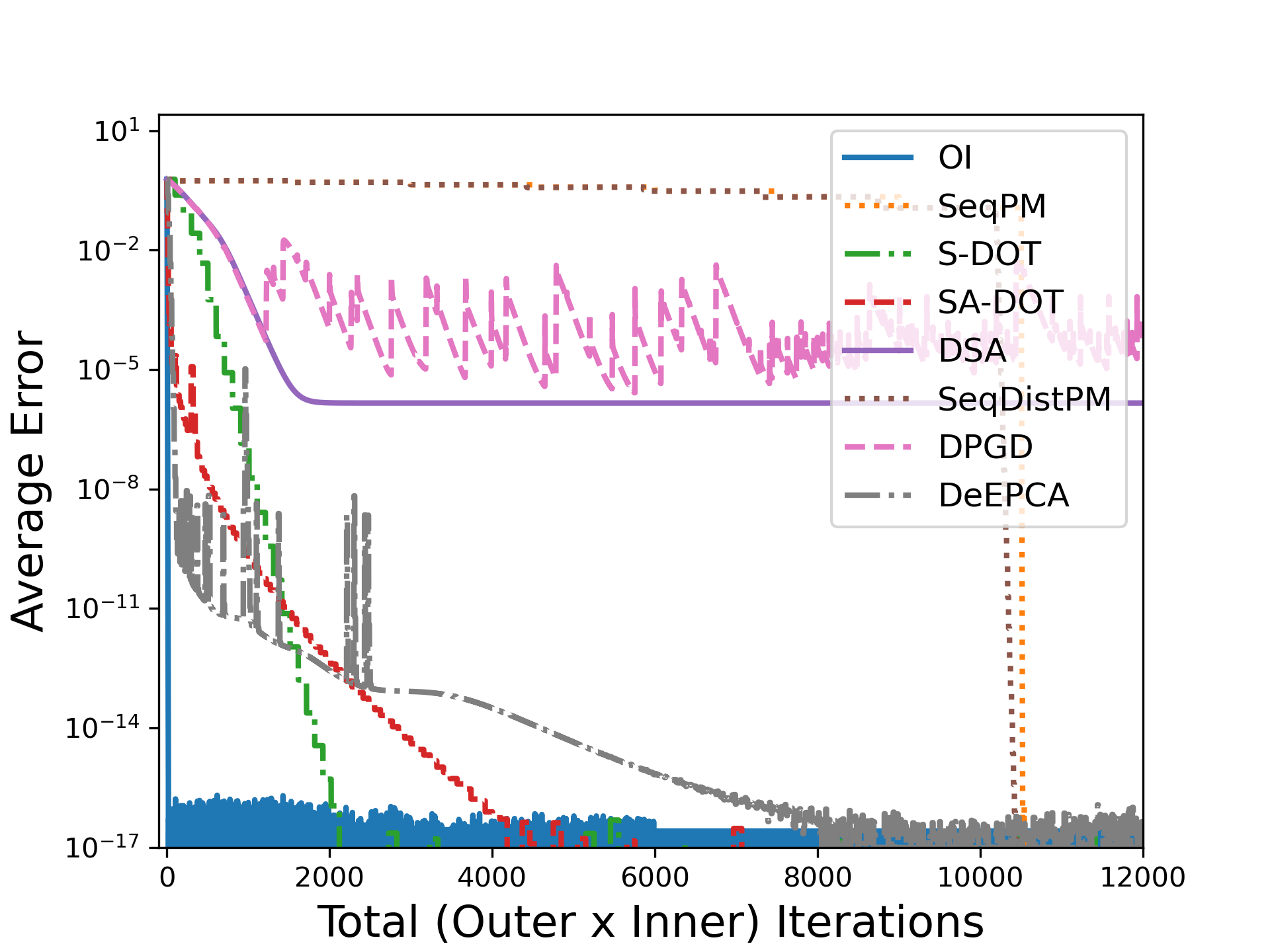}
                \caption{$r=8, \Delta_r = 0.4$}
                \label{fig:c1c}
        \end{subfigure}%
        \begin{subfigure}{0.25\textwidth}
                \includegraphics[width=\linewidth]{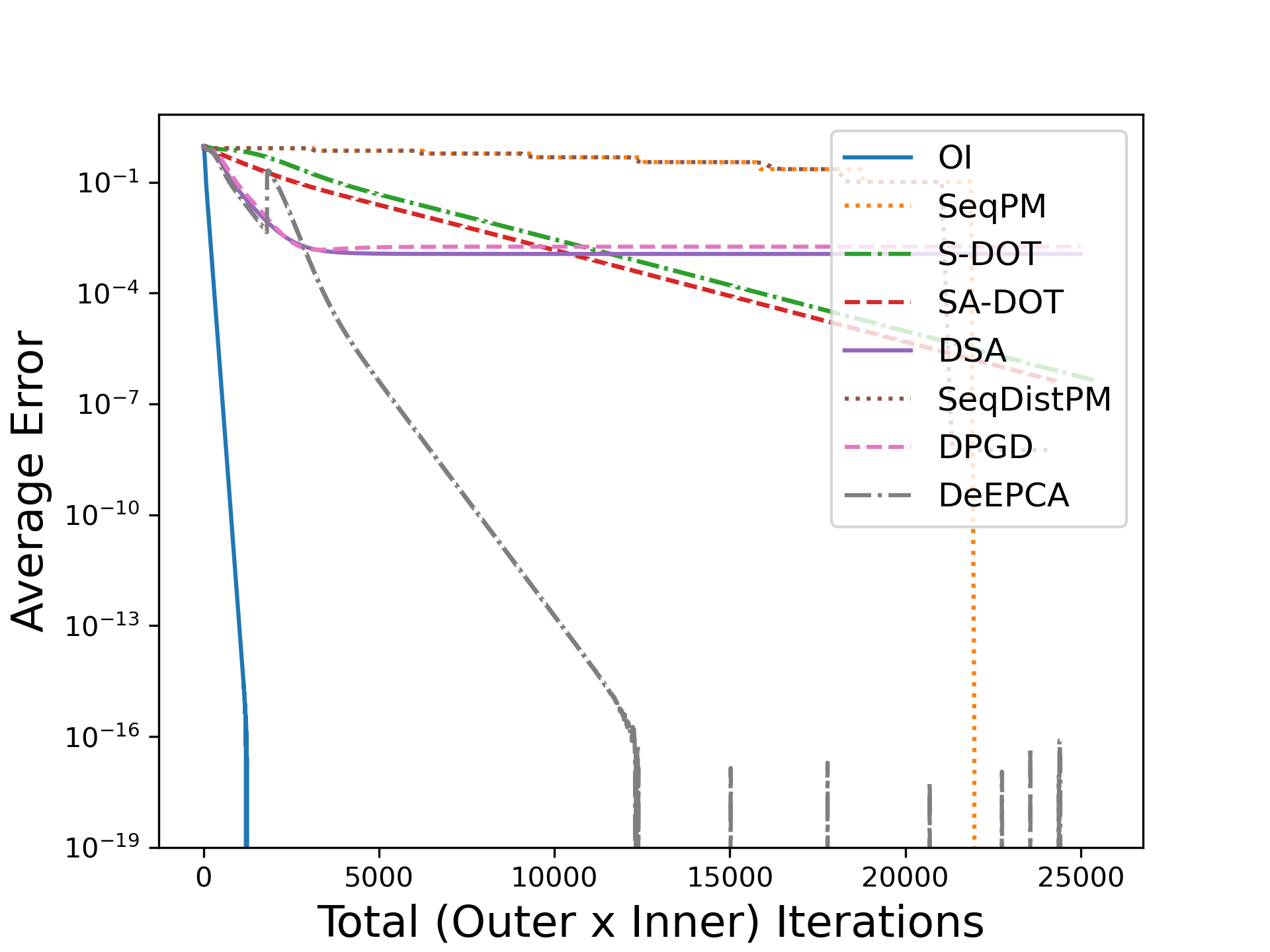}
                \caption{$r=8, \Delta_r = 0.85$}
                \label{fig:c1d}
        \end{subfigure}%
        \caption{Performance comparison of S-DOT and SA-DOT with various centralized and distributed algorithms when all eigenvalues are distinct.}\label{fig:comp1}
\end{figure}

\begin{figure}
        \centering
        \vspace{-10pt}
        \begin{subfigure}{0.25\textwidth}
                \includegraphics[width=\linewidth]{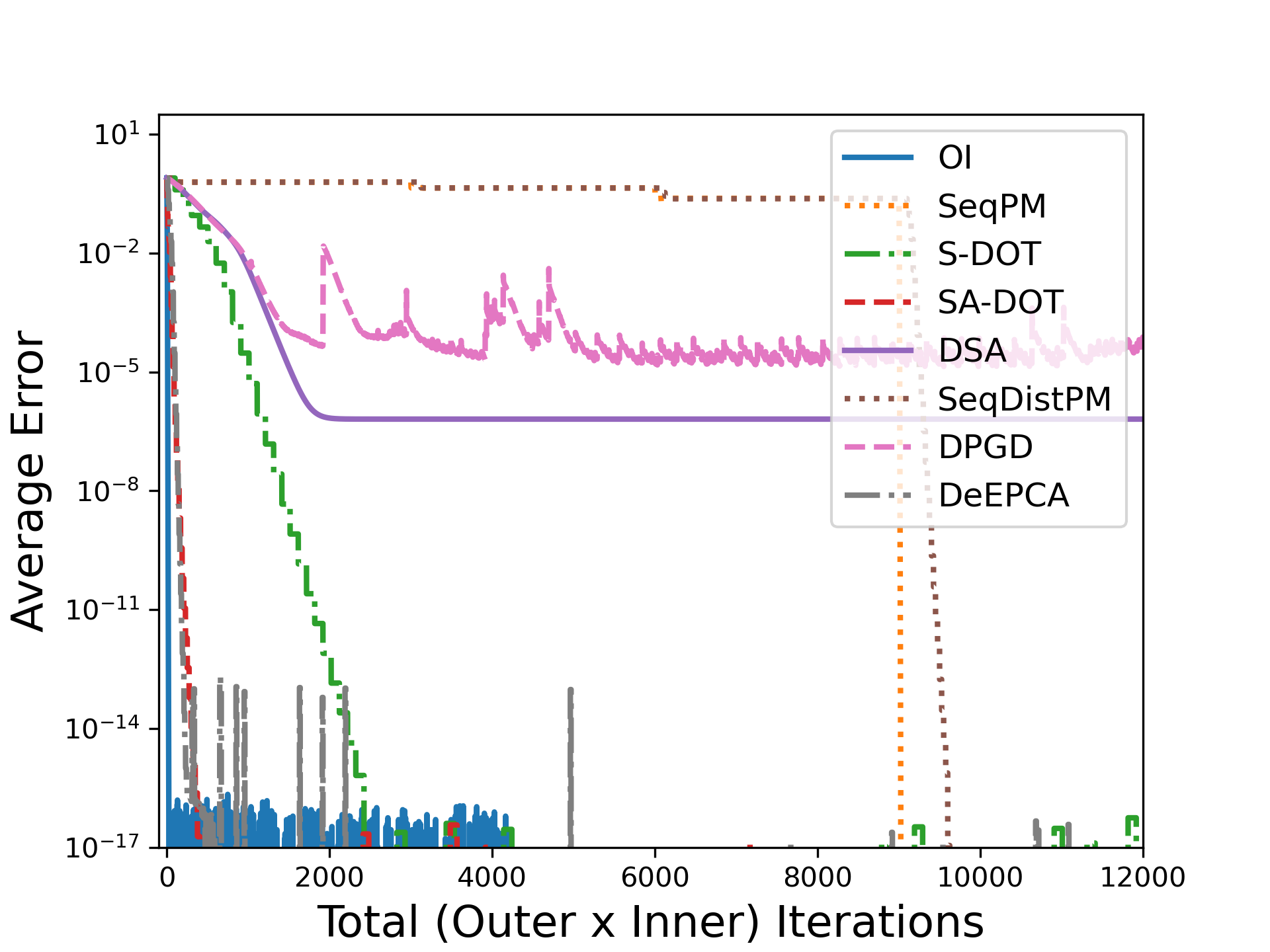}
                \caption{$r=4, \Delta_r = 0.4$}
                \label{fig:c2a}
        \end{subfigure}%
        \begin{subfigure}{0.25\textwidth}
                \includegraphics[width=\linewidth]{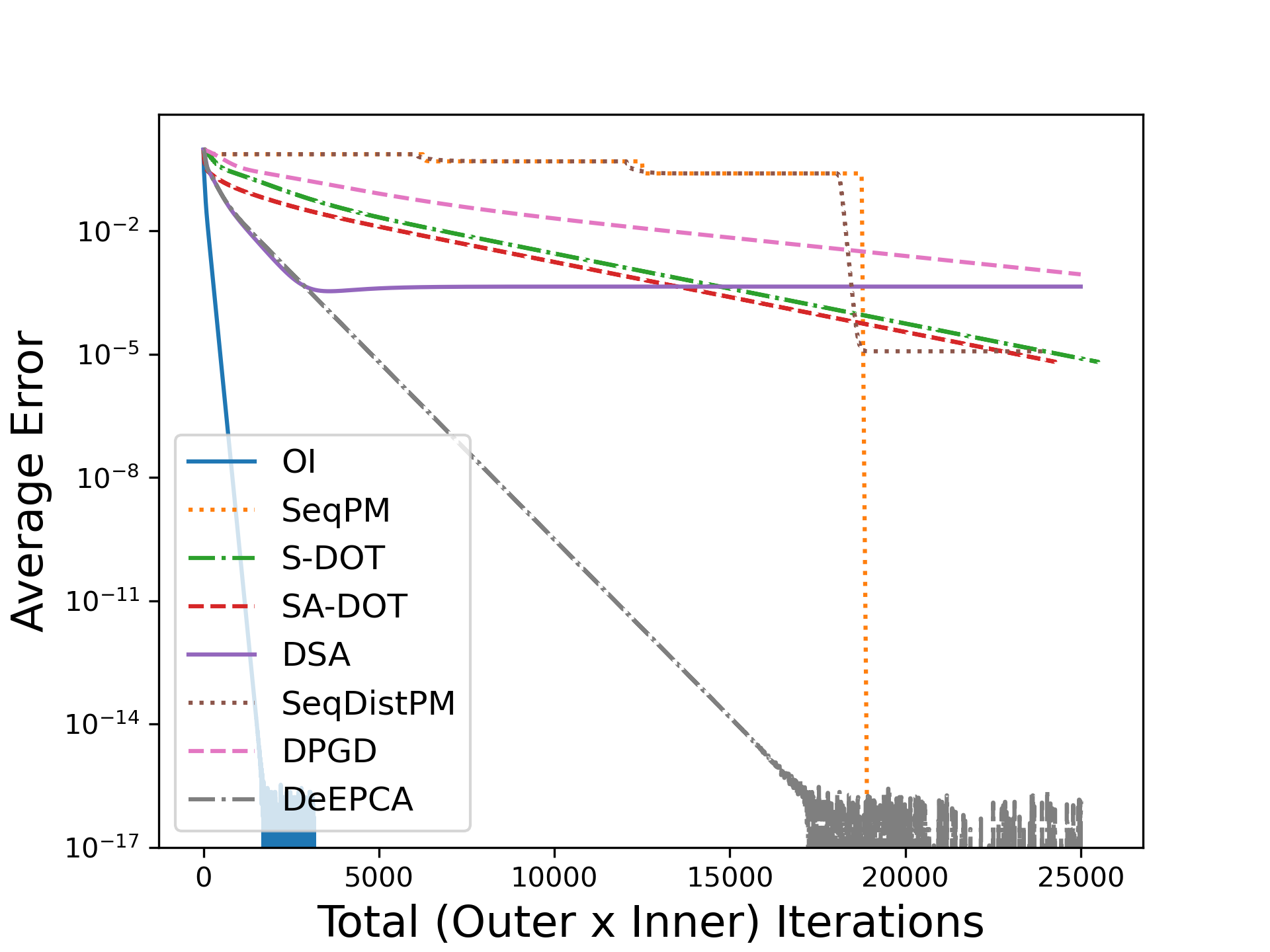}\
                \caption{$r=4, \Delta_r = 0.85$}
                \label{fig:c2b}
        \end{subfigure}%
        
        \begin{subfigure}{0.25\textwidth}
                \includegraphics[width=\linewidth]{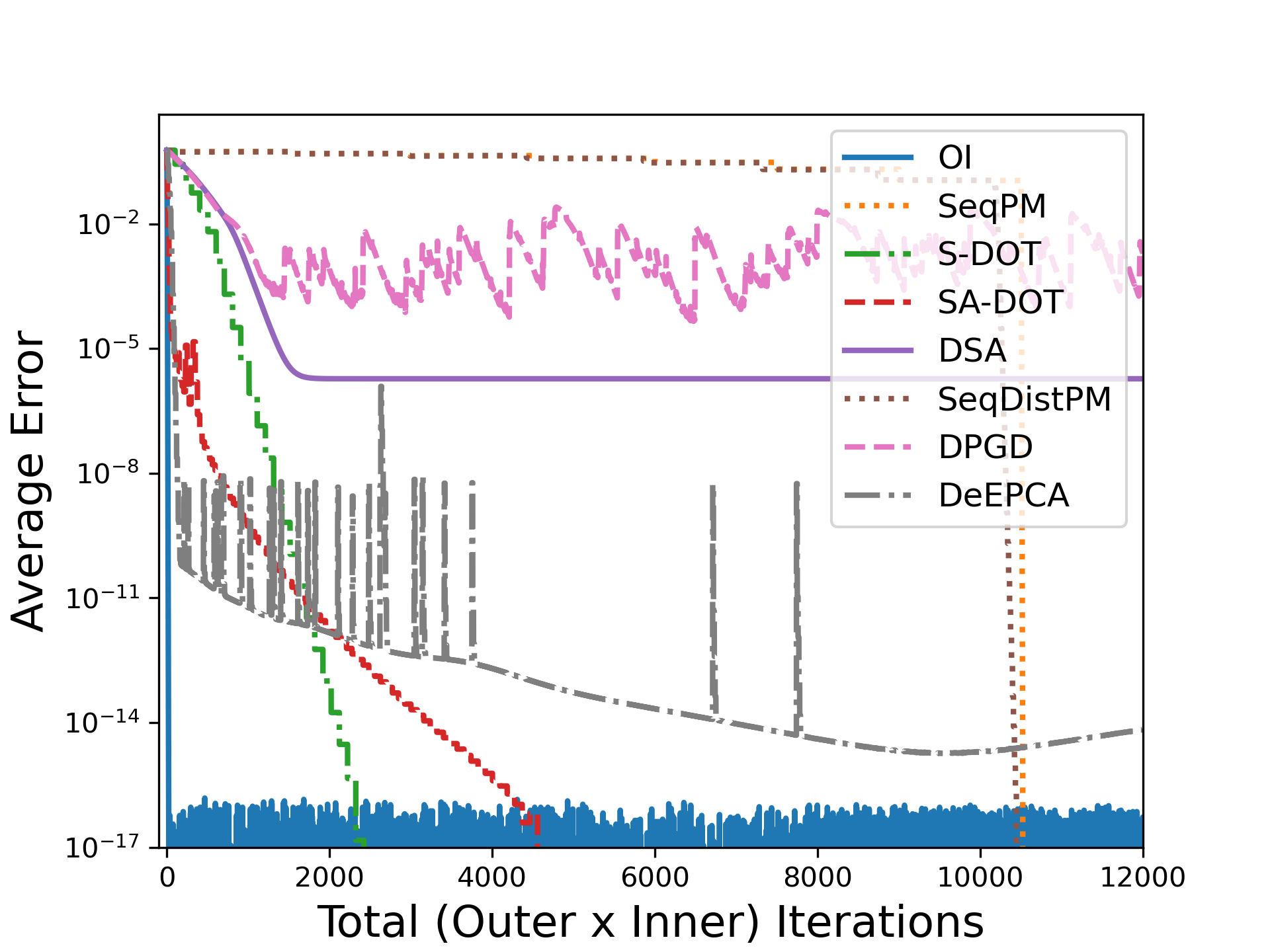}
                \caption{$r=8, \Delta_r = 0.4$}
                \label{fig:c2c}
        \end{subfigure}%
        \begin{subfigure}{0.25\textwidth}
                \includegraphics[width=\linewidth]{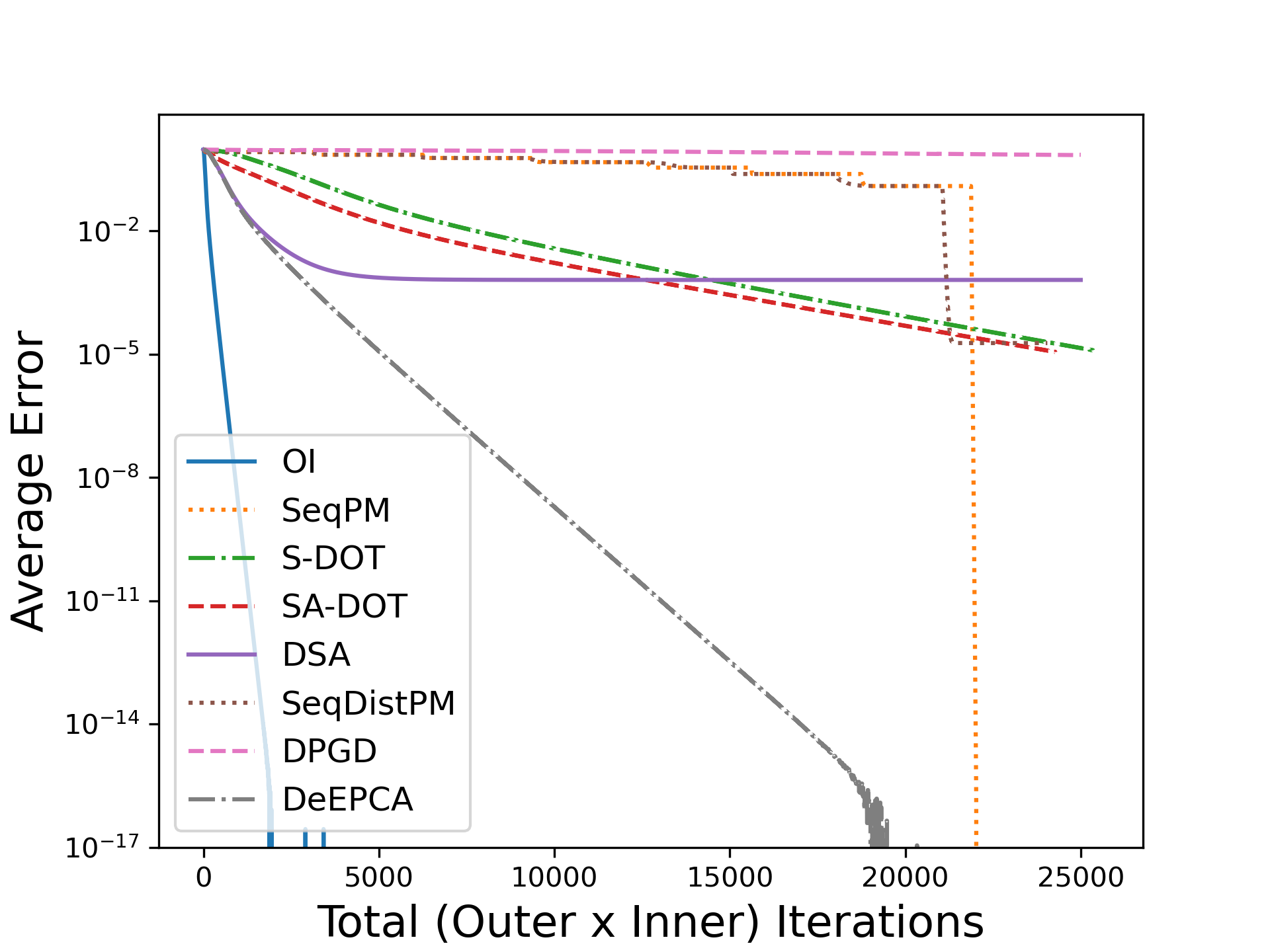}
                \caption{$r=8, \Delta_r = 0.85$}
                \label{fig:c2d}
        \end{subfigure}%
        \caption{Performance comparison of S-DOT and SA-DOT with various centralized and distributed algorithms in the case of non-distinct eigenvalues.}\label{fig:comp2}
\end{figure}

The convergence guarantees for S-DOT and SA-DOT algorithms show that estimation of the space spanned by the top $r$ eigenvectors of the global covariance matrix $\bM$ depends on the $r^{th}$ eigengap $\Delta_r$. Figure~\ref{fig:comp1} shows the comparisons for two different eigengaps and two values of $r$ and all the eigenvalues are distinct. It is clear that for all combinations of $\Delta_r$ and $r$, the proposed methods significantly outperform the sequential power methods (SeqPM, SeqDistPM) in terms of total number of iterations (inner x outer) required to converge. This is because the sequential methods compute one basis vector at a time and since the other lower-order estimates are still at their initial random values, they contribute a large error. It is only when the last basis vector is getting estimated do the errors come down significantly. There are no inner loops in case of OI, SeqPM, DSA and DPGD and hence the number of (outer x inner) loops are same as the number of outer loops. So in all the figures showing comparison with the other methods, the x-axis for OI, SeqPM, DSA and DPGD implies outer loop only while for the other algorithms it implies (outer x inner) loops. The methods DSA and DPGD both only converge to a neighborhood of the true solution and hence have a weaker performance compared to S-DOT and SA-DOT. Both our methods clearly have slightly inferior performance than DeEPCA in terms of total communication cost. This is due to the additional log factor in the total communication cost required by our proposed algorithm as compared to DeEPCA, as discussed in Remark~\ref{remark:remark1}. Next, as asserted by our analysis, S-DOT and SA-DOT only require $\lambda_r$ and $\lambda_{r+1}$ to be distinct. To investigate the effect on convergence when some of the other eigenvalues are equal, we generate data from a distribution such that $\lambda_1=\lambda_2=\ldots=\lambda_r > \lambda_{r+1}$ (note that for finite number of samples, the eigenvalues might not be exactly equal but very close). It is clear from Figure~\ref{fig:comp2} that the performance of our algorithms remains the same and better than the other algorithms in this case too. 

\begin{figure}
        \centering
        \vspace{-10pt}
        \begin{subfigure}{0.25\textwidth}
                \includegraphics[width=\linewidth]{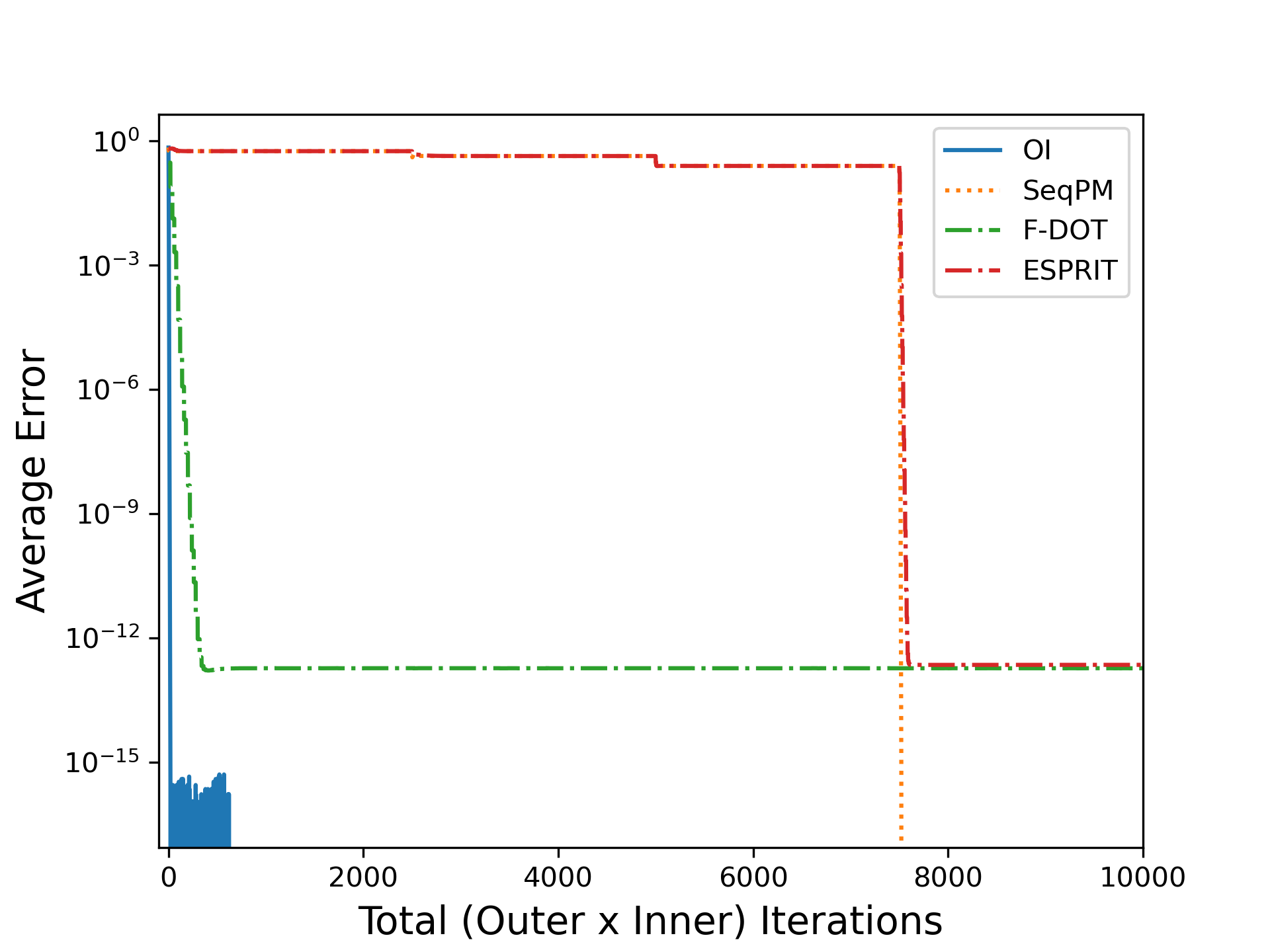}
                \caption{$r = 4, \Delta_r = 0.4$}
                \label{fig:fc1a}
        \end{subfigure}%
        \begin{subfigure}{0.25\textwidth}
                \includegraphics[width=\linewidth]{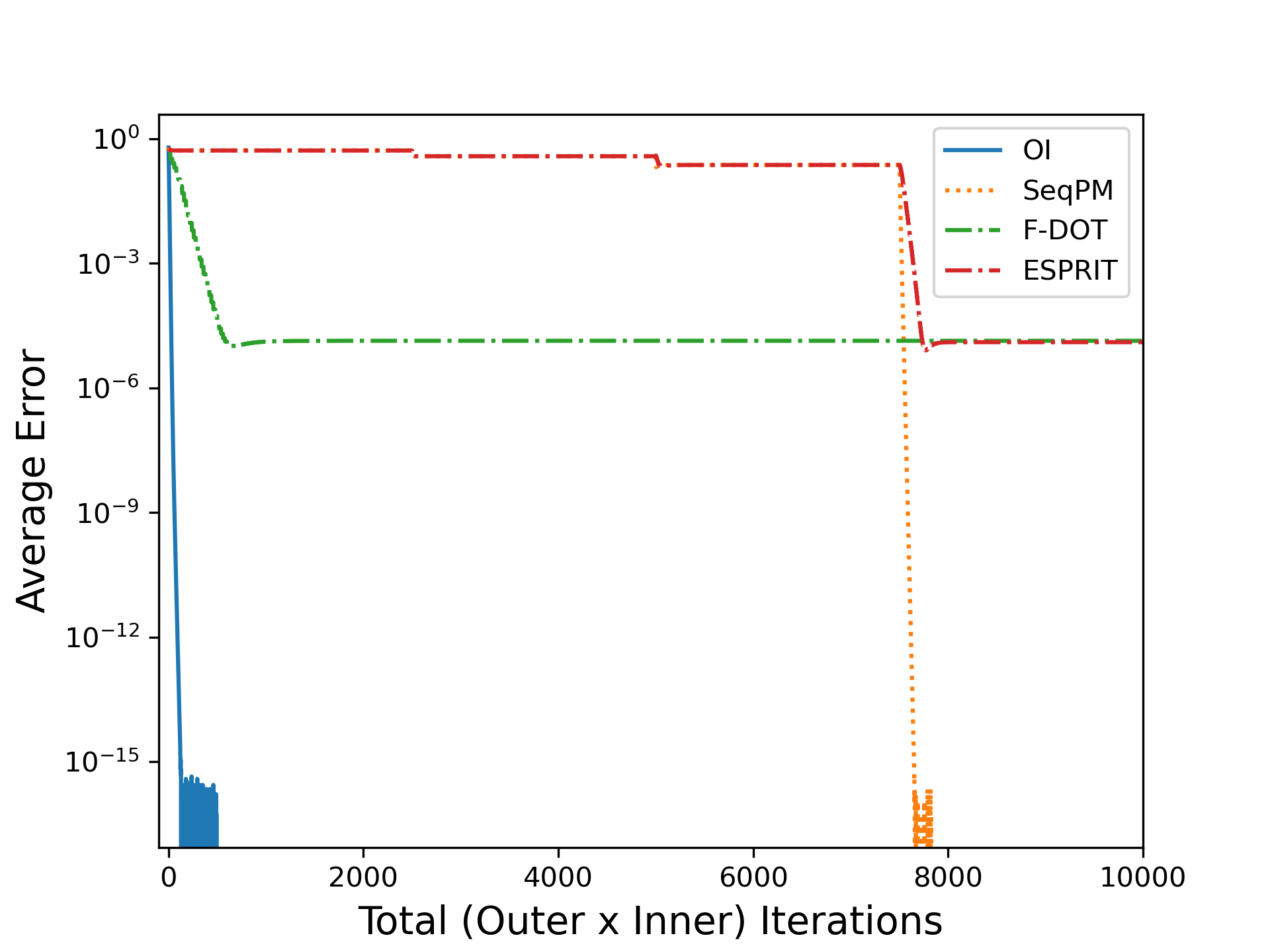}
                \caption{$r =4, \Delta_r = 0.85$}
                \label{fig:fc1b}
        \end{subfigure}%
        
        \begin{subfigure}{0.25\textwidth}
                \includegraphics[width=\linewidth]{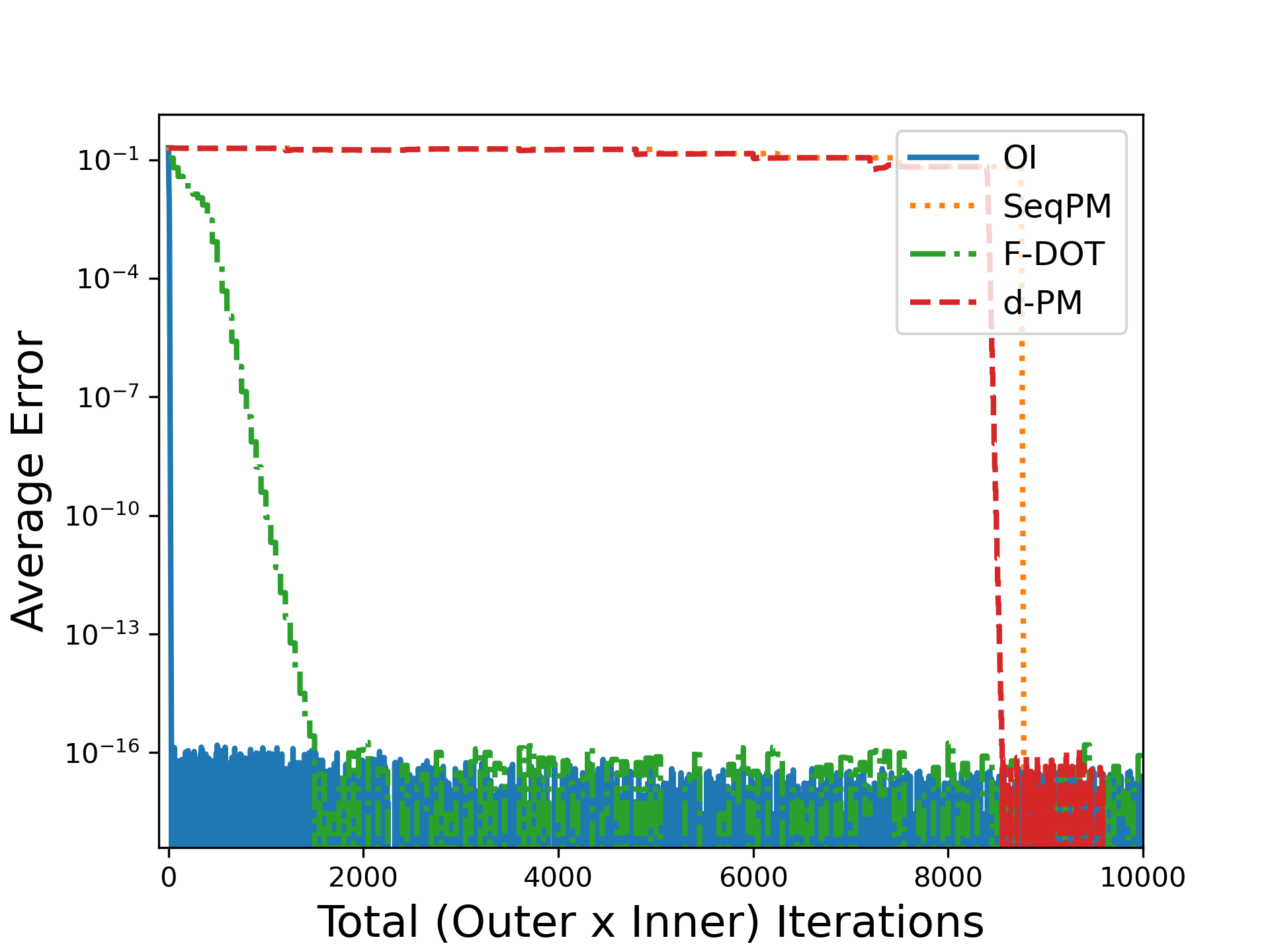}
                \caption{$r=8, \Delta_r = 0.4$}
                \label{fig:fc1c}
        \end{subfigure}%
        \begin{subfigure}{0.25\textwidth}
                \includegraphics[width=\linewidth]{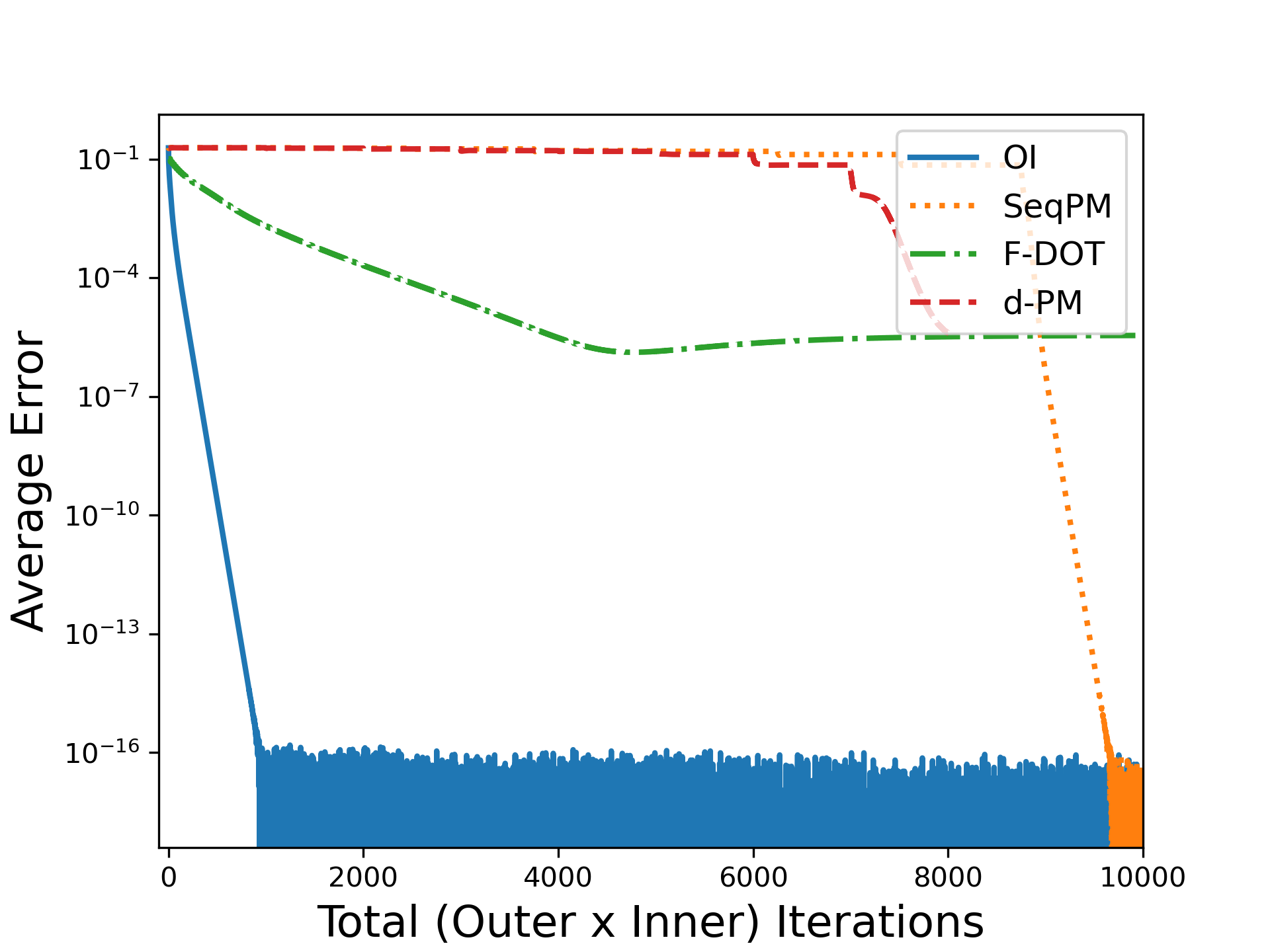}
                \caption{$r=8, \Delta_r = 0.85$}
                \label{fig:fc1d}
        \end{subfigure}%
        \caption{Performance comparison of F-DOT with OI, SeqPM and d-PM in the case of distinct eigenvalues.}\label{fig:feat_comp1}
\end{figure}

Next, we demonstrate the convergence behaviour of F-DOT algorithm for feature-wise partitioned data. There is not much work done for distributed PSA in this setting except the distributed power method (d-PM) in~\cite{scaglione2008decentralized}, which computes the $r$-dimensional subspace sequentially by estimating one vector at a time. Hence, we restrict comparison with only centralized OI, sequential power method (SeqPM) and d-PM. For this comparison, we generate Erd\H{o}s-Rènyi grapth with $N=10$ nodes and connectivity parameter $p=0.5$. The total dimension of the samples is $d=N$, i.e., each node carries one feature and $n=500$ samples. Figure~\ref{fig:feat_comp1} shows the comparison of our proposed algorithm F-DOT with OI for different eigenspace dimensions $r$ and eigengaps $\Delta_r$ when all the eigenvalues of the global covariance matrix $\bM$ are distinct. It is evident that in the case of feature-wise data partitioning our method once again significantly outperforms SeqPM and d-PM, thus emphasising the advantage of simultaneous estimation over sequential methods.

\subsection{Experiments Using Real-World Data}
In this section we demonstrate the performance of our proposed methods on real-world data for sample-wise partitioned data. For this purpose, we choose four widely used public datasets, viz., MNIST, CIFAR10, LFW and ImageNet. As pointed out earlier, the computation complexity of F-DOT is directly proportional to the number of samples $n$. Since all these real-world data sets have large $n$, we omit those experiments for feature-wise data partitioning case. The MNIST is a database of handwritten digits~\cite{deng2012mnist}. It contains $n=50,000$ gray-scale samples with each sample of dimension $d=784$. The Canadian Institute For Advanced Research 10 (CIFAR-10) dataset also consists of $n=50,000$ samples. Each sample has a dimension of $d=1024$~\cite{krizhevsky2014cifar}.
Labeled Faces in the Wild (LFW) face database is mainly a public benchmark for face recognition~\cite{huang2008labeled}, consisting of gray-scale images of a number of people's faces in different poses, distinct angles, and various light conditions. The number of training samples of LFW is $n=13,233$, with dimension of each being $d=2914$. The final dataset we use is ImageNet\cite{deng2009imagenet}. It is a huge dataset that contains 14 million color images over more than 20,000 categories. The dimension of the images are inconsistent and hence we reshape the images into a uniform dimension of $d=1024$. For each of these datasets, we show the comparison of P2P communications for S-DOT and SA-DOT. We also demonstrate the performance of our proposed algorithms with OI, SeqPM, DSA, DPGD, SeqDistPM, DeEPCA for MNIST and CIFAR10. The size of LFW and ImageNet datasets are too large to perform centralized OI and hence we leave out that comparison.
\begin{enumerate}
    \item \textbf{MNIST}: First, we compare the number of P2P communications for the two proposed algorithms S-DOT and SA-DOT in Table~\ref{Tab:m}. Each node in the connected network has $n_i=\left \lfloor \frac{50,000}{N} \right\rfloor$ local samples in $\R^{784}$. Figure~\ref{fig:MNIST} shows that we can achieve faster convergence with the SA-DOT algorithm compared to S-DOT (which uses a constant $T_c$). Figure~\ref{fig:mnist1} demonstrates how the average error of S-DOT and SA-DOT changes with the number of total iterations as compared to other methods. The number of nodes here is $N=10$.
\begin{table}
\caption{Parameters and P2P communication for MNIST experiments}
\centering
\begin{tabular}{|l|l|l|l|l|l|}
\hline
 $N$ & Erdős–Rényi: $p$ & $r$  & $T_o$ & Consensus Itr   & P2P $(K)$\\
\hline              
 20  &  0.25 & 5   & 400    & $t+1$     & 82.61    \\
     &       &     &        & $2t+1$    & 85.25    \\
     &       &     &        & 50      & 88       \\
\hline            
 20  &  0.25 & 10  & 400    & $t+1$   & 82.61   \\
     &       &     &        & $2t+1$  & 85.25   \\
     &       &     &        & 50    & 88      \\
\hline              
100  &  0.05 & 5   & 200    & $t+1$    & 43.88      \\
     &       &     &        & $2t+1$   & 46.875      \\
    &        &     &        & 50     & 50          \\
\hline
\end{tabular}
  \label{Tab:m}
\end{table}
\begin{figure}
        \centering
        \vspace{-10pt}
        \begin{subfigure}{0.25\textwidth}
                \includegraphics[width=\linewidth]{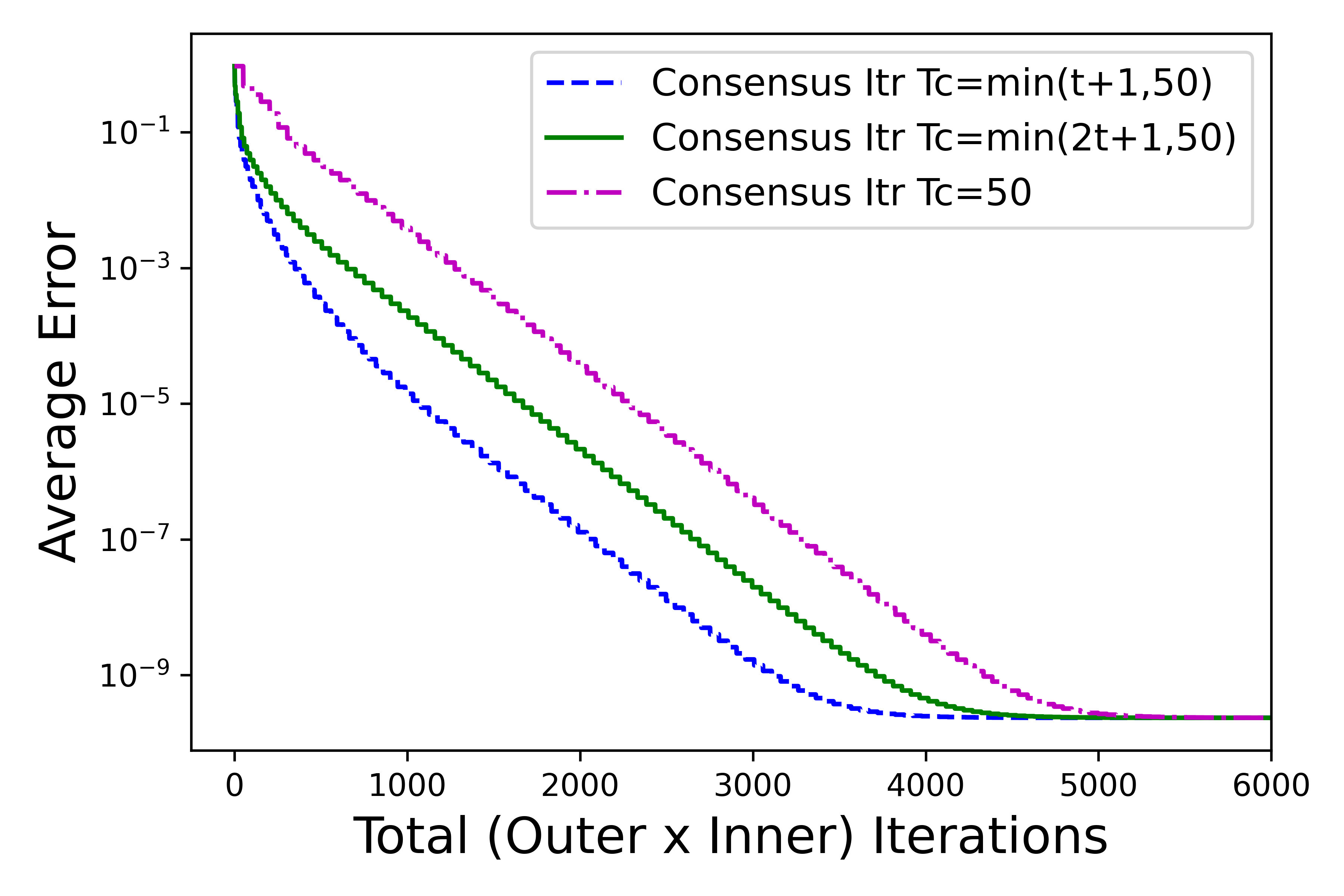}
                \caption{$N=20,r=5$}
                \label{fig:m2b}
        \end{subfigure}%
        \begin{subfigure}{0.25\textwidth}
                \includegraphics[width=\linewidth]{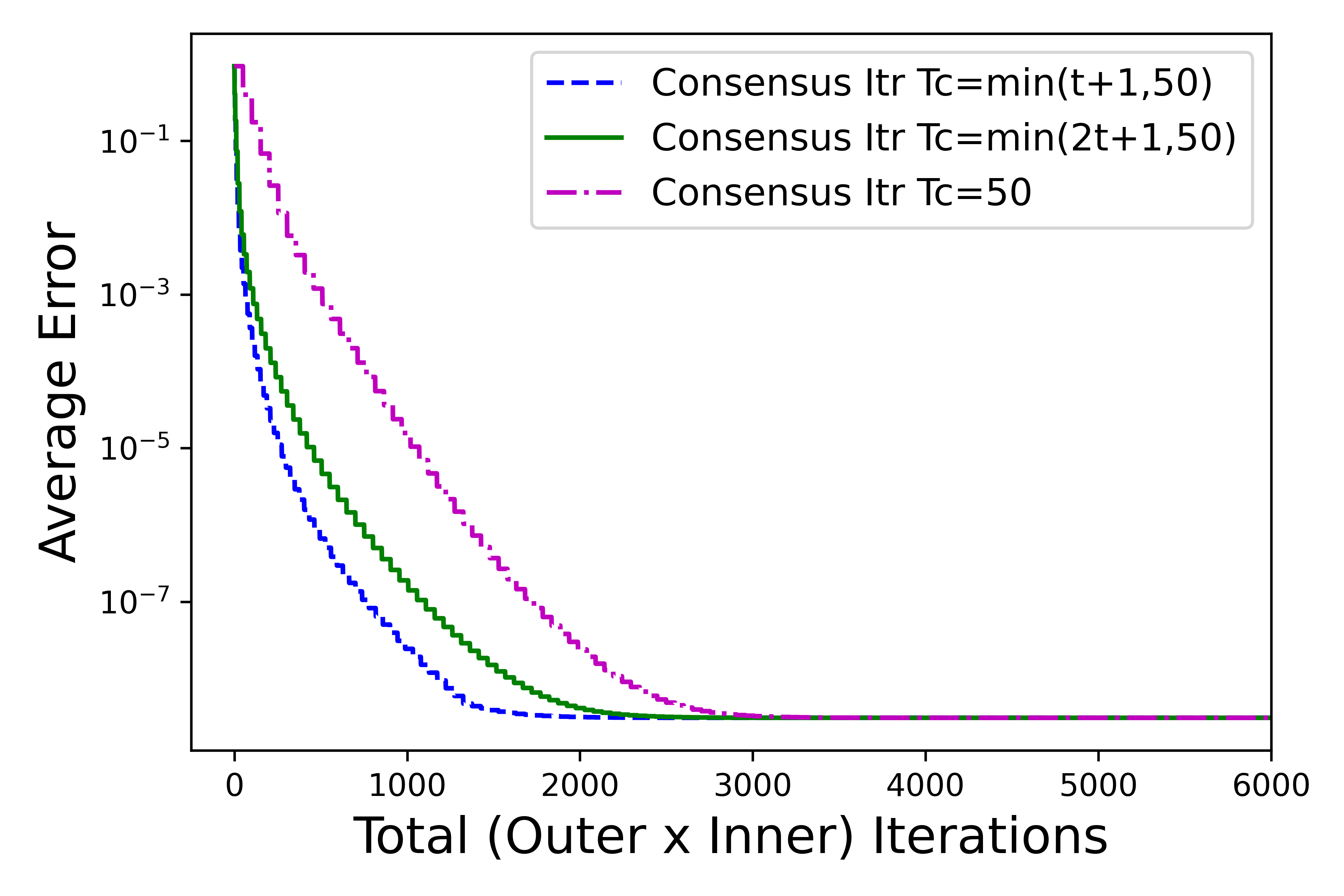}
                \caption{$N=100,r=5$}
                \label{fig:m3b}
        \end{subfigure}%
        \caption{Comparison of S-DOT and SA-DOT in terms of communication cost for MNIST dataset.}\label{fig:MNIST}
\end{figure}
\begin{figure}
        \centering
        \vspace{-10pt}
        \begin{subfigure}{0.25\textwidth}
                \includegraphics[width=\linewidth]{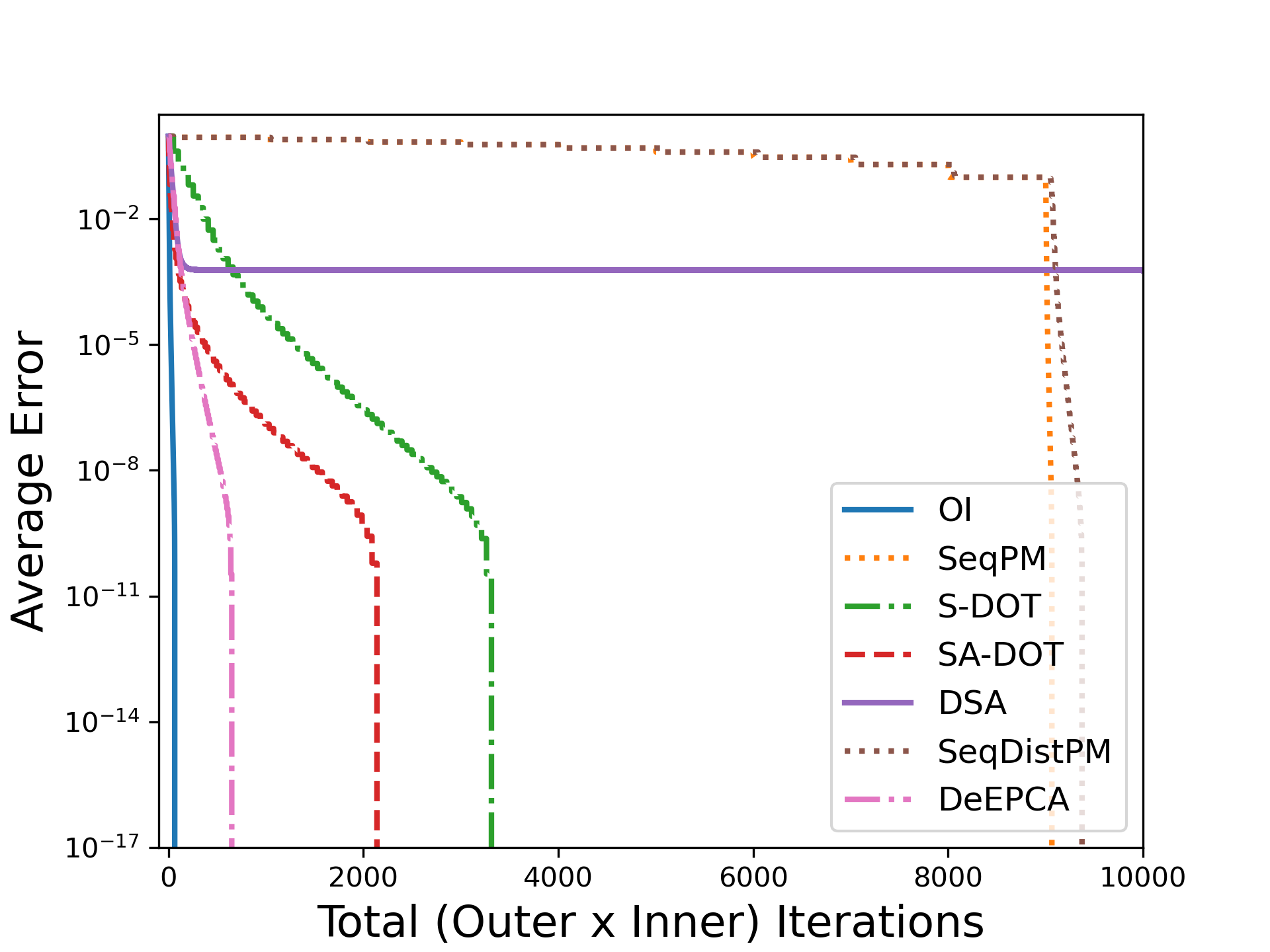}
                \caption{$r = 10$}
                \label{fig:m1a}
        \end{subfigure}%
        \begin{subfigure}{0.25\textwidth}
                \includegraphics[width=\linewidth]{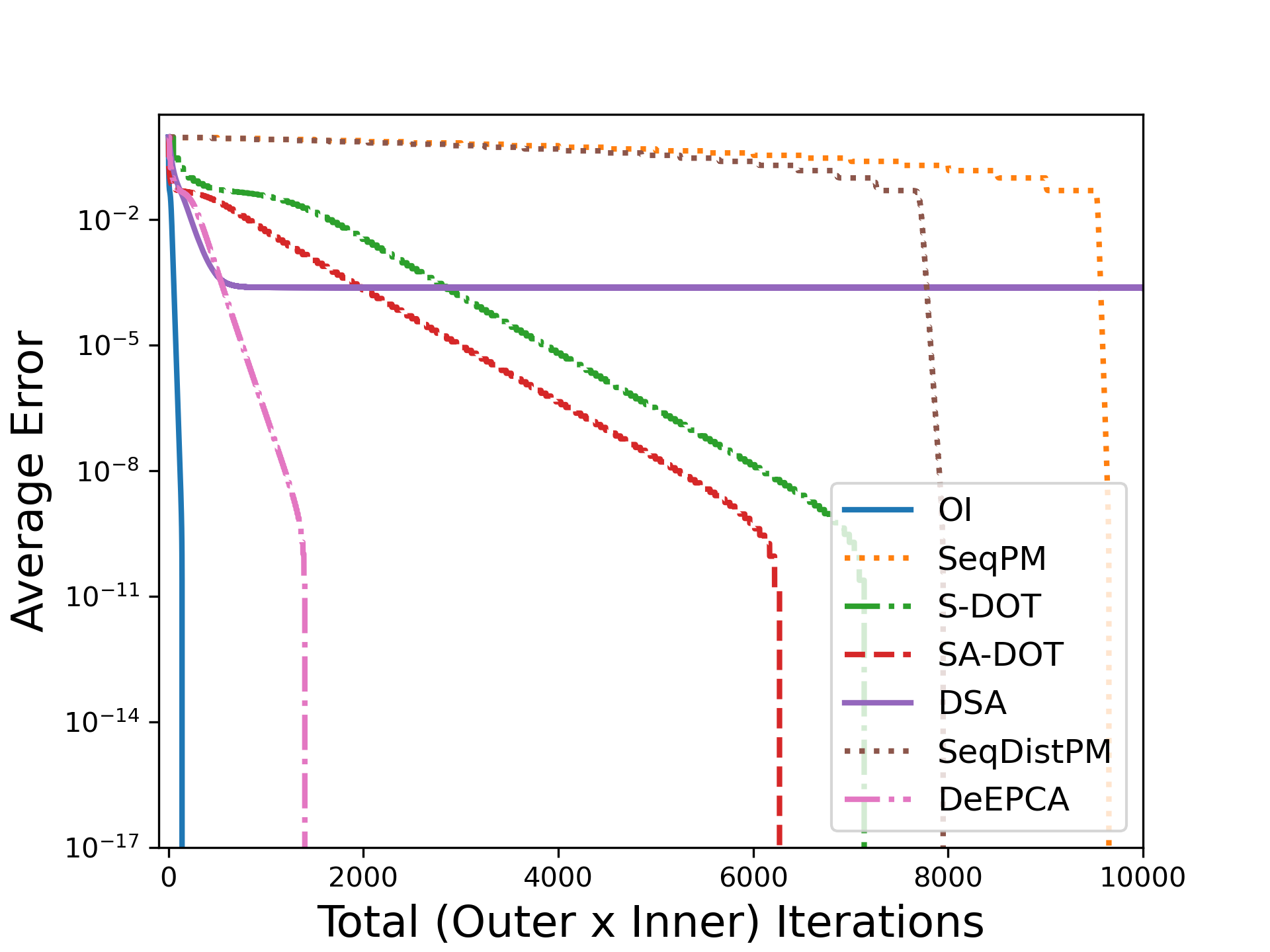}
                \caption{$r = 20$}
                \label{fig:m1b}
        \end{subfigure}%
        \caption{Performance comparison of S-DOT and SA-DOT with different centralized and distributed algorithms for MNIST data.}\label{fig:mnist1}
\end{figure}
 \item \textbf{CIFAR10}: Table~\ref{Tab:c} shows the comparison for P2P communications. Here, each node in the underlying connected network has $n_i=\left \lfloor \frac{50,000}{N} \right\rfloor$ local samples in $\R^{1024}$ and the plots in Fig.~\ref{fig:cifar} validate that SA-DOT algorithm again outperforms S-DOT in terms of communication cost. Figure~\ref{fig:cifar1} demonstrates how the average error of S-DOT and SA-DOT changes with the number of total iterations as compared to other methods.
 \begin{table}
\caption{Parameters and P2P communication for CIFAR-10 experiments}
\centering
\begin{tabular}{|l|l|l|l|l|l|}
\hline
 $N$ & Erdős–Rényi: $p$ & $r$  & $T_o $& Consensus Itr  & P2P $(K)$\\
\hline              
 20  &  0.25 & 5   & 400    & $t+1$    & 76.98       \\
     &      &      &        & $2t+1$   & 79.44       \\
     &      &      &        & 50     & 82          \\
\hline            
 20  &  0.25 & 7   & 400    & $t+1$    & 76.98       \\
     &      &      &        & $2t+1$   & 79.44       \\
     &      &      &        & 50     & 82          \\
\hline              
100  &  0.05 & 7   & 400    & $t+1$    & 44.4       \\
     &       &     &        & $2t+1$   & 98.4       \\
     &       &     &        & 50    & 101.12      \\
\hline
\end{tabular}
 \label{Tab:c}
\end{table}
\begin{figure}
        \centering
        \vspace{-10pt}
        \begin{subfigure}{0.25\textwidth}
                \includegraphics[width=\linewidth]{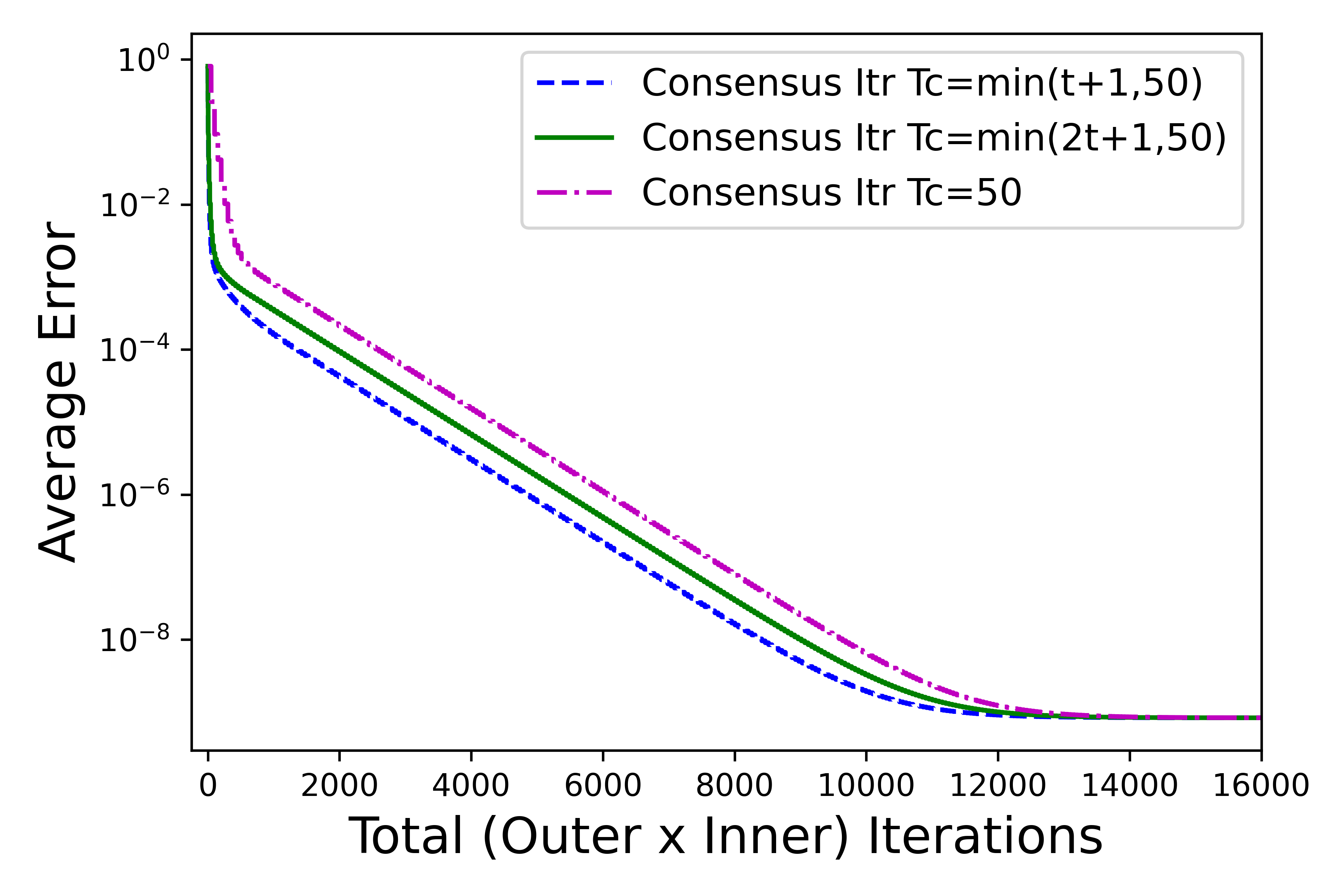}
                \caption{$N=20,r=5$}
                \label{fig:ca}
        \end{subfigure}%
        \begin{subfigure}{0.25\textwidth}
                \includegraphics[width=\linewidth]{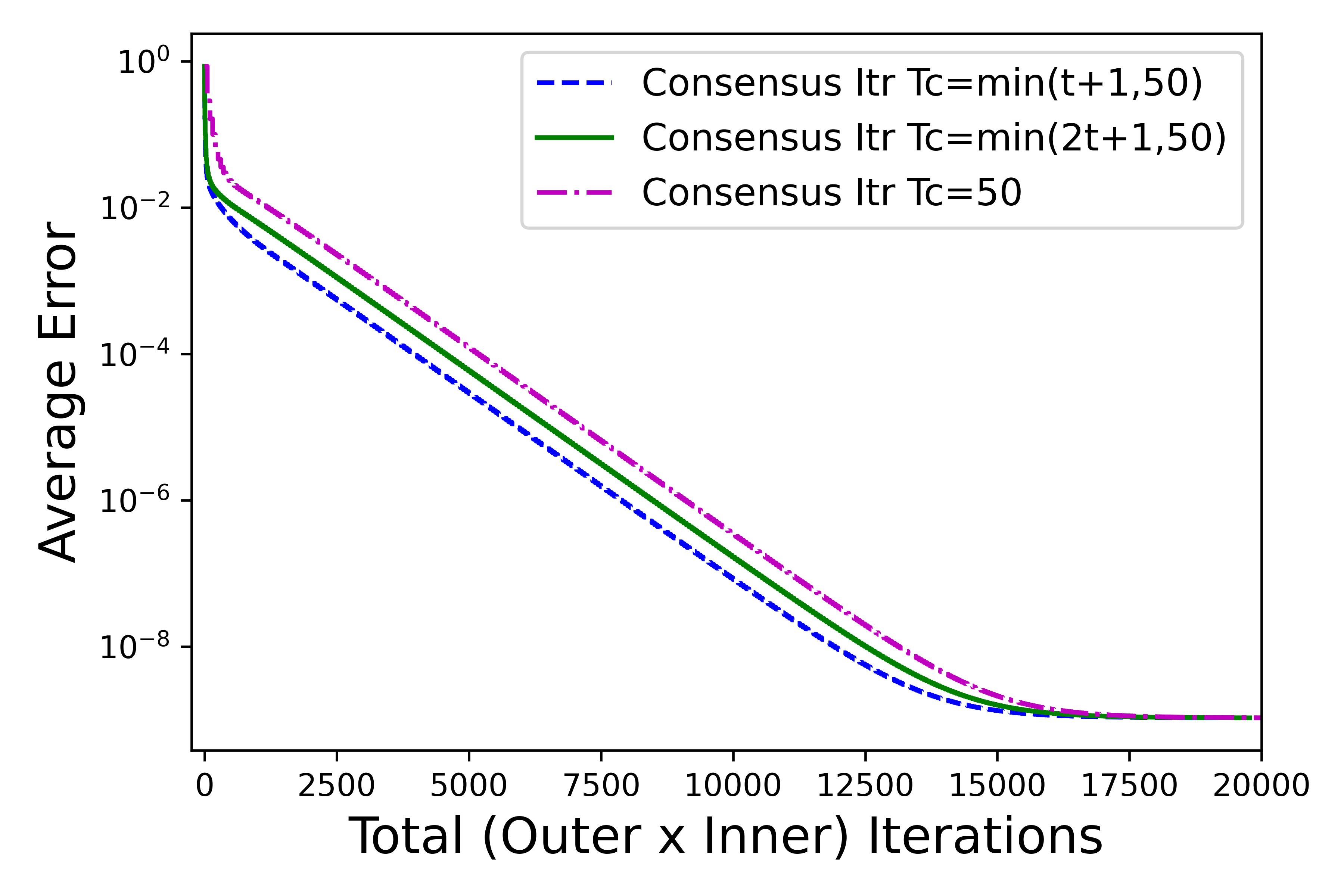}
                \caption{$N=20,r=7$}
                \label{fig:cc}
        \end{subfigure}
        \caption{Comparison of S-DOT and SA-DOT in terms of communication cost for CIFAR10 dataset.}\label{fig:cifar}
\end{figure}
\begin{figure}
        \centering
        \vspace{-10pt}
        \begin{subfigure}{0.25\textwidth}
                \includegraphics[width=\linewidth]{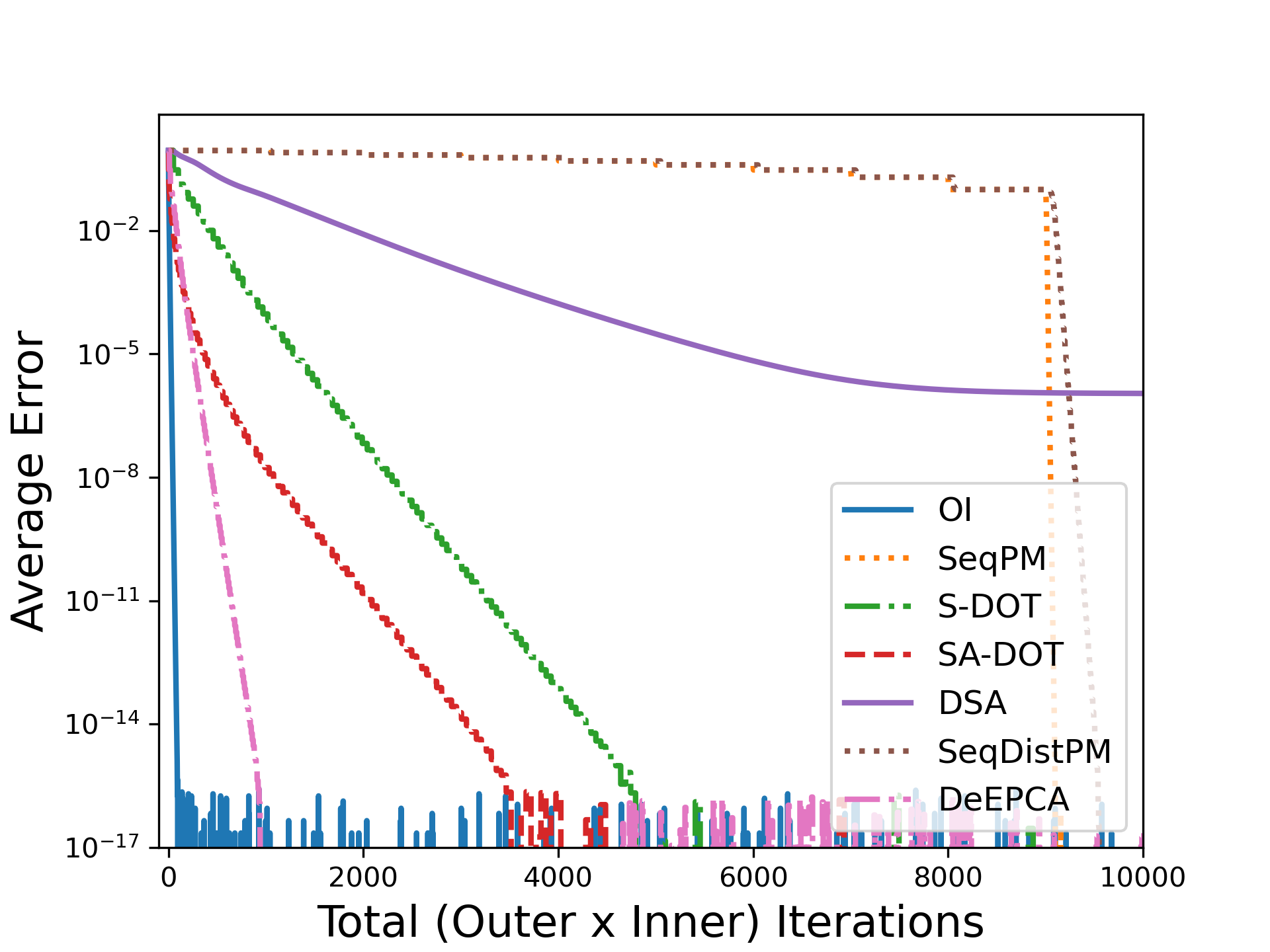}
                \caption{$r = 10$}
                \label{fig:cf1a}
        \end{subfigure}%
        \begin{subfigure}{0.25\textwidth}
                \includegraphics[width=\linewidth]{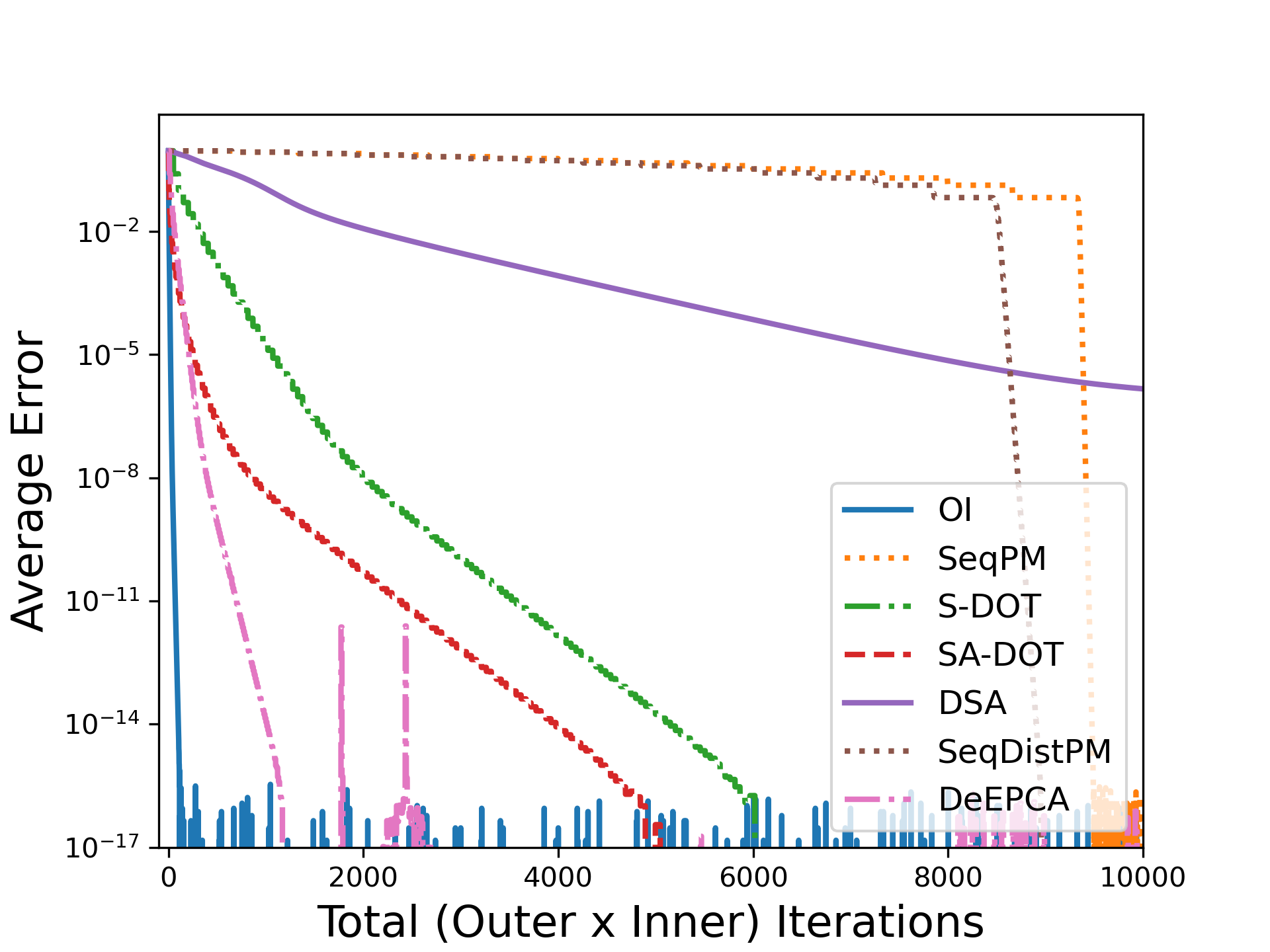}
                \caption{$r = 15$}
                \label{fig:cf1b}
        \end{subfigure}%
        \caption{Performance comparison of S-DOT and SA-DOT with different centralized and distributed algorithms for CIFAR10 data.}\label{fig:cifar1}
\end{figure}
    \item \textbf{LFW}: The experiment parameters for LFW are provided in Table~\ref{Tab:l}. Each node in the connected network has $n_i=\left \lfloor \frac{13233}{N} \right\rfloor$ local samples in $\R^{2914}$ and $r$ is set to be 7. Results in Fig.~\ref{fig:lfw} show how increasing number of consensus iterations per orthogonal iteration causes slower convergence because of unnecessary communications.
\begin{table}
\caption{Parameters and P2P communication for LFW experiments}
\centering
\begin{tabular}{|l|l|l|l|l|l|l|}
\hline
 $N$ & Erdős–Rényi: $p$ & $r$  & Consensus Itr  & P2P $(K)$\\
\hline              
 20  &  0.25 & 7    & $t+1$       & 42.12       \\
     &      &       & $2t+1$   & 45       \\
     &      &       & 50     & 48       \\
\hline
20   &  0.5 & 7    & $t+1$     & 82.49    \\
     &      &      & $2t+1$    & 88.13    \\
     &      &      & 50      & 94       \\
\hline
\end{tabular}
  \label{Tab:l}
\end{table}
\begin{figure}
        \centering
        \vspace{-10pt}
        \begin{subfigure}[b]{0.25\textwidth}
                \includegraphics[width=\linewidth]{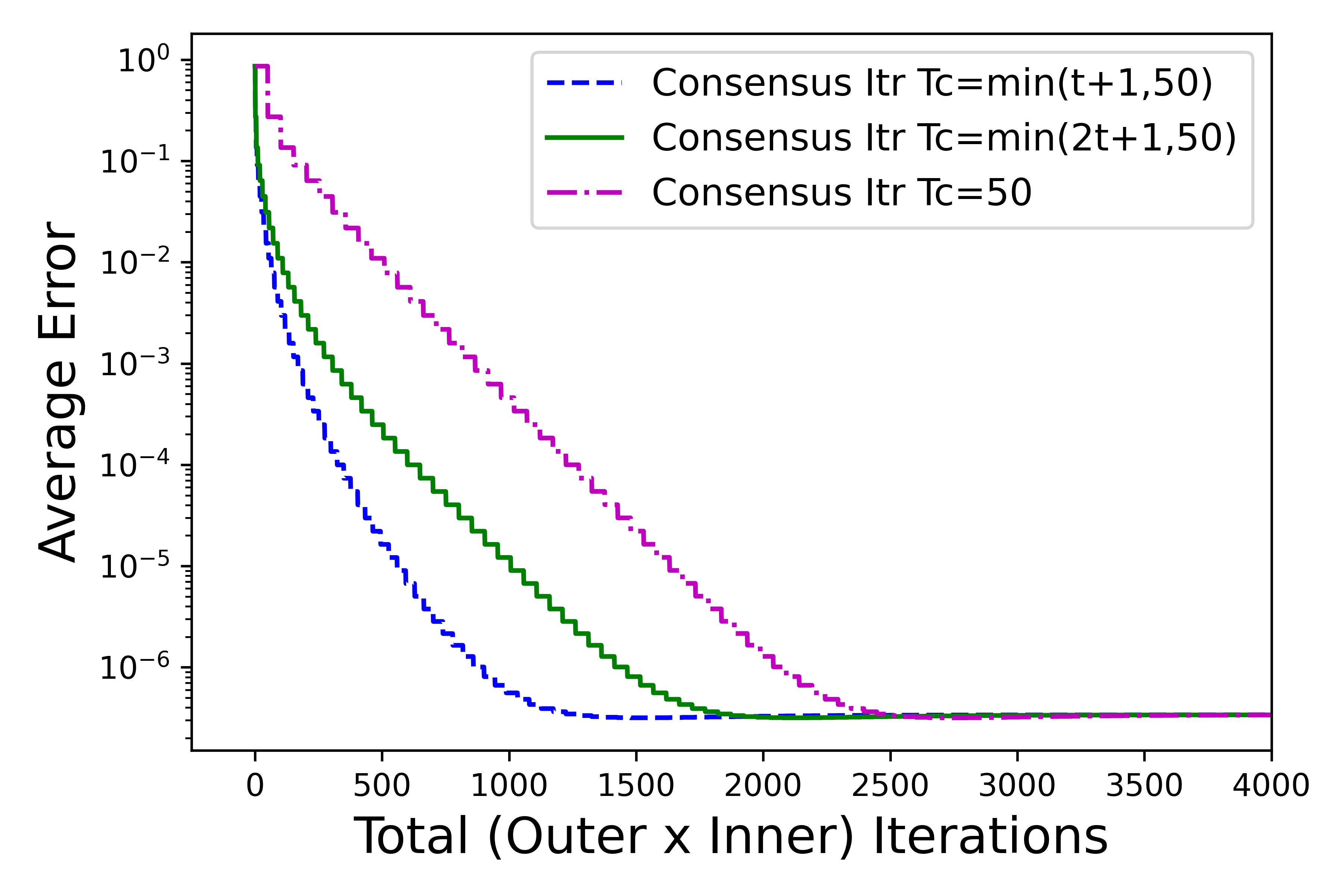}
                \caption{$N=10,p=0.5$}
                \label{fig:lc}
        \end{subfigure}%
        \begin{subfigure}[b]{0.25\textwidth}
                \includegraphics[width=\linewidth]{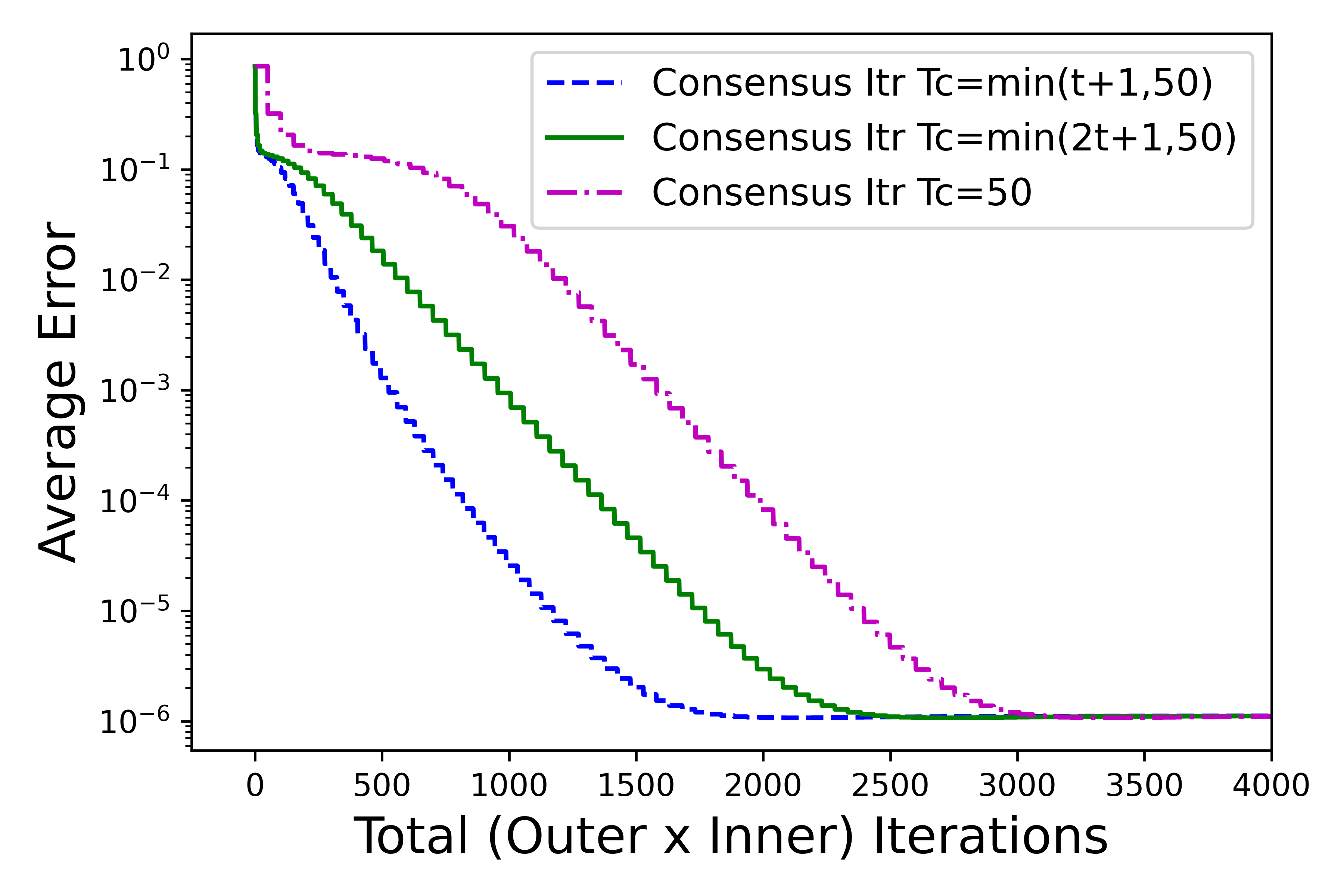}
                \caption{$N=20,p=0.5$}
                \label{fig:ld}
        \end{subfigure}%
        \caption{Comparison of S-DOT and SA-DOT in terms of communication cost for LFW dataset.}\label{fig:lfw}
\end{figure}
    \item \textbf{ImageNet}: The experiment parameters are given in Table~\ref{Tab:i}, where each node in the connected network has $n_i=5000$ local samples in $\R^{1024}$ and $r$ is set to be 5. The results for the ImageNet dataset are shown in Fig.~\ref{fig:imagenet}, which indicate that increasing the number of consensus iterations faster helps achieve faster convergence of the SA-DOT algorithm. 
\end{enumerate}
\begin{table}
\caption{Parameters and P2P communication for ImageNet experiments}
\centering
\begin{tabular}{|l|l|l|l|l|l|}
\hline
 $N$ & Erdős–Rényi: $p$ & $r$  & Consensus Itr  & P2P $(K)$\\
\hline              
 10  &  0.5 & 5   & $t+1$     & 35.1       \\
     &      &     & $2t+1$    & 37.5       \\
     &      &     & 50      & 40         \\
\hline              
 20  &  0.25 & 5  & $t+1$     & 32.47     \\
     &       &    & $2t+1$   & 34.69     \\
     &       &    & 50      & 37        \\
\hline 
100  &  0.05 & 5   & $t+1$    & 47.91         \\
     &       &     & $2t+1$   & 51.19          \\
     &       &     & 50     & 54.6         \\
\hline  
200  &  0.03 & 5   & $t+1$  & 50.37         \\
     &       &     & $2t+1$  & 53.81         \\
     &       &     & 50   & 57.4        \\
 \hline
\end{tabular}
  \label{Tab:i}
\end{table}
\begin{figure}
        \vspace{-10pt}
        \begin{subfigure}[b]{0.25\textwidth}
                \includegraphics[width=\linewidth]{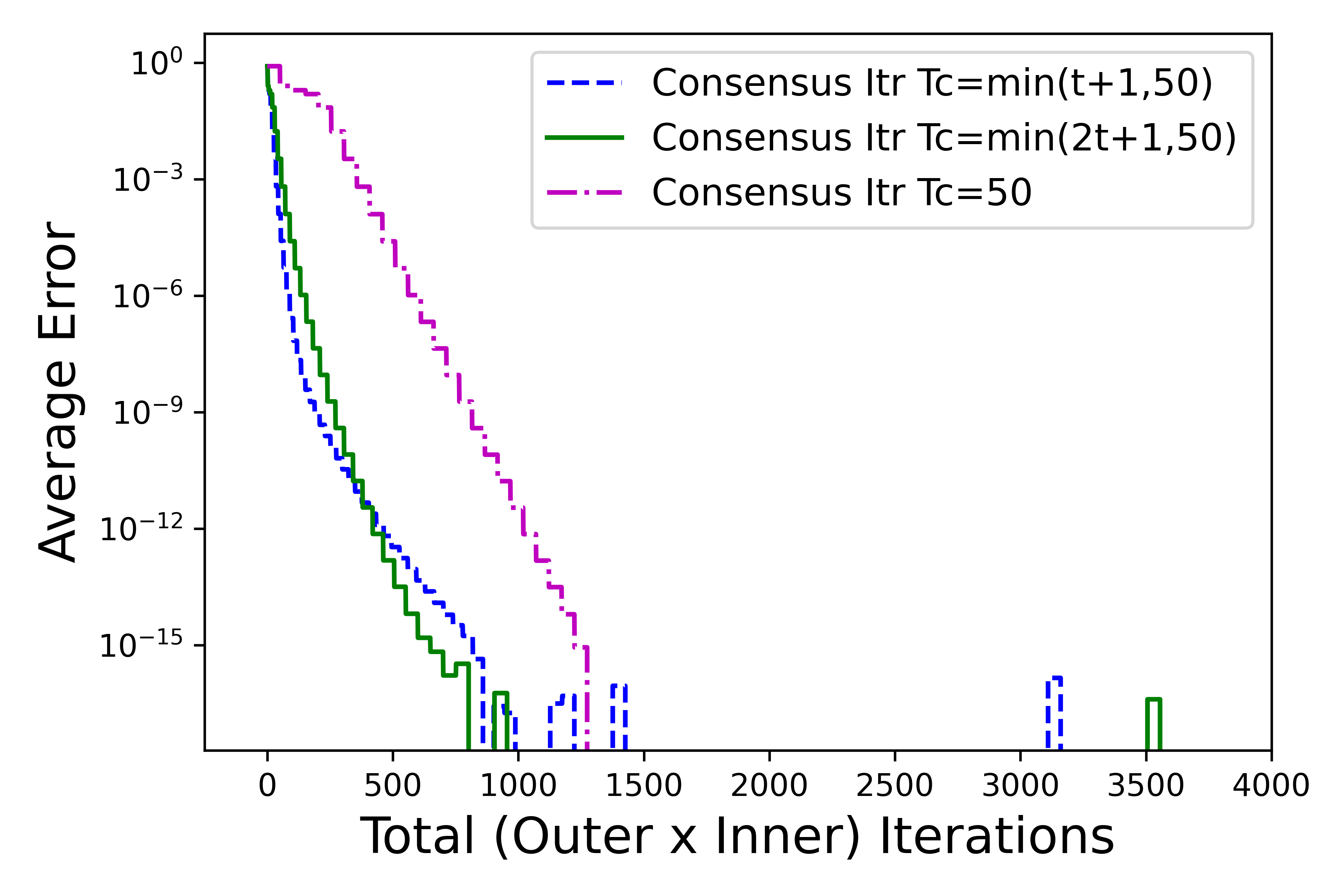}
                \caption{$N=10$}
                \label{fig:i1b}
        \end{subfigure}%
        \begin{subfigure}[b]{0.25\textwidth}
                \includegraphics[width=\linewidth]{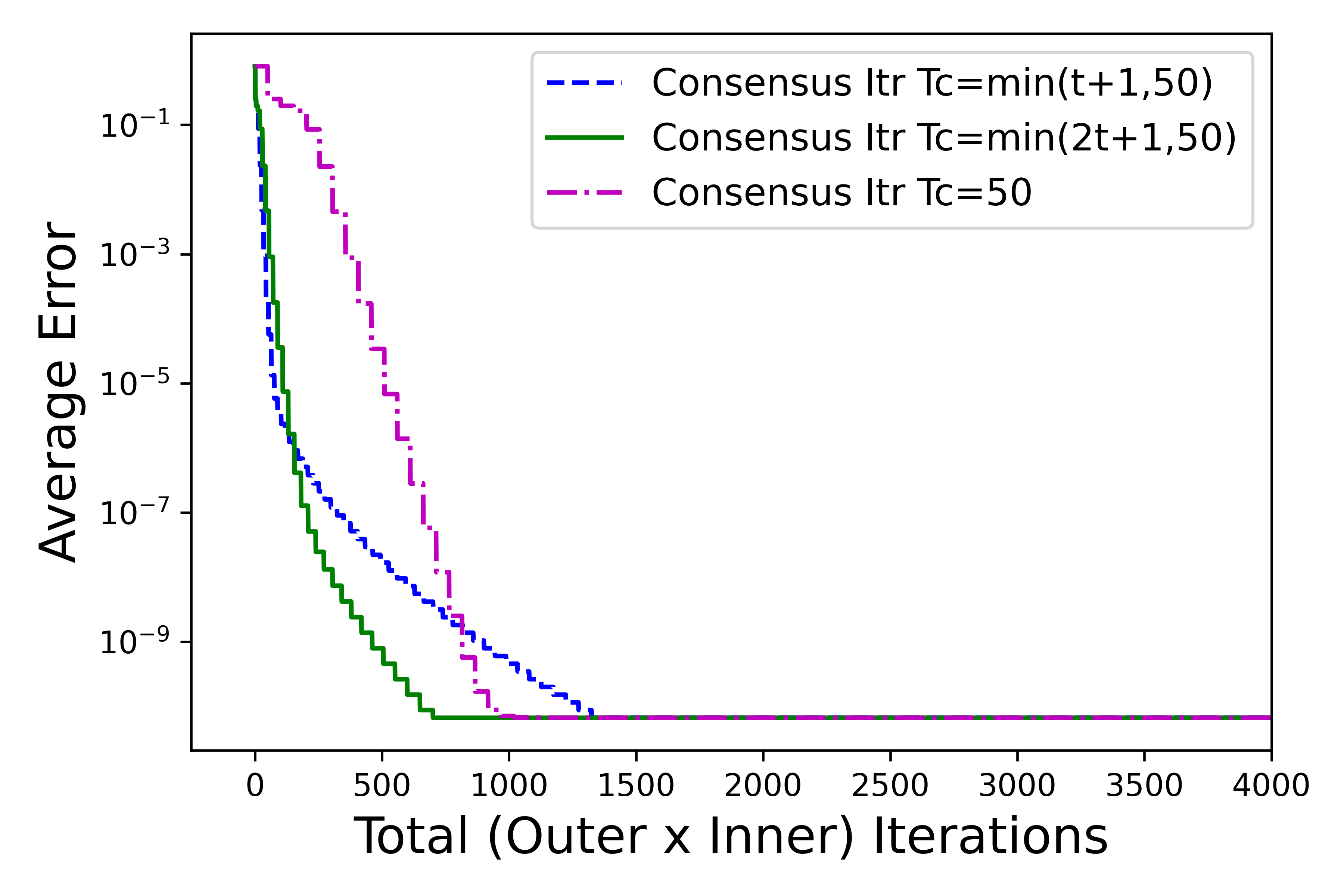}
                \caption{$N=20$}
                \label{fig:i2b}
        \end{subfigure}%
        
        \begin{subfigure}[b]{0.25\textwidth}
                \includegraphics[width=\linewidth]{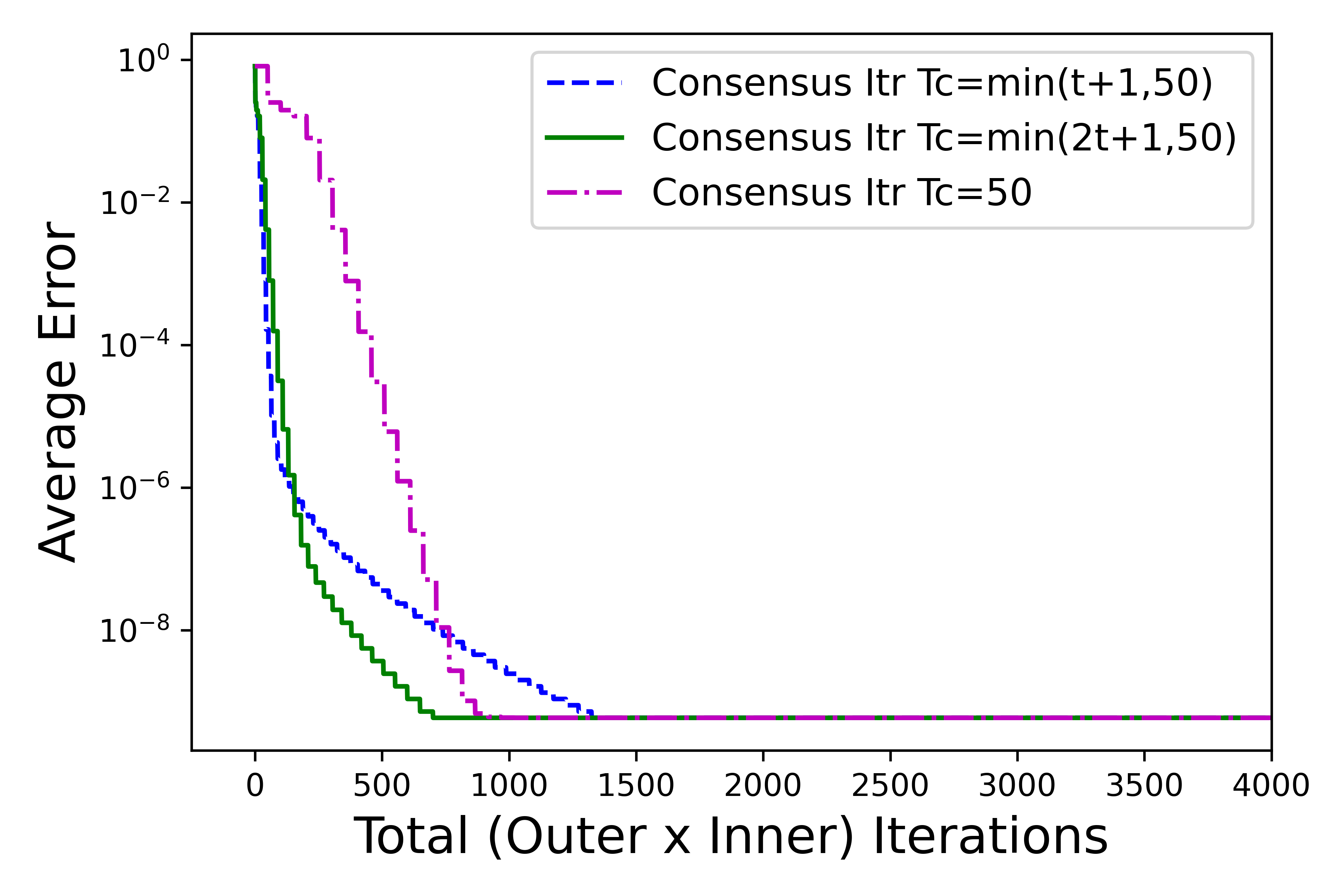}
                \caption{$N=100$}
                \label{fig:i100}
        \end{subfigure}%
        \begin{subfigure}[b]{0.25\textwidth}
                \includegraphics[width=\linewidth]{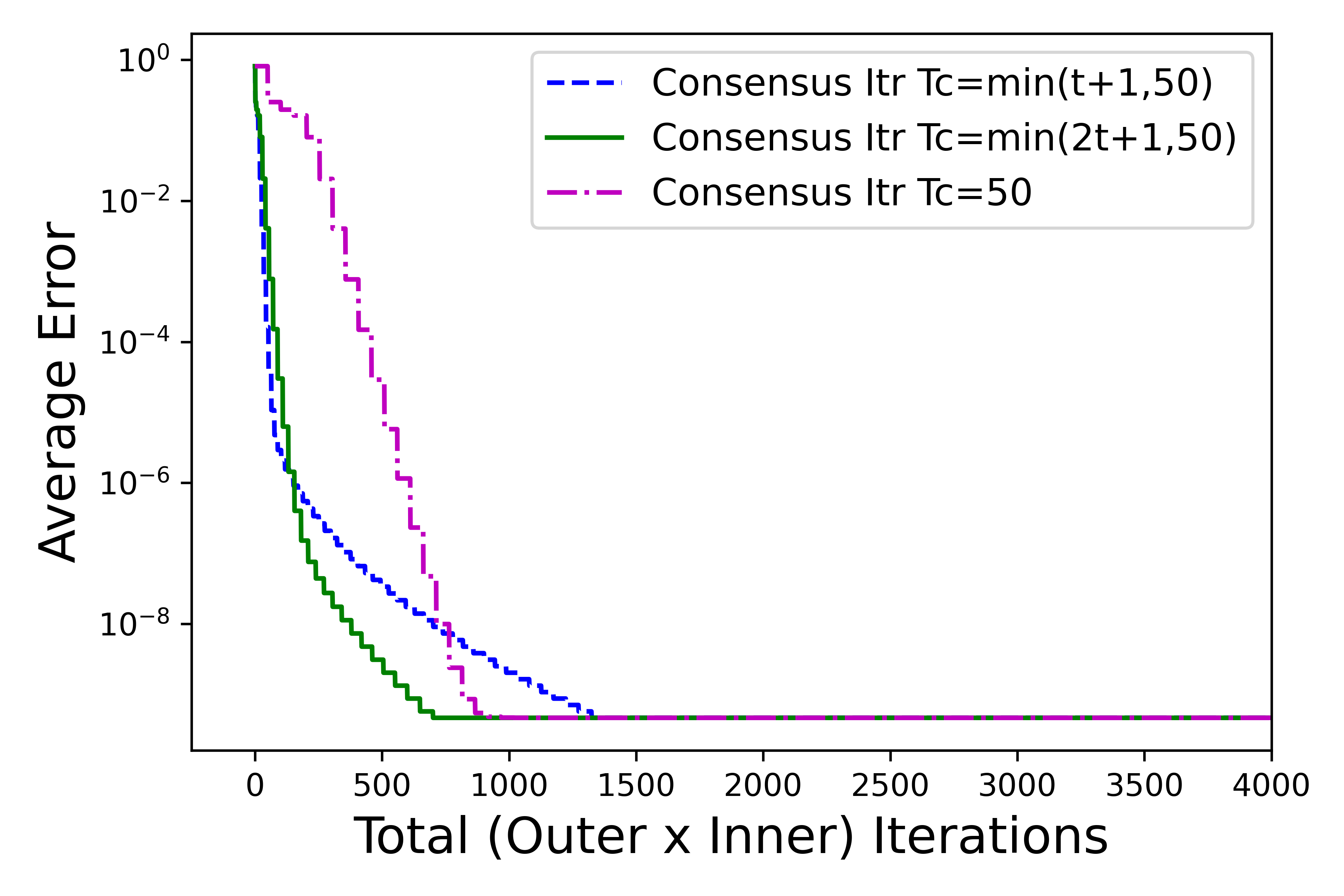}
                \caption{$N=200$}
                \label{fig:i200}
        \end{subfigure}%
        \caption{Comparison of S-DOT and SA-DOT in terms of communication cost for ImageNet dataset}\label{fig:imagenet}
\end{figure}
\label{sec:num_res}

\section{Conclusion}\label{sec:conc}
In this paper, we addressed the problem of Principal Component Analysis (PCA) in a distributed setting defined by an arbitrarily connected network without any central server. Data can be partitioned in different ways in a network and here we considered two kinds of data partitioning: by samples and by features. For sample-wise partitioned data, we proposed an algorithm Sample-wise Distributed Orthogonal iTeration (S-DOT) and an adaptive variant of it called Sample-wise Adaptive Distributed Orthogonal iTeration (SA-DOT). Theoretical convergence guarantees for both these algorithms were provided, which show that for sufficient number of consensus iterations per orthogonal iteration, both S-DOT and SA-DOT have a linear convergence rate. Numerical results on synthetic as well as real-world data were presented to further demonstrate the efficacy of our proposed algorithms. Furthermore, we also proposed an algorithm for feature-wise partitioned data called Feature-wise Distributed Orthogonal iTeration (F-DOT). Even though we do not provide theoretical guarantees for F-DOT, extensive numerical experiments on synthetic data show the effectiveness of the proposed solution.

In the future, providing theoretical guarantees for F-DOT is an obvious extension. Also, as pointed out earlier, in case of data that has both high dimension and large number of samples the proposed F-DOT algorithm will have high communication and computation costs. Randomly block-wise partitioned data, i.e., data partitioned by both samples and features, can be a possible way to handle big data that is massive in both dimension and size. Thus, block-partitioning is a probable solution for such massive data and developing solutions for such partitioning is a direction for future.
\begin{appendices}
\section{Proof of Lemma 1}\label{app:lemma1n2}

Let $\bV_{s,i}$ be the value from Step~\ref{algostep:consensus_result} in Algorithm \ref{alg:cdot} during the $(t_o+1)^{th}$ iteration of S-DOT and SA-DOT at node $i$ and let $\bV_c = \bM\bQ_c$ be the corresponding value in case of centralized OI. From Step 12, we know $\bQ_{s,i}^\prime\bR_{s,i} = \bV_{s,i}$. Similarly, in case of OI we will have $\bQ_{c}^\prime\bR_{c} = \bV_c$. Thus $\bQ_c^\prime =\bV_c\bR_c^{-1}$, and $\bQ_{s,i}^\prime =\bV_{s,i} \bR_{s,i}^{-1}$. Therefore,
\begin{align}
(\bQ_c^\prime-\bQ_{s,i}^\prime) &=\bV_c\bR_c^{-1}-\bV_{s,i}\bR_{s,i}^{-1} \nonumber\\
                                &=\bV_c\bR_c^{-1}-\bV_c\bR_{s,i}^{-1}+\bV_c\bR_{s,i}^{-1}-\bV_{s,i}\bR_{s,i}^{-1} \nonumber\\
                                &=\bV_c(\bR_c^{-1}-\bR_{s,i}^{-1})+(\bV_c-\bV_{s,i})\bR_{s,i}^{-1}.
\end{align}
Using the triangle inequality, we obtain
\begin{align}\nonumber
\norm{\bQ_c^\prime-\bQ_{s,i}^\prime}_F \leq &\norm{\bV_c-\bV_{s,i}}_F\cdot\max_i\norm{\bR_{s,i}^{-1}}_F +\\\label{eq:2_15} &\norm{\bV_c}_F\cdot\max_i\norm{\bR_c^{-1}-\bR_{s,i}^{-1}}_F.
\end{align}
Therefore, if we want to bound $\norm{\bQ_c^\prime-\bQ_{s,i}^\prime}_F$ we need to bound $\norm{\bV_c-\bV_{s,i}}_F$, $\norm{\bR_{s,i}^{-1}}_F$, $\norm{\bV_c}_F$, and $\norm{\bR_c^{-1}-\bR_{s,i}^{-1}}_F$. Let $\bV_{s}=\sum_{i=1}^N(\bM_i\bQ_{s,i})$ and note that $\bV_{s,i}=\bV_{s}+\bcE_{c,i}$, where $\bcE_{c,i}$ is the consensus error after $T_c$ consensus iteration at node $i$. Suppose $\bZ_i^{(0)}=\bM_i\bQ_{s,i} \in \R^{d\times r}$, then using Proposition~\ref{theo:consensus}, we have that
\begin{align}\label{eq:2_16}
    \norm{\bcE_{c,i}}_F &=\norm{\bV_{s,i}-\bV_{s}}_F \nonumber\\\
                            &=\norm{\bM\bQ_{s,i}-\sum_{j=1}^N(\bM_j\bQ_{s,j})}_F\leq\delta\norm{\bZ^\prime}_F,
\end{align}
where $\bZ^\prime(j,k)=\sum_{i=1}^{N}\abs{\bZ_{i}^{(0)}(j,k)}$. We know
\begin{equation}
    \norm{\bZ^\prime}_F^2 =\sum_{j=1}^n\sum_{k=1}^r\left(\sum_{i=1}^N\abs{\bZ_{i}^{(0)}(j,k)}\right)^2.
\end{equation}
Using Cauchy-Schwarz inequality, $\abs{\sum_{i=1}^Na_i\cdot 1}^2\leq\left(\sum_{i=1}^N a_i^2\right)\cdot N$, we obtain
\begin{align}
    \norm{\bZ^\prime}_F^2 & \leq N\sum_{j=1}^n\sum_{k=1}^r\sum_{i=1}^N\abs{\bZ_{i}^{(0)}(j,k)}^2 \nonumber \\
                        & = N\sum_{i=1}^N\norm{\bZ_i^{(0)}}_F^2 = N\sum_{i=1}^N(\norm{\bM_i\bQ_{s,i}}_F^2).
\end{align}
Using the property $\norm{\bA\bB}_F\leq \norm{\bA}_2 \norm{\bB}_F$ and the fact that $\bQ_{s,i}$ are orthonormal matrices with rank $r$, we have
\begin{align}
 \norm{\bZ^\prime}_F^2 & \leq N \sum_{i=1}^N\left(\norm{\bM_i}^2_2\cdot \norm{\bQ_{s,i}}^2_F\right)\nonumber\\
                     & \leq N \left(\sum_{i=1}^N\norm{\bM_i}^2_2\right) \cdot r  \leq N\gamma^2r.
\end{align}
Therefore,
\begin{equation}\label{eq:2_20}
    \norm{\bZ^\prime}_F \leq \gamma\sqrt{Nr}.
\end{equation}
From \eqref{eq:2_16} and \eqref{eq:2_20} we have that
\begin{equation}\label{eq:2_21}
    \norm{\bcE_{c,i}}_F \leq \delta\gamma\sqrt{Nr}.
\end{equation}
From \eqref{eq:2_21} and $\bV_{s,i}=\bV_{s}+\bcE_{c,i}$, we have
\begin{align}
        \bV_c-\bV_{s,i}&=\bV_c-(\bV_{s}+\bcE_{c,i})\nonumber\\
        & = \bM \bQ_c-\sum_{i=1}^N \bM_i \bQ_{s,i}-\bcE_{c,i}\nonumber\\
        & = \sum_{i=1}^N\bM_i(\bQ_c-\bQ_{s,i})-\bcE_{c,i}.
\end{align}
Therefore, we get
\begin{align}
        \norm{\bV_c-\bV_{s,i}}_F & \leq  \sum_{i=1}^N\norm{\bM_i(\bQ_c-\bQ_{s,i})}_F+\norm{\bcE_{c,i}}_F \nonumber\\
        & \leq \sum_{i=1}^N \norm{\bM_i}_2 \norm{\bQ_c-\bQ_{s,i}}_F + \delta\gamma\sqrt{N r} \nonumber\\
        & \leq \alpha \max_i \norm{\bQ_c-\bQ_{s,i}}_F+\delta\gamma\sqrt{N r}.
\end{align}
Next, we bound $\norm{\bV_c}_F$ and $\norm{\bV_{s,i}}_F$ as follows:
\begin{align}
    \norm{\bV_c}_F &= \norm{\bM \bQ_c}_F \leq\norm{\bM}_2\norm{\bQ_c}_F \nonumber\\
    & = \norm{\sum_{i=1}^N \bM_i}_2\norm{\bQ_c}_F \leq\sum_{i=1}^N \norm{\bM_i}_2\norm{\bQ_c}_F  \leq \alpha\sqrt{r},
\end{align}
and
\begin{align}
    \|\bV_{s,i}\|_F & = \|\bV_{s}+\bcE_{c,i}\|_F \nonumber\\
            & = \|\sum_{i=1}^N (\bM_i\bQ_{s,i})+\bcE_{c,i}\|_F \nonumber\\
            & \leq \norm{\sum_{i=1}^N (\bM_i\bQ_{s,i})}_F+\delta\gamma\sqrt{N r}, \quad \text{from~\eqref{eq:2_21}}\nonumber\\
            & \leq \sum_{i=1}^N \norm{\bM_i\bQ_{s,i}}_F+\delta\gamma\sqrt{N r}\nonumber\\
            & \leq \sum_{i=1}^N \norm{\bM_i}_2\sqrt{r}+\delta\gamma\sqrt{N r} \leq \alpha\sqrt{r}+\delta\gamma\sqrt{N r} \text{.}
\end{align}
Next, we bound $\norm{\bR_{s,i}^{-1}}_F$ and $\norm{\bR_c^{-1}-\bR_{s,i}^{-1}}_F$. Define $\bK_c\coloneqq \bV_c^\tT\bV_c=\bR_c^\tT\bR_c$, and $\bK_{s,i}\coloneqq \bV_{s,i}^\tT\bV_{s,i}=\bR_{s,i}^\tT\bR_{s,i}$. Thus, $\bR_c$ and $\bR_{s,i}$ are non-singular matrices that denote the Cholesky decomposition of symmetric matrices $\bK_c$ and $\bK_{s,i}$, respectively. For such non-singular matrices $\bR_c$ and $\bR_{s,i}$, a theorem by Wedin \cite{wedin1973perturbation} states that
\begin{align}\nonumber
   &\norm{\bR_c^{-1}-\bR_{s,i}^{-1}}_2 \\
   &\leq \frac{1+\sqrt{5}}{2}\norm{\bR_c-\bR_{s,i}}_2\max\left\{\norm{\bR_c^{-1}}_2^2,\norm{\bR_{s,i}^{-1}}_2^2\right\}.
 \end{align}
Another theorem in~\cite{stewart1997perturbation} states that if $\bK_c=\bR_c^\tT\bR_c$, and $\bK_{s,i}=\bR_{s,i}^\tT\bR_{s,i}$ are Cholesky factorizations of symmetric matrices, then
\begin{align}\label{eq:bound_r}
    \norm{\bR_c-\bR_{s,i}}_F & \leq \norm{\bK_c^{-1}}_2\norm{\bR_c}_2\norm{\bK_{s,i}-\bK_c}_F \nonumber\\
    & = \norm{\bR_c^{-1}}_2^2\norm{\bR_c}_2\norm{\bK_{s,i}-\bK_c}_F.
\end{align}
Thus,
\begin{align}\nonumber
   &\norm{\bR_c^{-1}-\bR_{s,i}^{-1}}_2 \\ \label{eq:bound_r_inv}
    &\leq \frac{1+\sqrt{5}}{2}\max\left\{\norm{\bR_c^{-1}}_2^2,\norm{\bR_{s,i}^{-1}}_2^2\right\}\norm{\bR_c^{-1}}_2^2\norm{\bR_c}_2\norm{\bK_{s,i}-\bK_c}_F.
\end{align}
Also, from the definitions of $\bK_c$ and $\bK_{s,i}$, we know
\begin{align}
    \bK_c-\bK_{s,i} &= \bV_c^\tT\bV_c-\bV_{s,i}^\tT\bV_{s,i}\nonumber\\
    & = \bV_c^\tT\bV_c-\bV_{s,i}^\tT\bV_c+\bV_{s,i}^\tT\bV_c-\bV_{s,i}^\tT\bV_{s,i}.
\end{align}
Therefore, we have
\begin{align}\nonumber
    &\norm{\bK_c-\bK_{s,i}}_F \\
    &\leq \norm{\bV_c}_F\norm{\bV_c-\bV_{s,i}}_F+\norm{\bV_{s,i}}_F\norm{\bV_c-\bV_{s,i}}_F\nonumber\\
    &\leq (\norm{\bV_c}_F+\norm{\bV_{s,i}}_F)\norm{\bV_c-\bV_{s,i}}_F\nonumber\\
    &\leq \left(\alpha\sqrt{r}+\alpha\sqrt{r}+\delta\gamma\sqrt{N r}\right)\left(\alpha \max_i \norm{\bQ_c-\bQ_{s,i}}_F+\delta\gamma\sqrt{N r}\right)\nonumber\\ \label{eq:k_diff_bound}
     &= \alpha^2\left(2\sqrt{r}+\frac{\delta\gamma\sqrt{N r}}{\alpha}\right)\left(\max_i \norm{\bQ_c-\bQ_{s,i}}_F+\frac{\delta\gamma\sqrt{N r}}{\alpha}\right).
\end{align}

Also, note that $\bV_c=\bQ_c^\prime \bR_c$, hence $\norm{\bR_c}_2=\norm{\bV_c}_2 \leq \|\bV_c\|_F \leq \alpha\sqrt{r}$. Since
$\beta = \underset{t_o=1,\ldots,T_o}{\max} \norm{\bR_c^{-1^{(t_o)}}}_2$, from~\eqref{eq:bound_r_inv} and~\eqref{eq:k_diff_bound} we have
\begin{align}\nonumber
    &\norm{\bR_c^{-1}-\bR_{s,i}^{-1}}_2 \leq \frac{1+\sqrt{5}}{2}\max\left\{\norm{\bR_c^{-1}}_2^2,\norm{\bR_{s,i}^{-1}}_2^2\right\}\\ \nonumber
    &\beta^2\alpha\sqrt{r}\alpha^2\left(2\sqrt{r}+\frac{\delta\gamma\sqrt{N r}}{\alpha}\right)\left(\max_i \norm{\bQ_c-\bQ_{s,i}}_F+\frac{\delta\gamma\sqrt{N r}}{\alpha}\right)\\\nonumber
    & \leq \frac{1+\sqrt{5}}{2}\max\left\{\beta^2,\norm{\bR_{s,i}^{-1}}_2^2\right\}\alpha^3\beta^2\sqrt{r}\left(2\sqrt{r}+\frac{\delta\gamma\sqrt{N r}}{\alpha}\right)\\ \label{eq:2_31}
    & \qquad \times\left(\max_i \norm{\bQ_c-\bQ_{s,i}}_F+\frac{\delta\gamma\sqrt{N r}}{\alpha}\right).
\end{align}
The bound for $\norm{\bR_{s,i}^{-1}}_2$ is obtained as follows: The perturbation bound for singular values of a matrix~\cite{stewart1998perturbation} gives $\sigma_r(\bR_c)-\sigma_r(\bR_{s,i})\leq\norm{\bR_c-\bR_{s,i}}_2$, where $\sigma_r(\bR_c)$ and $\sigma_r(\bR_{s,i})$ represents the $r^{th}$ singular value of matrices $\bR_c$ and $\bR_{s,i}$ respectively. As $\sigma_r(\bR_c)=\norm{\bR_c^{-1}}_2^{-1}$ and $\sigma_r(\bR_{s,i})=\norm{\bR_{s,i}^{-1}}_2^{-1}$, we obtain that
\begin{align*}
    \norm{\bR_c^{-1}}_2^{-1}-\norm{\bR_{s,i}^{-1}}_2^{-1} \leq \norm{\bR_c-\bR_{s,i}}_2.
\end{align*}
Thus, from~\eqref{eq:bound_r}
\begin{align}\label{eq:2_32}
    \norm{\bR_c^{-1}}_2^{-1} &\leq \norm{\bR_{s,i}^{-1}}_2^{-1}+\norm{\bR_c^{-1}}_2^2\norm{\bR_c}_2\norm{\bK_{s,i}-\bK_c}_F\nonumber\\
    &\leq \norm{\bR_{s,i}^{-1}}_2^{-1}+\alpha^3\beta^2\sqrt{r}\left(2\sqrt{r}+\frac{\delta\gamma\sqrt{N r}}{\alpha}\right)\nonumber\\
    & \qquad \times \left(\max_i \norm{\bQ_c-\bQ_{s,i}}_F+\frac{\delta\gamma\sqrt{N r}}{\alpha}\right).
\end{align}
Using the assumption $\norm{\bQ_c-\bQ_{s,i}}_F+\frac{\delta \gamma \sqrt{N r}}{\alpha} \leq \frac{1}{2\alpha^2\beta^3\sqrt{r}(2\alpha\sqrt{r}+\delta \gamma \sqrt{N r})}$ in \eqref{eq:2_32}, we get
\begin{equation}
    \norm{\bR_c^{-1}}_2^{-1} \leq \norm{\bR_{s,i}^{-1}}_2^{-1}+\frac{1}{2\beta}.
\end{equation}
From our definition for $\beta$, we have $\beta^{-1} \leq \norm{\bR_c^{-1}}_2^{-1}$. So,
\begin{align}
     &\norm{\bR_{s,i}^{-1}}_2^{-1}+\frac{1}{2\beta} \geq   \beta^{-1}\nonumber\\
    \implies &\norm{\bR_{s,i}^{-1}}_2^{-1} \geq \frac{1}{2\beta} \implies \norm{\bR_{s,i}^{-1}}_2 \leq 2\beta\text{.}
\end{align}
Plugging-in the bound for $\norm{\bR_{s,i}^{-1}}_2$ into \eqref{eq:2_31}, we get
\begin{align}
    &\norm{\bR_c^{-1}-\bR_{s,i}^{-1}}_2 \leq \frac{1+\sqrt{5}}{2}\max\left\{\beta^2,\norm{\bR_{s,i}^{-1}}_2^2\right\}\alpha^3\beta^2\sqrt{r}\nonumber\\
    &\left(2\sqrt{r}+\frac{\delta\gamma\sqrt{N r}}{\alpha}\right)\left(\max_i \norm{\bQ_c-\bQ_{s,i}}_F+\frac{\delta\gamma\sqrt{N r}}{\alpha}\right)\nonumber\\
     & \leq  2\left(1+\sqrt{5}\right)\alpha^3\beta^4\sqrt{r}\left(2\sqrt{r}+\frac{\delta\gamma\sqrt{N r}}{\alpha}\right)\nonumber\\
     & \qquad \times\left(\max_i \norm{\bQ_c-\bQ_{s,i}}_F+\frac{\delta\gamma\sqrt{N r}}{\alpha}\right).
\end{align}

We know that for any matrix $\bX$ of rank $r$, $\norm{\bX}_F\leq\sqrt{r}\norm{\bX}_2$. Using this fact in \eqref{eq:2_15}, we obtain
\begin{align}\nonumber
\norm{\bQ_c^\prime-\bQ_{s,i}^\prime}_F  &\leq \sqrt{r}\norm{\bV_c-\bV_{s,i}}_F\cdot\max_i\norm{\bR_{s,i}^{-1}}_2 +\\ &\sqrt{r}\norm{\bV_c}_F\cdot\max_i\norm{\bR_c^{-1}-\bR_{s,i}^{-1}}_2.
\end{align}
Plugging in bounds for $\norm{\bV_c-\bV_{s,i}}_F$, $\norm{\bR_{s,i}^{-1}}_F$, $\norm{\bV_c}_F$, and $\norm{\bR_c^{-1}-\bR_{s,i}^{-1}}_F$, we have
\begin{align} \label{eq:bound_q_diff}
& \norm{\bQ_c^\prime-\bQ_{s,i}^\prime}_F   \leq 2\alpha\beta\sqrt{r} \left(\max_i \norm{\bQ_c-\bQ_{s,i}}_F+\frac{\delta\gamma\sqrt{N r}}{\alpha}\right) \nonumber\\
& + 2\left(1+\sqrt{5}\right)\alpha r\alpha^3\beta^4\sqrt{r}\left(2\sqrt{r}+\frac{\delta\gamma\sqrt{N r}}{\alpha}\right)\nonumber\\
& \qquad \times\left(\max_i \norm{\bQ_c-\bQ_{s,i}}_F+\frac{\delta\gamma\sqrt{N r}}{\alpha}\right)\nonumber\\
& = \Bigg(2\alpha\beta\sqrt{r} + 4(1+\sqrt{5})\alpha^4\beta^4r^2 + 2\left(1+\sqrt{5}\right) \nonumber\\
 &\alpha^4\beta^4r^{\frac{3}{2}}\frac{\delta\gamma\sqrt{N r}}{\alpha}\Bigg)\left(\max_i \norm{\bQ_c-\bQ_{s,i}}_F+\frac{\delta\gamma\sqrt{N r}}{\alpha}\right).
\end{align}
For the orthonormal matrix $\bQ_c^\prime$, we know $1=\norm{\bQ_c^\prime}_2= \norm{\bM\bQ_c\bR_c^{-1}}_2\leq \norm{\bM}_2\norm{\bR_c^{-1}}_2\leq \sum_{i=1}^N\norm{\bM_i}_2\norm{\bR_c^{-1}}_2\leq \alpha\beta$. Therefore $\alpha^4\beta^4 \geq \alpha\beta \geq 1$. Recall that
\begin{itemize}
    \item For S-DOT algorithm, we defined $\delta=\frac{\alpha}{\gamma\sqrt{Nr}}\epsilon^{T_o}(\frac{1}{3\alpha\beta\sqrt{r}})^{4T_o}$. Thus $\frac{\delta\gamma\sqrt{N r}}{\alpha}=\epsilon^{T_o}(\frac{1}{3\alpha\beta\sqrt{r}})^{4T_o}\leq \epsilon^{T_o}(\frac{1}{3})^{4T_o}\leq 1$.\\

    \item For SA-DOT algorithm, we defined $\delta=\frac{\alpha}{T_o\gamma\sqrt{Nr}}{\epsilon}^{T_o}(\frac{1}{3\alpha\beta\sqrt{r}})^{4t_o}$, where $\frac{\delta\gamma\sqrt{N r}}{\alpha}=\frac{{\epsilon}^{T_o}}{T_o}(\frac{1}{3\alpha\beta\sqrt{r}})^{4t_o}\leq \frac{{\epsilon}^{T_o}}{T_o}(\frac{1}3)^{4t_o}\leq 1$.
\end{itemize}
Plugging these facts into~\eqref{eq:bound_q_diff}, we can see that for both algorithms:
\begin{align*}
    \norm{\bQ_c^\prime-\bQ_{s,i}^\prime}_F  & \leq \Bigg(2\alpha^4\beta^4r^2 + 4(1+\sqrt{5})\alpha^4\beta^4r^2 + 2\left(1+\sqrt{5}\right) \nonumber\\
    & \alpha^4\beta^4r^2\Bigg)\left(\max_i \norm{\bQ_c-\bQ_{s,i}}_F+\frac{\delta\gamma\sqrt{N r}}{\alpha}\right) \nonumber\\
    &\leq \left(3\alpha\beta\sqrt{r}\right)^4\left(\max_i \norm{\bQ_c-\bQ_{s,i}}_F+\frac{\delta\gamma\sqrt{N r}}{\alpha}\right).
\end{align*}\qed

\section{Proof of Theorem 1}\label{app:thm1n2}
Let $\bQ_c$ be the estimate of $\bQ$ obtained after $T_o$ iterations of centralized OI. Now, we know that $\forall i$,
\begin{align}\label{eq:2_39}\nonumber
    \norm{\bQ\bQ^\tT-\bQ_{s,i}\bQ_{s,i}^\tT}_2 &\leq \norm{\bQ\bQ^\tT-\bQ_c\bQ_c^\tT}_2+\\
    &\qquad \norm{\bQ_c\bQ_c^\tT-\bQ_{s,i}\bQ_{s,i}^\tT}_2.
\end{align}
We drop the superscript of $\bQ_{s,i}$ here for convenience. The first term on the right-hand side of~\eqref{eq:2_39} is the error of centralized orthogonal iteration. It is proved in~\cite{van1983matrix} that $\norm{\bQ\bQ^\tT-\bQ_c\bQ_c^\tT}_2\leq c\left|\frac{\lambda_{r+1}}{r}\right|^{T_o}$ for some positive constant $c$. We now bound the second term in \eqref{eq:2_39}. We know $\norm{\bQ_c\bQ_c^\tT-\bQ_{s,i}\bQ_{s,i}^\tT}_2 \leq \norm{\bQ_c\bQ_c^\tT-\bQ_{s,i}\bQ_{s,i}^\tT}_F$. Now,
\begin{align*}
    \bQ_c\bQ_c^\tT-\bQ_{s,i}\bQ_{s,i}^\tT = \bQ_c\bQ_c^\tT-\bQ_{s,i}\bQ_{s,i}^\tT+\bQ_c\bQ_{s,i}^\tT-\bQ_c\bQ_{s,i}^\tT.
\end{align*}
Thus,
\begin{align}\label{eq:2_40}
    \norm{\bQ_c\bQ_c^\tT-\bQ_{s,i}\bQ_{s,i}^\tT}_F &\leq \left(\norm{\bQ_c}_2+\norm{\bQ_{s,i}}_2\right)\norm{\bQ_c-\bQ_{s,i}}_F\nonumber\\
                                                    &\leq 2\norm{\bQ_c-\bQ_{s,i}}_F .
\end{align}

We first prove that the assumption and hence the statement of Lemma 1 hold true for all $t_o < T_o$ in case of S-DOT. We initialize OI and S-DOT with same value $\bQ^{\text{init}}=\bQ_c^{(0)}=\bQ_{s,i}^{(0)}$. Therefore, we have $\norm{\bQ_c^{(0)}-\bQ_{s,i}^{(0)}}_F+\frac{\delta\gamma\sqrt{Nr}}{\alpha}=\frac{\delta\gamma\sqrt{Nr}}{\alpha} \leq \epsilon^{T_o}(\frac{1}{3})^{4T_o}$ $\leq \frac{1}{2\alpha^2\beta^3\sqrt{r}(2\alpha\sqrt{r}+\delta \gamma \sqrt{N r})}$. Thus the assumption of Lemma 1 is true for $t_o=0$. Through mathematical induction, it can be shown that the assumption of the lemma is true for all $t_o < T_o$. Now, applying Lemma \ref{lemm:c-dot} recursively for $(t_o + 1)$, we obtain
\begin{align}\label{eq:2_41}
    \norm{\bQ_c^{(t_o+1)}-\bQ_{s,i}^{(t_o+1)}}_F+\frac{\delta\gamma\sqrt{Nr}}{\alpha} &\leq \frac{\delta\gamma\sqrt{Nr}}{\alpha}\sum_{j=0}^{t_o}(3\alpha\beta\sqrt{r})^{4j}\nonumber\\
    \norm{\bQ_c^{(t_o)}-\bQ_{s,i}^{(t_o)}}_F &\leq \frac{\delta\gamma\sqrt{Nr}}{\alpha}\sum_{j=0}^{t_o}(3\alpha\beta\sqrt{r})^{4j}.
\end{align}
Note that $(3\alpha\beta\sqrt{r})^4>3$, and $\frac{1}{(3\alpha\beta\sqrt{r})^4}<\frac{1}{3}$. Then we have  $1-\frac{1}{(3\alpha\beta\sqrt{r})^4}>1-\frac{1}{3}=\frac{2}{3}$, and $\frac{(3\alpha\beta\sqrt{r})^4}{(3\alpha\beta\sqrt{r})^4-1}<\frac{3}{2}$. Applying geometric series, we obtain
\begin{align}\label{eq:2_42}
    \sum_{j=0}^{t_o}(3\alpha\beta\sqrt{r})^{4j}&=\frac{(3\alpha\beta\sqrt{r})^{4(t_o+1)}-1}{(3\alpha\beta\sqrt{r})^{4}-1}\nonumber\\
    &\leq(3\alpha\beta\sqrt{r})^{4t_o}\frac{(3\alpha\beta\sqrt{r})^4}{(3\alpha\beta\sqrt{r})^{4}-1}\nonumber\\
    &\leq \frac{3}{2}(3\alpha\beta\sqrt{r})^{4t_o}.
\end{align}
Plugging \eqref{eq:2_42} into \eqref{eq:2_41}, we have
\begin{equation}\label{eq:2_43}
    \norm{\bQ_c^{(t_o)}-\bQ_{s,i}^{(t_o)}}_F\leq \frac{3}{2}\frac{\delta\gamma\sqrt{Nr}}{\alpha}(3\alpha\beta\sqrt{r})^{4t_o}.
\end{equation}\label{eq:2_44}
We now plug in $\frac{\delta\gamma\sqrt{Nr}}{\alpha}=\epsilon^{T_o}\left(\frac{1}{3\alpha\beta\sqrt{r}}\right)^{4T_o}$ into \eqref{eq:2_43}. As $t_o<T_o$ and  $3\alpha\beta\sqrt{r}>3$, we have
\begin{align}
    \norm{\bQ_c^{(t_o)}-\bQ_{s,i}^{(t_o)}}_F & \leq \frac{3}{2}\epsilon^{T_o}\left(\frac{1}{3\alpha\beta\sqrt{r}}\right)^{4T_o}(3\alpha\beta\sqrt{r})^{4t_o}\nonumber\\
    & \leq \frac{3}{2}\epsilon^{T_o}\frac{(3\alpha\beta\sqrt{r})^{4t_o}}{(3\alpha\beta\sqrt{r})^{4T_o}} \leq \frac{3}{2}\epsilon^{T_o}.
\end{align}
From \eqref{eq:2_40}, we have
\begin{equation}
    \norm{\bQ_c\bQ_c^\tT-\bQ_{s,i}\bQ_{s,i}^\tT}_F \leq 2\norm{\bQ_c-\bQ_{s,i}}_F\leq 3\epsilon^{T_o}.
\end{equation}
Therefore,
\begin{equation}
    \norm{\bQ\bQ^\tT-\bQ_{s,i}\bQ_{s,i}^\tT}_2 \leq c\left|\frac{\lambda_{r+1}}{\lambda_r}\right|^{T_o}+3\epsilon^{T_o}.
\end{equation}
This completes the proof for S-DOT.

For SA-DOT, we prove convergence in a similar way. We first prove that the assumption and hence the statement of Lemma 1 hold true for all $t_o < T_o$. For same initialization for OI and SA-DOT $\bQ^{\text{init}}=\bQ_c^{(0)}=\bQ_{s,i}^{(0)}$, we have $\norm{\bQ_c^{(0)}-\bQ_{s,i}^{(0)}}_F+\frac{\delta^{(0)}\gamma\sqrt{Nr}}{\alpha}=\frac{\delta^{(0)}\gamma\sqrt{Nr}}{\alpha} \leq \frac{{\epsilon}^{T_o}}{T_o}(\frac{1}3)^{4t_o}\leq \frac{1}{2\alpha^2\beta^3\sqrt{r}(2\alpha\sqrt{r}+\delta \gamma \sqrt{N r})}$. Thus the assumption of Lemma 2 is true for $t_o=0$. Through mathematical induction, it can be shown that the assumption of the lemma is true for all $t_o < T_o$. Next, applying Lemma \ref{lemm:c-dot} recursively for $T_o^{th}$ iteration
\begin{equation}\label{eq:2_47}
    \norm{\bQ_c^{(T_o)}-\bQ_{s,i}^{(T_o)}}_F+\frac{\delta^{(T_o)}\gamma\sqrt{Nr}}{\alpha} \leq \frac{\gamma\sqrt{Nr}}{\alpha}\sum_{j=0}^{T_o}(3\alpha\beta\sqrt{r})^{4j}\delta^{(j)}.
\end{equation}
Plugging in $\delta^{(j)}$ into \eqref{eq:2_47}, where $\delta^{(j)} \coloneqq \frac{\alpha}{T_o\gamma\sqrt{N r}}\epsilon^{T_o}\left(\frac{1}{3\sqrt{r}\alpha\beta}\right)^{4j}$, and $\epsilon \in (0,1)$, we obtain
\begin{align}
    \frac{\gamma\sqrt{Nr}}{\alpha}\sum_{j=0}^{T_o}\left(3\alpha\beta\sqrt{r}\right)^{4j} \delta^{(j)}&=\sum_{i=0}^{T_o}\left(3\alpha\beta\sqrt{r}\right)^{4j} \frac{\epsilon^{T_o}}{T_o}\left(\frac{1}{3\sqrt{r}\alpha\beta}\right)^{4j}\nonumber\\
    & =\frac{\epsilon^{T_o}}{T_o}\sum_{i=0}^{T_o} 1 =\frac{(T_o+1)}{T_o}\epsilon^{T_o} \leq \epsilon^{T_o}\text{.}
\end{align}
Thus,
\begin{equation}
    \norm{\bQ_c^{(T_o)}-\bQ_{s,i}^{(T_o)}}_F  \leq \epsilon^{T_o},
\end{equation}
and from \eqref{eq:2_40}, we have
\begin{equation}
    \norm{\bQ_c\bQ_c^\tT-\bQ_{s,i}\bQ_{s,i}^\tT}_F \leq 2\norm{\bQ_c-\bQ_{s,i}}_F\leq 2\epsilon^{T_o}.
\end{equation}
Thus,
\begin{align*}
    \norm{\bQ\bQ^\tT-\bQ_{s,i}\bQ_{s,i}^\tT}_2 \leq c\left|\frac{\lambda_{r+1}}{\lambda_r}\right|^{T_o}+2\epsilon^{T_o}.
\end{align*}
This completes the proof for SA-DOT. \qed
\end{appendices}

\balance

\end{document}